\documentclass[review,hidelinks,onefignum,onetabnum]{siamart220329}

\usepackage{lipsum}
\usepackage{amsfonts}
\usepackage{graphicx}
\usepackage{epstopdf}
\usepackage{algorithmic}

\usepackage{amssymb}
\usepackage{amsmath}
\usepackage{mathtools}
\usepackage{xcolor}
\usepackage{subfigure}

\newcommand{\D}{\mathcal{D}}
\newcommand{\T}{\mathcal{T}}
\newcommand{\F}{\mathcal{F}}
\newcommand{\B}{\mathcal{B}}
\newcommand{\R}{\mathbb{R}}
\DeclareMathOperator*{\argmin}{arg\,min}

\ifpdf
  \DeclareGraphicsExtensions{.eps,.pdf,.png,.jpg}
\else
  \DeclareGraphicsExtensions{.eps}
\fi

\newsiamremark{remark}{Remark}
\newsiamremark{hypothesis}{Hypothesis}
\crefname{hypothesis}{Hypothesis}{Hypotheses}
\newsiamthm{claim}{Claim}

\headers{L-HYDRA}{Z. Zou and G. E. Karniadakis}

\title{L-HYDRA: Multi-Head Physics-Informed Neural Networks}

\author{Zongren Zou\thanks{Division of Applied Mathematics, Brown University, Providence, RI 02912, USA (\email{zongren\_zou@brown.edu}).}
\and George Em Karniadakis\thanks{Corresponding author. Division of Applied Mathematics, Brown University, Providence, RI 02912, USA (\email{george\_karniadakis@brown.edu}).}}

\usepackage{amsopn}

\ifpdf
\hypersetup{
  pdftitle={L-HYDRA},
  pdfauthor={Z. Zou and G.E. Karniadakis}
}
\fi

\begin{document}

\maketitle
\begin{abstract}
We introduce multi-head neural networks (MH-NNs) to physics-informed machine learning, which is a type of neural networks (NNs) with all nonlinear hidden layers as the body and multiple linear output layers as multi-head. Hence, we construct multi-head physics-informed neural networks (MH-PINNs) as a potent tool for multi-task learning (MTL), generative modeling, and few-shot learning for diverse problems in scientific machine learning (SciML). MH-PINNs connect multiple functions/tasks via a shared body as the basis functions as well as a shared distribution for the head. The former is accomplished by solving multiple tasks with MH-PINNs with each head independently corresponding to each task, while the latter by employing normalizing flows (NFs) for density estimate and generative modeling. To this end, our method is a two-stage method, and both stages can be tackled with standard deep learning tools of NNs, enabling easy implementation in practice. MH-PINNs can be used for various purposes, such as approximating stochastic processes, solving multiple tasks synergistically, providing informative prior knowledge for downstream few-shot learning tasks such as meta-learning and transfer learning, learning representative basis functions, and uncertainty quantification. We demonstrate the effectiveness of MH-PINNs in five benchmarks, investigating also the possibility of synergistic learning in regression analysis. We name the open-source code ``Lernaean Hydra" (L-HYDRA), since this mythical creature possessed many heads for performing important multiple tasks, as in the proposed method. 
\end{abstract}



\section{Introduction}

Learning across tasks has drawn great attention recently in deep learning and is an emerging theme in scientific machine learning (SciML), due to the fact that several classes of scientific problems are similar and/or related intrinsically by their common physics. Intuitively, if tasks are similar, e.g., in the context of approximating stochastic processes \cite{yang2020physics}, learning solution operators of ordinary/partial differential equations (ODEs/PDEs) \cite{lu2021learning}, and solving parametric PDEs \cite{wang2021learning, khoo2021solving, bhattacharya2020model}, it may be beneficial to relate them in the modeling, algorithm design, and/or solving procedure. In this regard, machine learning solvers, developed rapidly in the past few years, are considerably more flexible and of higher potential compared to traditional numerical solvers. Significant progress has been witnessed in the general area, including meta-learning for solving ODEs/PDEs \cite{meng2022learning, liu2022novel, penwarden2021physics, chen2021learning}, transfer learning for physics-informed neural networks (PINNs) \cite{bahmani2021training, desai2021one}, transfer learning for domain shift in solving PDEs \cite{goswami2022deep}, multi-task learning for PINNs \cite{thanasutives2021adversarial}, and generative methods for solving stochastic differential equations (SDEs) \cite{yang2020physics, zhong2022pi, guo2022normalizing}. More recently, operator learning~\cite{lu2021learning, li2020fourier} in which direct operator mapping is learned and subsequently used for other tasks in one-shot format has attracted a lot of attention. 

Multi-head neural networks (MH-NNs) fit perfectly different scenarios of learning across tasks. They were originally proposed as members of hard-parameter sharing neural networks (NNs) for deep multi-task learning (MTL) \cite{Caruana1993MultitaskLA}, in which multiple tasks, denoted as $\T_k, k=1,...,M$, where $M$ is the number of total tasks, are solved simultaneously. 
The general goals of using MH-NNs in MTL are diverse: achieving better performance for all tasks, learning good and useful representations for downstream tasks, and/or boosting the learning of main tasks with the help of auxiliary tasks. Moreover, although originally designed for solving multiple tasks, MH-NNs in recent years have also been extensively used for meta-learning. For example, in \cite{wang2021bridging}, the connection between MTL and meta-learning was analyzed, and meta-learning algorithms for MH-NN were discussed; in \cite{lin2021to}, it was shown that MH-NNs, trained in MTL fashion also perform task-specific adaptation in meta-learning; \cite{raghu2019rapid} argued that the effectiveness of model-agnostic meta-learning \cite{finn2017model}, a well-known meta-learning algorithm, may be due to successfully learned good representations rather than learned adaptation, and MH-NNs were used to study the detailed contributions of NNs in fast task adaptations. 
Overall, it is commonly acknowledged in the literature that when used to solve previous tasks, MH-NNs are capable of distilling useful shared information and storing it in their bodies and heads.

\begin{figure}[ht]
    \centering
    \includegraphics[scale=0.45]{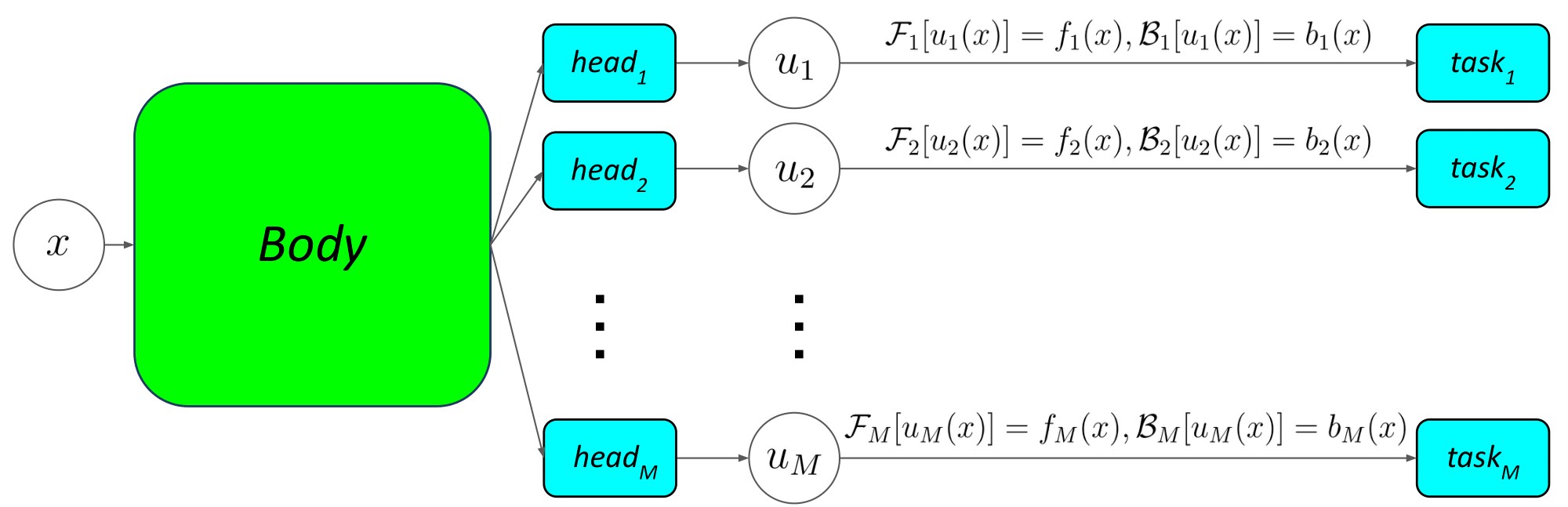}
    \caption{Schematic view of the structure of multi-head physics-informed neural networks (MH-PINNs) with $M$ different heads, which are built upon conventional multi-head neural networks. The shared layers are often referred to as \textit{body} and the task-specific layer as \textit{head}. Generally, $u_k, k=1,...,M$ represent $M$ solutions to $M$ different ODEs/PDEs, formulated in Eq.~\eqref{eq:problem}, which may differ in source terms $f_k$, boundary/initial condition terms $b_k$, or differential operator $\F_k$.}
    \label{fig:schematic}
\end{figure}

In this paper, we develop MH-NNs for physics-informed machine learning \cite{karniadakis2021physics}, propose multi-head physics-informed neural networks (MH-PINNs), and further investigate their applicability and capabilities to MTL, generative modeling, and meta-learning. 
A MH-PINN, as shown in Fig.~\ref{fig:schematic}, is built upon a conventional MH-NN and consists of two main parts, the \textit{body} and multiple \textit{heads}, and each head connects to a specific ODE/PDE task. Many architecture splitting strategies for MH-NNs are adopted in different applications scenarios; e.g., for some computer vision problems, a NN is split such that the body consists of convolutional layers and is followed by fully-connected layers as heads. In this paper, however, we choose the simplest one, i.e., the body consists of all nonlinear layers and the head is the last linear layer, for the following two reasons: (1) the dimensionality of the head is reduced, which enables fast density estimation (see next section); and (2) the body spontaneously provides a set of basis functions.

The novelty and major contributions of this work are as follows:
\begin{enumerate}
    \item We propose a new physics-informed generative method using MH-PINNs for learning stochastic processes from data and physics.
    \item We propose a new method for physics-informed few-shot regression problems with uncertainty quantification using MH-PINNs.
    \item We study and demonstrate the effectiveness of MTL and synergistic learning with MH-NNs in regression problems.
\end{enumerate}

The paper is organized as follows. In Sec.~\ref{sec:method}, we present the problem formulation, details of MH-PINNs, and the general methodology, including how to use MH-PINNs for MTL, generative modeling, downstream few-shot physics-informed learning with uncertainty quantification (UQ). In Sec.~\ref{sec:related}, we discuss existing research closely related to our work and compare them conceptually. In Sec.~\ref{sec:examples}, we test MH-PINNs with five benchmarks, each of which corresponds to one or more learning purposes, e.g., MTL and generative modeling. In Sec.~\ref{sec:MTL}, we investigate MTL and synergistic learning with the function approximation example. We conclude and summarize in Sec.~\ref{sec:dicussion}. The details of our experiments, such as NN architectures and training strategies, can be found in Appendix~\ref{sec:nn} and~\ref{sec:hmc}, as well as in the L-HYDRA open-source codes on GitHub, which will be released once the paper is accepted.

\section{Methodology}
\label{sec:method}
We assume that we have a family of tasks, $\{\mathcal{T}_k\}_{k=1}^M$, each of which is associated with data $\mathcal{D}_k, k=1,..., M$. The primary focus of this paper is on scientific computing and ODEs/PDEs, and therefore we further assume $\{\mathcal{T}_k\}_{k=1}^M$ are physics-informed regression problems \cite{karniadakis2021physics}.

Consider a PDE of the following form:
\begin{subequations}\label{eq:problem}
    \begin{align}
        \mathcal{F}_k[u_k(x)] &= f_k(x), x\in\Omega_k,\\
        \mathcal{B}_k[u_k(x)] &= b_k(x), x\in\partial\Omega_k,
    \end{align}
\end{subequations}
where $k$ denotes the index of the task and $k=1,..., M$, $x$ is the general spatial-temporal coordinate of $D_x$ dimensions, $\Omega_k$ are bounded domains, $f_k$ and $u_k$ are the $D_u$-dimensional source terms and solutions to the PDE, respectively, $\mathcal{F}_k$ are general differential operators, $\mathcal{B}_k$ are general boundary/initial condition operators, and $b_k$ are boundary/initial condition terms. For simplicity, throughout this paper, the domain and the boundary/initial operator, denoted as $\Omega$ and $\mathcal{B}$, are assumed to be the same for all tasks, and the solutions $u_k$ to be task-specific.
The task $\T_k$ is described as approximating $u_k$, and/or $f_k$, and/or $\mathcal{F}_k$, and/or $b_k$, from data $\D_k$ and Eq.~\eqref{eq:problem}.

Traditional numerical solvers often tackle $\{\mathcal{T}_k\}_{k=1}^M$ independently, without leveraging or transferring knowledge across tasks. The PINN method \cite{raissi2019physics} was designed to solve ODEs/PDEs independently using NNs, which, however, yields $M$ uncorrelated results. 
In this paper instead we treat $\{\T_k\}_{k=1}^M$ as a whole and connect them with MH-PINNs, the architecture of which, shown in Fig.~\ref{fig:schematic}, enforces basis-functions-sharing predictions on the solutions $u_k$. In addition to the informative representation/body, we further relate $\{\mathcal{T}_k\}_{k=1}^M$ by assuming that their corresponding heads in MH-PINNs, denoted as $\{H_k\}_{k=1}^M$, are samples of a random variable with unknown probability density function (PDF), denoted as $H$ and $p(H)$, respectively. The shared body and a generative model of $H$ immediately form a generative model of the solution $u$, and generators of the source term $f$ and the boundary/initial term $b$ as well by substituting $u$ into Eq.~\eqref{eq:problem} and automatic differentiation \cite{tensorflow2015-whitepaper}, from which a generative method for approximating stochastic processes is seamlessly developed.

\begin{figure}[ht]
    \centering
    \includegraphics[scale=0.5]{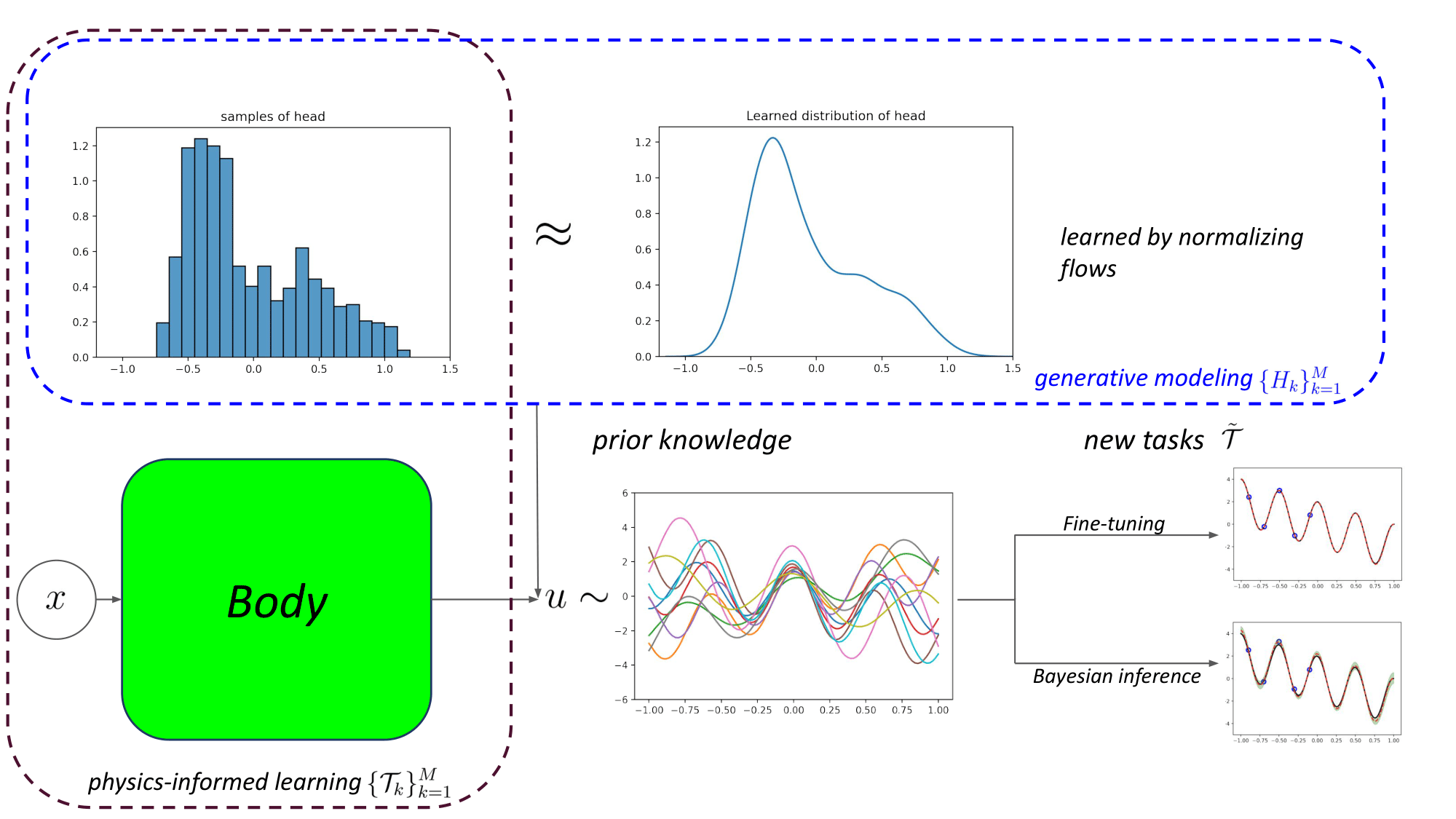}
    \caption{Schematic view of the learning framework and the proposed method. Three general types of learning are addressed: physics-informed learning, generative modeling, and few-shot learning. The physics-informed learning is performed with MH-PINNs; the generative modeling is done afterwards by density estimate over the head via normalizing flows (NFs); in the end the few-shot physics-informed learning is accomplished with prior knowledge obtained from previous two via either fine-tuning with the learned regularization or Bayesian inference with the learned prior distribution. The body represents the set of basis functions learned from solving $\{\T_k\}_{k=1}^M$ with MH-PINNs, and the density of the head, estimated from its samples using NFs, acts as the regularization, the prior distribution, or the generator together with the body, depending on the usage of MH-PINNs in applications.}
    \label{fig:methodology}
\end{figure}

Generators of $u$, $f$ and $b$, as discussed in \cite{meng2022learning}, are able to provide an informative prior distribution in physics-informed Bayesian inference \cite{yang2021b, linka2022bayesian} as well as in UQ for SciML \cite{zou2022neuraluq, psaros2022uncertainty}, where the informative prior compensates for the insufficiency of observational data to address the physics-informed learning problems with even a few noisy measurements. In this paper, we generalize such problem to deterministic cases as well, where the data is noiseless and methods and results are deterministic, and refer to it as \textit{few-shot physics-informed learning}. The general idea is to apply prior knowledge learned from connecting $\{\mathcal{T}_k\}_{k=1}^M$ with MH-PINNs to new tasks, denoted as $\tilde{\T}$, associated with insufficient data $\tilde{\D}$, for \textit{{accurate}} and \textit{{trustworthy}} predictions. The schematic view of the learning framework is illustrated in Fig.~\ref{fig:methodology}, and the details are explained next.

\subsection{Multi-head physics-informed neural networks (MH-PINNs)}
\label{subsec:mhpinn}

Hard parameter sharing is the most commonly used approach when MTL with NNs are considered, and MH-NNs, as its simplest instance, are frequently adopted \cite{Caruana1993MultitaskLA, ruder2017overview}. A MH-PINN, as described earlier, is composed of a body and multiple heads. We denote by $\Phi$ the body and by $H_k$ the head for $\T_k$. Notice that here $\Phi: \R^{D_x} \rightarrow \R^{L}$ is a function parameterized by a neural network with parameter $\theta$, and $H_k \in \R^{L+1}$ is a vector, where $L$ is the number of neurons on the last layer of the body. Let us define $H_k=[h_k^0, h_k^1, ..., h_k^L]^T$, $\Phi(x) = [\phi^1(x), ..., \phi^L(x)]^T$, where $\phi: \R^{D_x} \rightarrow \R$, and then the surrogate for the solution in $\T_k$ can be rewritten as $\hat{u}_k(x) = h_k^0 + \sum_{l=1}^L h^l_k \phi^l(x), \forall x\in\Omega$. The approximated source terms and boundary/initial terms are derived from Eq.~\eqref{eq:problem} accordingly. In the MTL framework , given data $\{\D_k\}_{k=1}^M$ and physics Eq.~\eqref{eq:problem}, the loss function $\mathcal{L}$ is formulated as follows:
\begin{equation}\label{eq:loss}
    \mathcal{L}(\{\D_k\}_{k=1}^M; \theta, \{H_k\}_{k=1}^M) = \frac{1}{M}\sum_{k=1}^M \mathcal{L}_k(\D_k; \theta, H_k),
\end{equation}
where $\mathcal{L}_k$ denotes the common loss function in PINNs. Conventionally, the data for $\T_k$ is expressed as $\D_k = \{\D_k^f, \D_k^b, \D_k^u\}$, where $\D_k^f = \{x_k^i, f_k^i\}_{i=1}^{N_k^f}$, $\D_k^b = \{x_k^i, b_k^i\}_{i=1}^{N_k^b}$ and $\D_k^u = \{x_k^i, u_k^i\}_{i=1}^{N_k^u}$, and $\mathcal{L}_k$ as follows:
\begin{equation}
\begin{aligned}
    \mathcal{L}_k(\D_k; \theta, H_k) = &\frac{w_k^f}{N_k^f}\sum_{i=1}^{N_k^f} ||\F_k(\hat{u}_k(x_k^i)) - f_k^i||^2 + \frac{w_k^b}{N_k^b}\sum_{i=1}^{N_k^b} ||\B(\hat{u}_k(x_k^i)) - b_k^i||^2 \\
    + &\frac{w_k^u}{N_k^u}\sum_{i=1}^{N_k^u} ||\hat{u}_k(x_k^i) - u_k^i||^2 + \mathcal{R}(\theta, H_k),
\end{aligned} 
\end{equation}
where $||\cdot||$ represents a properly chosen norm,  $\mathcal{R}(\cdot)$ is a regularization method over the parameters of NNs, $N_k^f, N_k^b, N_k^u$ are the numbers of data points for $f_k, b_k, u_k$, and $w_k^f, w_k^b, w_k^u$ are weights to balance different terms in the loss function.

\subsection{Generative modeling and normalizing flows (NFs)}
\label{subsec:nf}

As mentioned earlier, MH-PINNs connect $\{\T_k\}_{k=1}^M$ by making two assumptions: (1) the solutions $u_k, k=1,...,M$ share the same set of basis functions, $\Phi$; and (2) the corresponding coefficients are samples of the same random variable, $H$. In \cite{desai2021one}, $\Phi$ was used as a carrier of prior knowledge from $\{\T_k\}_{k=1}^M$ in downstream physics-informed learning tasks. In this work, we extend it by utilizing the information from the head as well by estimating the PDF and a generator of $H$ from its samples, $\{H_k\}_{k=1}^M$, using normalizing flows (NFs). The interested readers are directed to \cite{papamakarios2021normalizing, kobyzev2020normalizing} for reviews of NFs as well as \cite{dinh2016density, papamakarios2017masked, kingma2016improved} for developments of some popular NFs.

We choose NFs over other commonly used generative models, e.g.,  generative adversarial networks (GANs) \cite{goodfellow2020generative}, variational auto-encoders (VAEs) \cite{kingma2013auto}, or diffusion models \cite{ho2020denoising}, because the NF serves as both a density estimator and a generator. The former is able to provide proper regularization in the downstream few-shot physics-informed learning tasks, while the latter leads to a physics-informed generative method for approximating stochastic processes. 
It is worth noting that in previous works on physics-informed generative methods \cite{yang2020physics, zhong2022pi, guo2022normalizing}, NNs are trained by measurements over $u_k$, and/or $f_k$, and/or $b_k$. Our model, on the other hand, learns through samples of the head, which is obtained from MTL in the first step. This learning strategy brings two substantial advantages: (1) flexibility in dealing with unstructured data, e.g., inconsistent measurements across tasks; (2) simplicity and controlability of the training by decoupling the physics-informed learning and the generative modeling.

\subsection{Prior knowledge utilized in the downstream tasks}
\label{subsec:downstream}
Here, we describe details on how to utilize the prior knowledge stored in MH-PINNs, for downstream few-shot physics-informed learning task, $\tilde{\T}$, which is defined the same as all other tasks in the upstream training, but with much fewer measurements. Training of MH-PINNs and NFs yield a body, $\Phi$, samples of heads, $\{H_k\}_{k=1}^M$, and an estimated PDF of the head, $\hat{p}(H) \approx p(H)$. In solving $\tilde{\T}$ with $\tilde{\D}$, we fix the body $\Phi$ and find the head $\tilde{H}$ that best explains the data $\tilde{\D}$ and the physics in Eq.~\eqref{eq:problem}. 
Noiseless and noisy data are considered in this paper: for noiseless data, regular NN training is performed on the head for new tasks to provide deterministic predictions, where the learned PDF of the head, $\hat{p}(H)$, acts as a regularization term in the loss function; for noisy data, Bayesian inference is performed on the head as well, in which $\hat{p}(H)$ denotes the prior distribution. Details are presented in the following.

\subsubsection{Regularization in optimization}
Limited data in few-shot learning often leads to over-fitting and/or poor inter-/extrapolation performance. In this regard, regularizing the head according to its PDF is able to prevent over-fitting and provide additional prior knowledge for better inter-/extrapolation performance. The optimization problem is cast as
\begin{equation}\label{eq:optimization}
\begin{aligned}
    \tilde{H}^* = \argmin_{\tilde{H}} \mathcal{L}^*(\tilde{\D};\tilde{H}), \text{where }\mathcal{L}^*(\tilde{\D};\tilde{H}) = \mathcal{L}(\tilde{\D}; \tilde{H}) - \alpha\log p(\tilde{H})
    \\\approx \mathcal{L}(\tilde{\D}; \tilde{H}) - \alpha\log\hat{p}(\tilde{H}),
\end{aligned}
\end{equation}
where $\mathcal{L}$ is the regular loss function in physics-informed learning for data $\tilde{\D}$ and parameter $\tilde{H}$, and $\alpha\geq 0$ is the coefficient to adjust the regularization effect. Problem~\eqref{eq:optimization} in this work is solved with gradient descent.

\subsubsection{Prior distribution in Bayesian inference}
As opposed to point estimate obtained by solving the optimization problem~\eqref{eq:optimization}, the posterior distribution of the head in $\tilde{\T}$ is obtained using Bayesian inference. Similar as in \cite{meng2022learning, zou2022neuraluq}, the posterior distribution of $\tilde{H}$ is established as follows:
\begin{equation}\label{eq:posterior}
    p(\tilde{H}|\tilde{\D}) \propto p(\tilde{\D}|\tilde{H})p(\tilde{H}) \approx p(\tilde{\D}|\tilde{H})\hat{p}(\tilde{H}),
\end{equation}
where $p(\tilde{H}|\tilde{\D})$ is the posterior distribution, $p(\tilde{\D}|\tilde{H})$ is the likelihood distribution, which is often assumed to be independent Gaussian over all measurements in $\tilde{\D}$, and $\hat{p}$ is the estimated PDF of the head via NFs. Intractability of distribution~\eqref{eq:posterior} requires approximation methods, among which Markov chain Monte Carlo methods, such as Hamiltonian Monte Carlo (HMC) \cite{neal2011mcmc}, generally provide the most accurate estimation. Moreover, the relatively low dimension of $\tilde{H}$ also enables the use of Laplace's approximation (LA) \cite{kass1990validity}, which is employed in this paper as an alternative to HMC.
\section{Related works}
\label{sec:related}

Deep NNs in recent years have been extensively investigated for solutions of ODEs/PDEs, SDEs as well as operator learning. Although not explicitly introduced as MH-PINNs, MH-NNs were first used to solve ODEs/PDEs by \cite{desai2021one}, in which MH-PINNs were pre-trained on multiple similar tasks, and then the heads were discarded while the body was kept and transferred to solving new tasks, by either least square estimate for linear ODEs/PDEs, or fine-tuning with gradient descent for nonlinear ones. 
In \cite{desai2021one}, a one-shot transfer learning algorithm for linear problems was proposed but  other potential uses of MH-NNs, e.g., MTL and generative modeling, were not discussed, as opposed to the work presented herein. Furthermore, \cite{desai2021one} focused only on fast and deterministic predictions with high accuracy using sufficient clean data, while in this paper, we study the applicability of MH-NNs to few-shot physics-informed learning as well, where data is insufficient and/or noisy, and address such cases with UQ. We note that MH-NN was also used as a multi-output NN in \cite{yang2022multi}, which, however, focused on solving single tasks and obtaining uncertainties.

Generative modeling in the context of scientific computing has also been studied recently, and a few attempts for adopting deep generative NNs to SciML problems have been made in \cite{yang2020physics, zhong2022pi, guo2022normalizing}, most of which have focused on approximating stochastic processes and on solving SDEs. We propose a new physics-informed generative method, as an alternative to the current ones, using MH-PINNs, and test it in approximating stochastic processes. In this regard, our method is functionally the same as the current ones, but technically different. 
All previous methods address physics-informed generative modeling using end-to-end learning strategies by coupling two dissimilar types of learning, physics-informed learning and generative modeling, which may be problematic for implementation and usage in practice when either type of learning becomes more complicated. Our method, on the other hand, addresses the problem in an entirely new angle by {\em decoupling} those two: physics-informed learning is performed first and is followed by learning generators. To this end, our method is a two-step method, and with the help of well-developed algorithms from both fields, our method has advantages both in flexibility and simplicity in implementation.


\section{Results}
\label{sec:examples}

In this section, we test our method using five benchmarks. The first one is a pedagogical function regression, in which we aim to demonstrate the basic applicability and capabilities of our method, showing the importance of incorporating the distribution of the head in the downstream tasks and in obtaining results with or without uncertainty. The second example is a nonlinear ODE system, in which we test our method in approximating stochastic processes through a differential operator, compare different NFs, and eventually compare our method with another well-known physics-informed generative model, physics-informed GANs (PI-GANs) \cite{yang2020physics} in generative modeling. The third is a 1-D nonlinear reaction-diffusion equation, the fourth is a 2-D nonlinear Allen-Cahn equation, and the fifth is the 2-D stochastic Helmholtz equation with $20$ dimensions. 
In all examples unless stated otherwise, data for $\{\D_k\}_{k=1}^M$ are noise-free and task-wisely sufficient, while $\tilde{\D}$ in downstream tasks is insufficient, which makes the downstream tasks of the few-shot type. In addition, except for the first example, results from Bayesian inference are obtained by employing HMC, and the predicted mean denoted as $\mu$ and predicted standard deviation denoted as $\sigma$ are computed from the posterior samples of functions or unknown parameters. The predicted uncertainty is defined as $2 \sigma$ in this paper.

\subsection{Function approximation}
\label{subsec:example_1}
We start with a function regression problem using only data and no physics, which is a degenerate instance of Eq.~\eqref{eq:problem} with $\F_k$ being fixed as an identity operator, no $\B$ and $b_k$, and $u_k = f_k$ being task-specific. In this case, $\D_k$ and $\tilde{\D}$ are given as $\{(x_k^i, f_k^i)\}_{i=1}^{N_k}$ and $\{(x^i, f^i)\}_{i=1}^{N}$, respectively, and $\T_k$ and $\tilde{\T}$ are defined as approximating functions $f_k$ and $\tilde{f}$ from $\D_k$ and $\tilde{\D}$, respectively. The stochastic function $f$ in this example is defined as follows:
\begin{equation}\label{eq:example_1}
    \begin{aligned}
        &f(x) = A \cos(\omega x) + 2 \beta x, x\in[-1, 1],\\
        &A\sim U[1, 3), \omega\sim U[2\pi, 4\pi), \mathcal{P}(\beta=\pm1) = 0.5,
    \end{aligned}
\end{equation}
where $U$ stands for uniform distribution and $\mathcal{P}(\Xi)$ is defined as the probability of the event $\Xi$. Our goal is to approximate $f$ from data with MH-NNs and NFs, and solving the downstream few-shot regression tasks $\tilde{\T}$ as well, in which two functions, $2\cos(2\pi x) + 2x$ and $2\cos(4\pi x) - 2x$, are regressed from $4$ and $5$ measurements equidistantly distributed on $[-0.9, -0.1]$, respectively.

\begin{figure}[ht]
    \centering
    \subfigure[]{
    \includegraphics[scale=.25]{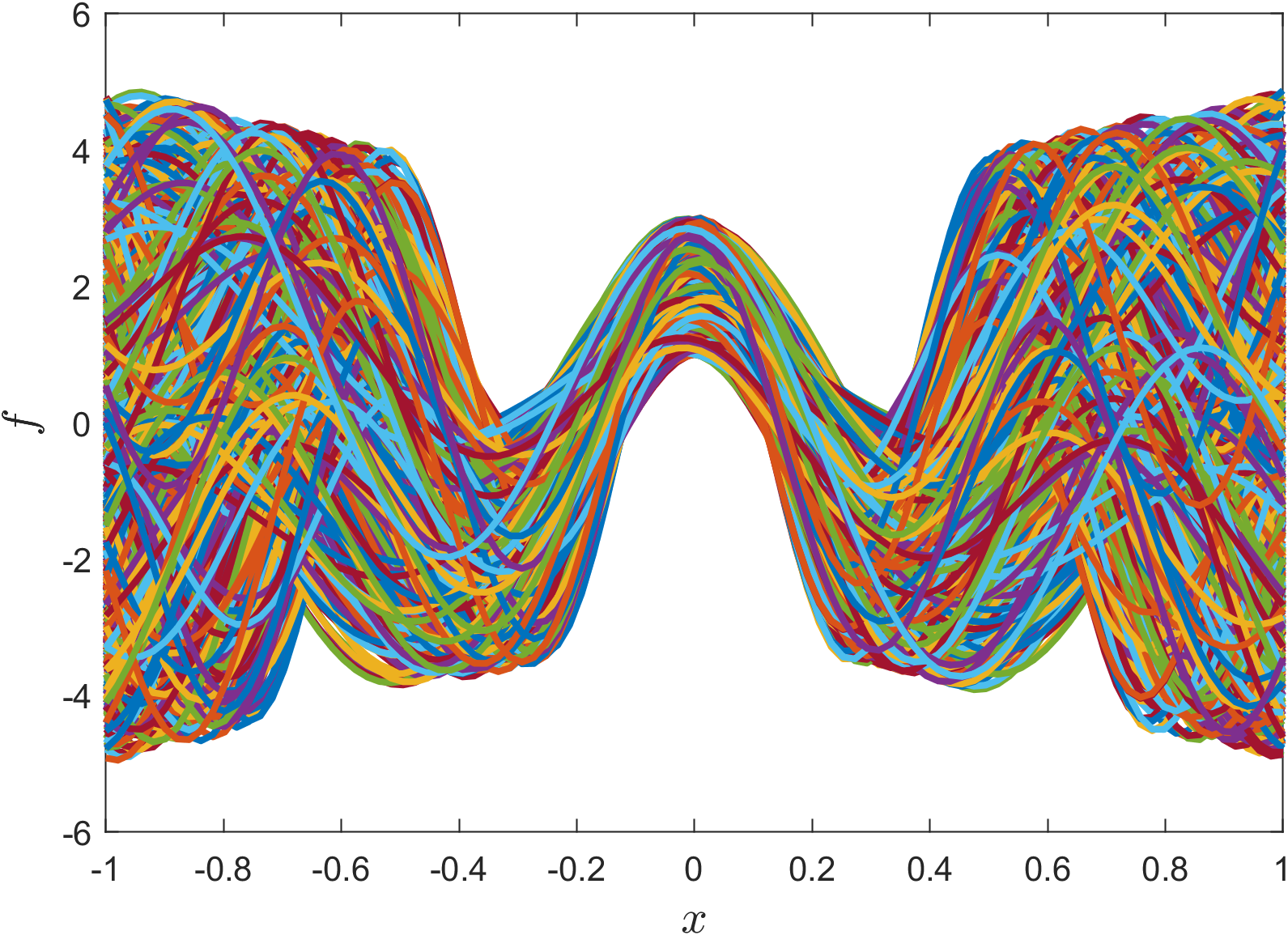}
    \includegraphics[scale=.25]{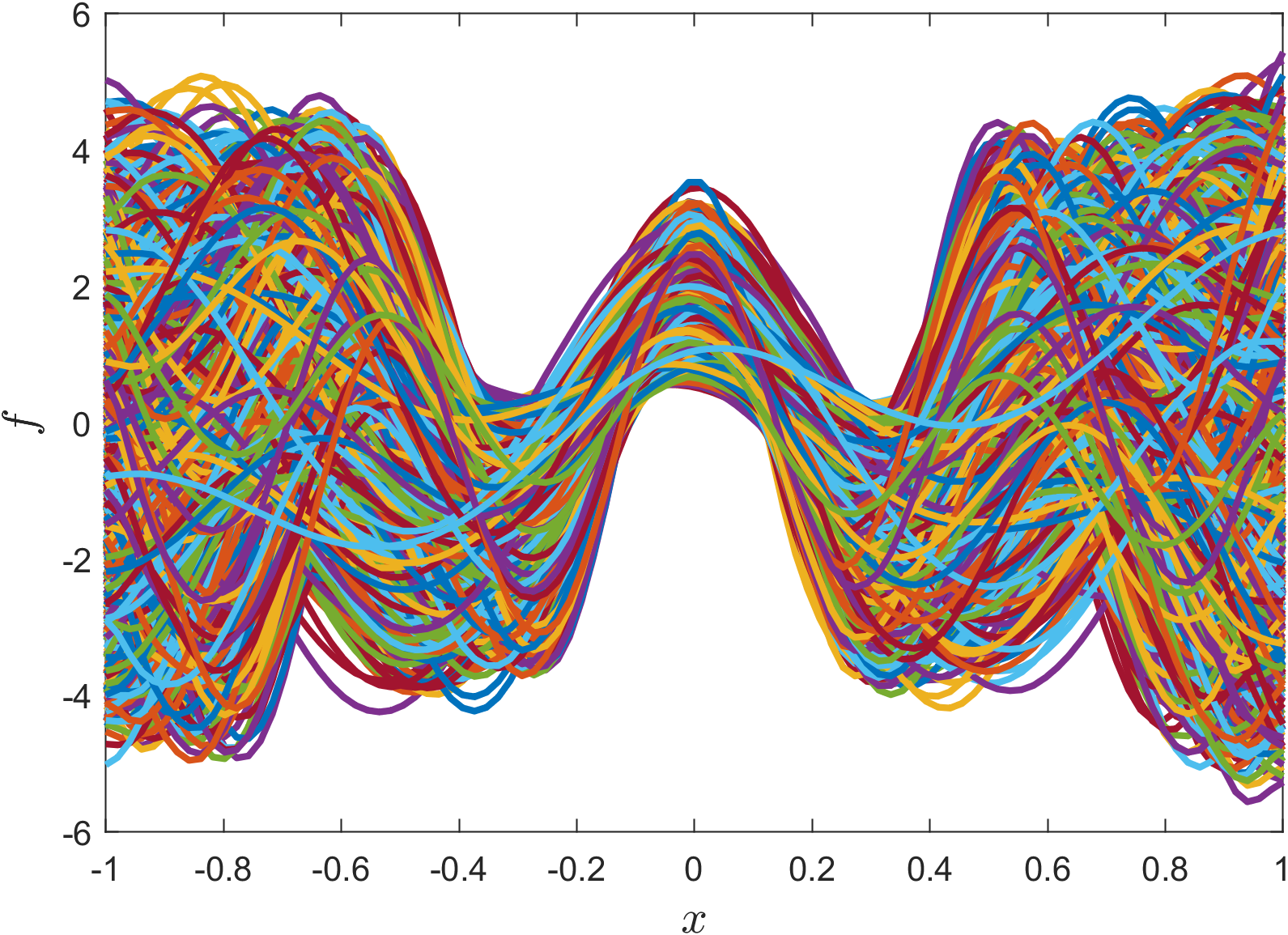}
    \includegraphics[scale=.25]{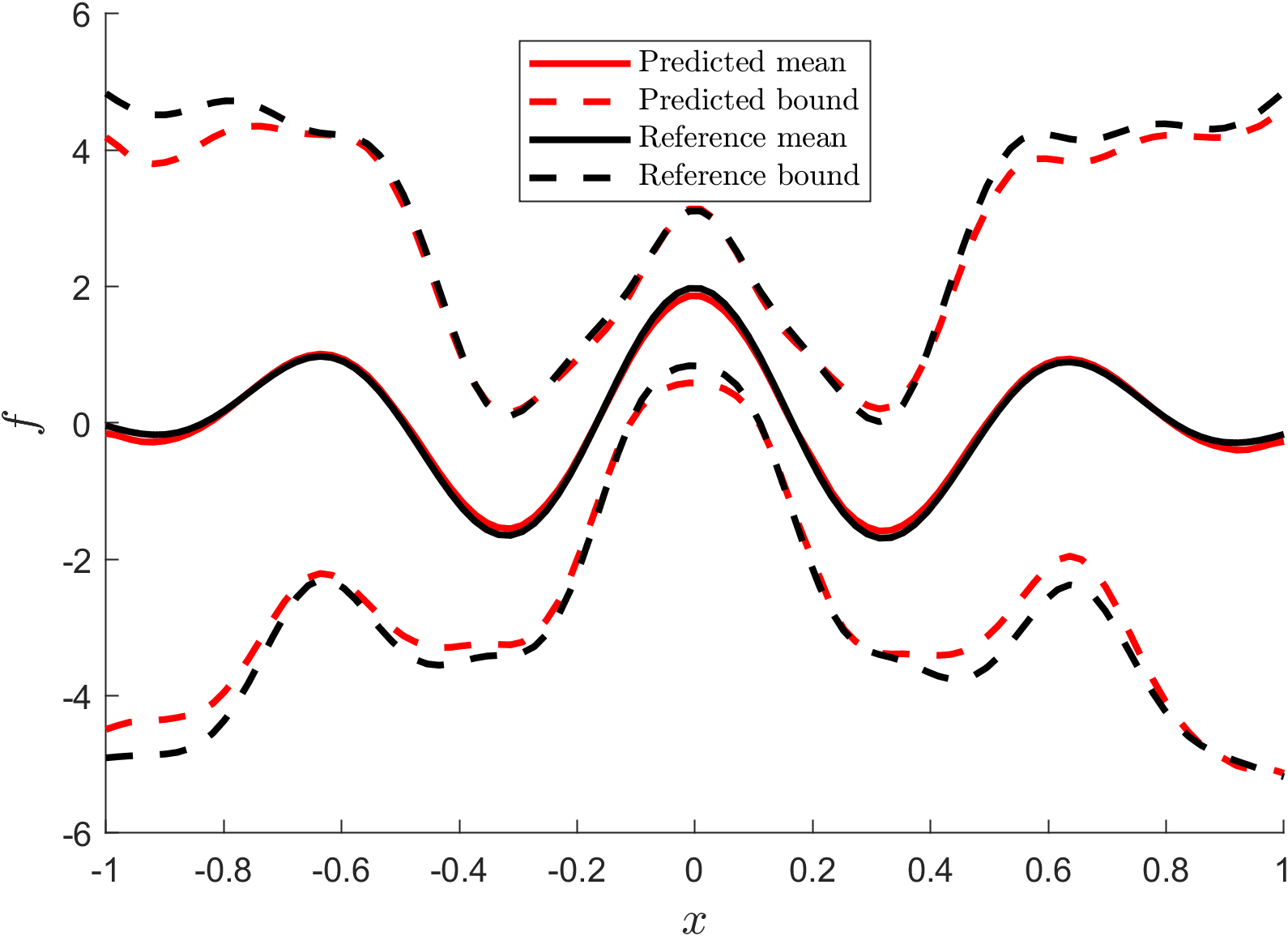}
    }
    \subfigure[]{
    \includegraphics[scale=.25]{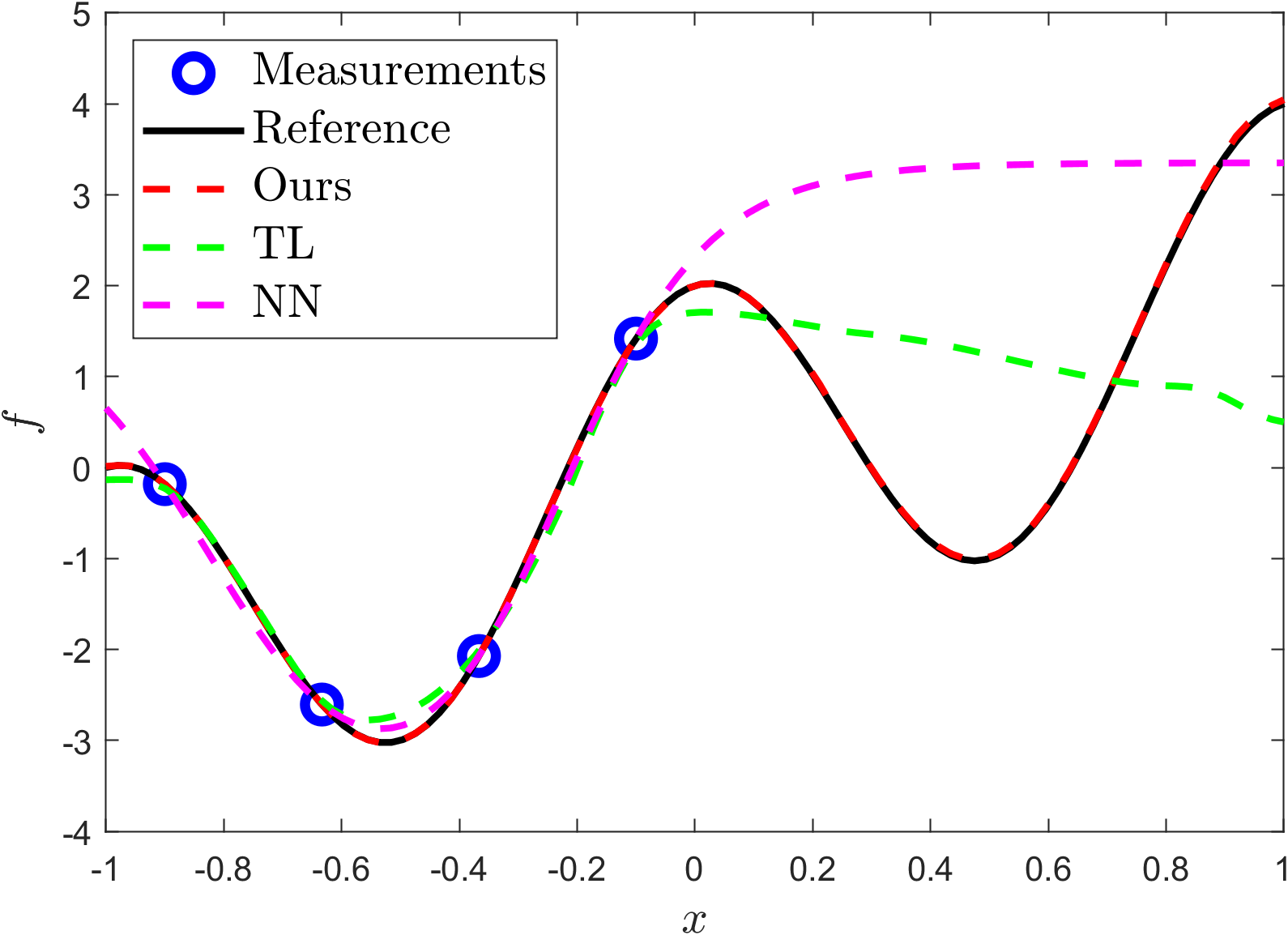}
    \includegraphics[scale=.25]{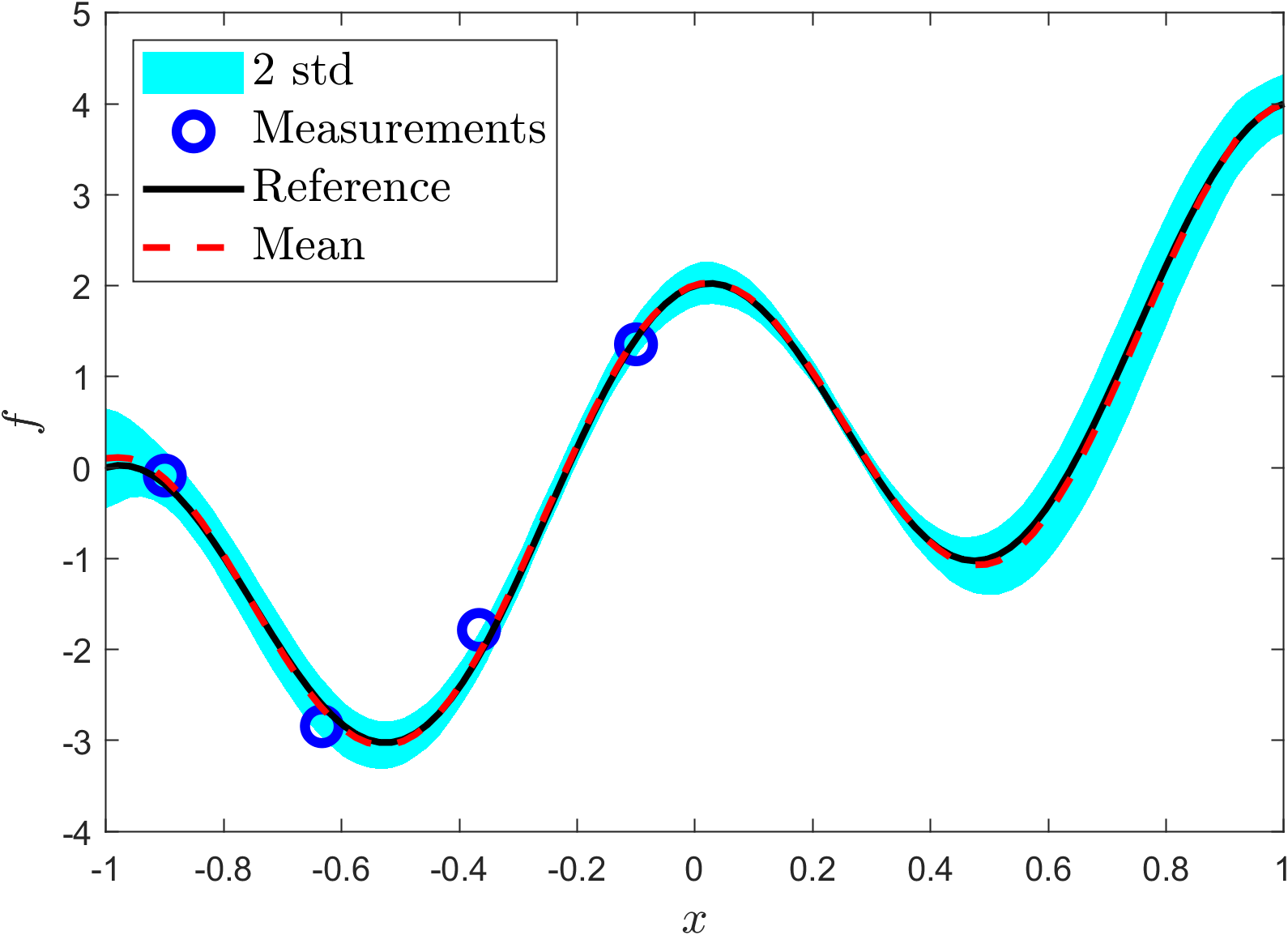}
    \includegraphics[scale=.25]{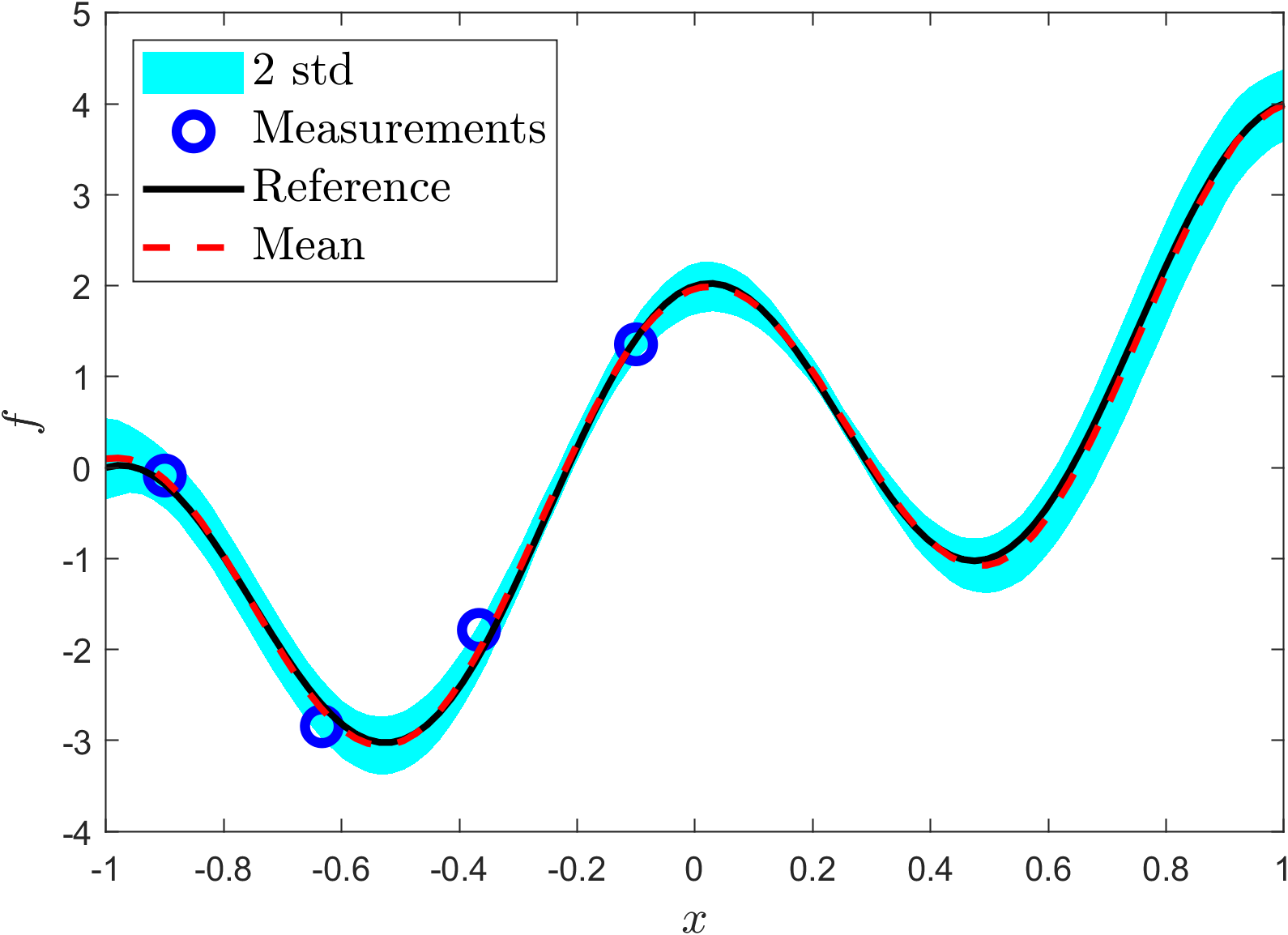}
    }
    \subfigure[]{
    \includegraphics[scale=.25]{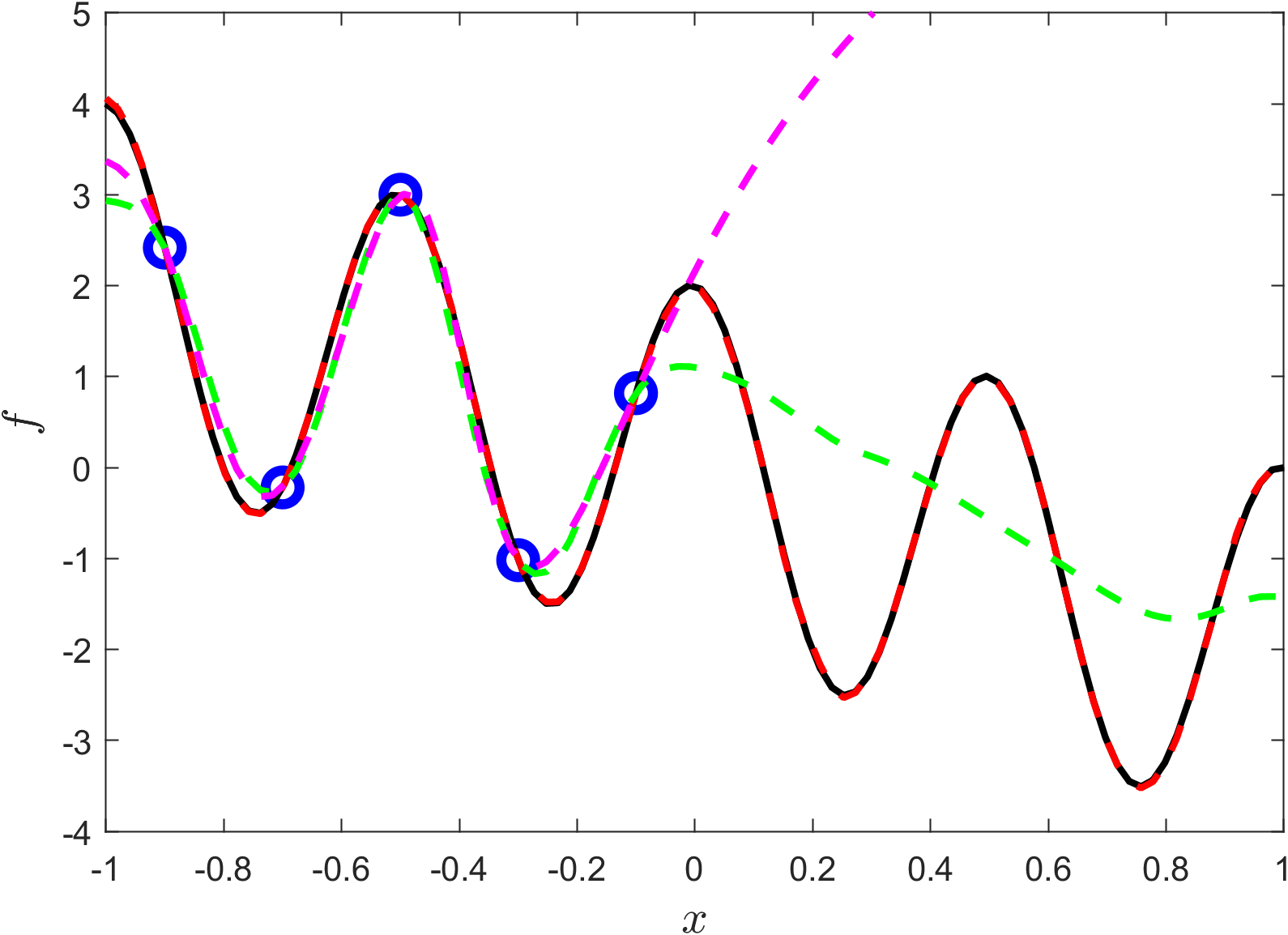}
    \includegraphics[scale=.25]{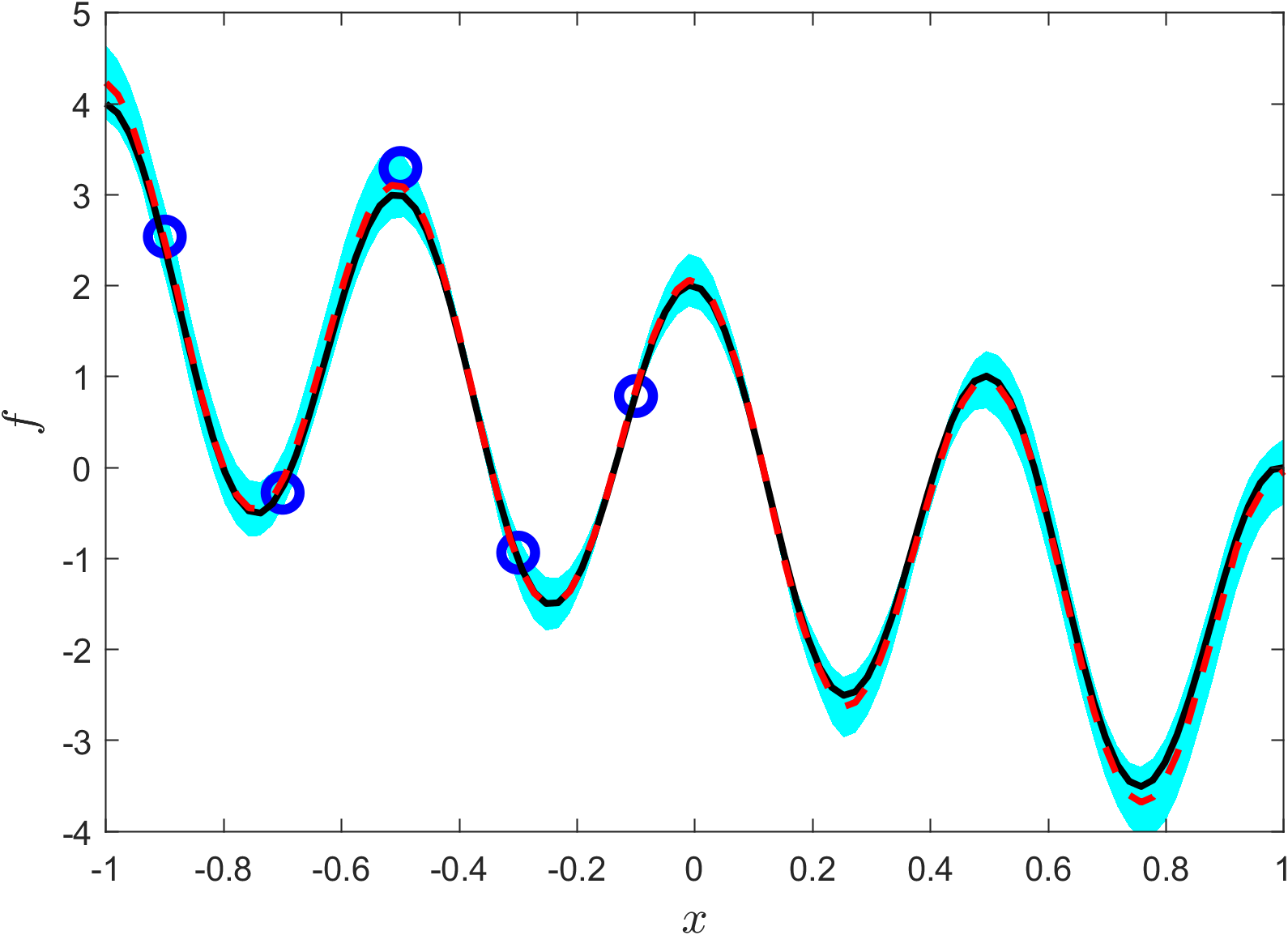}
    \includegraphics[scale=.25]{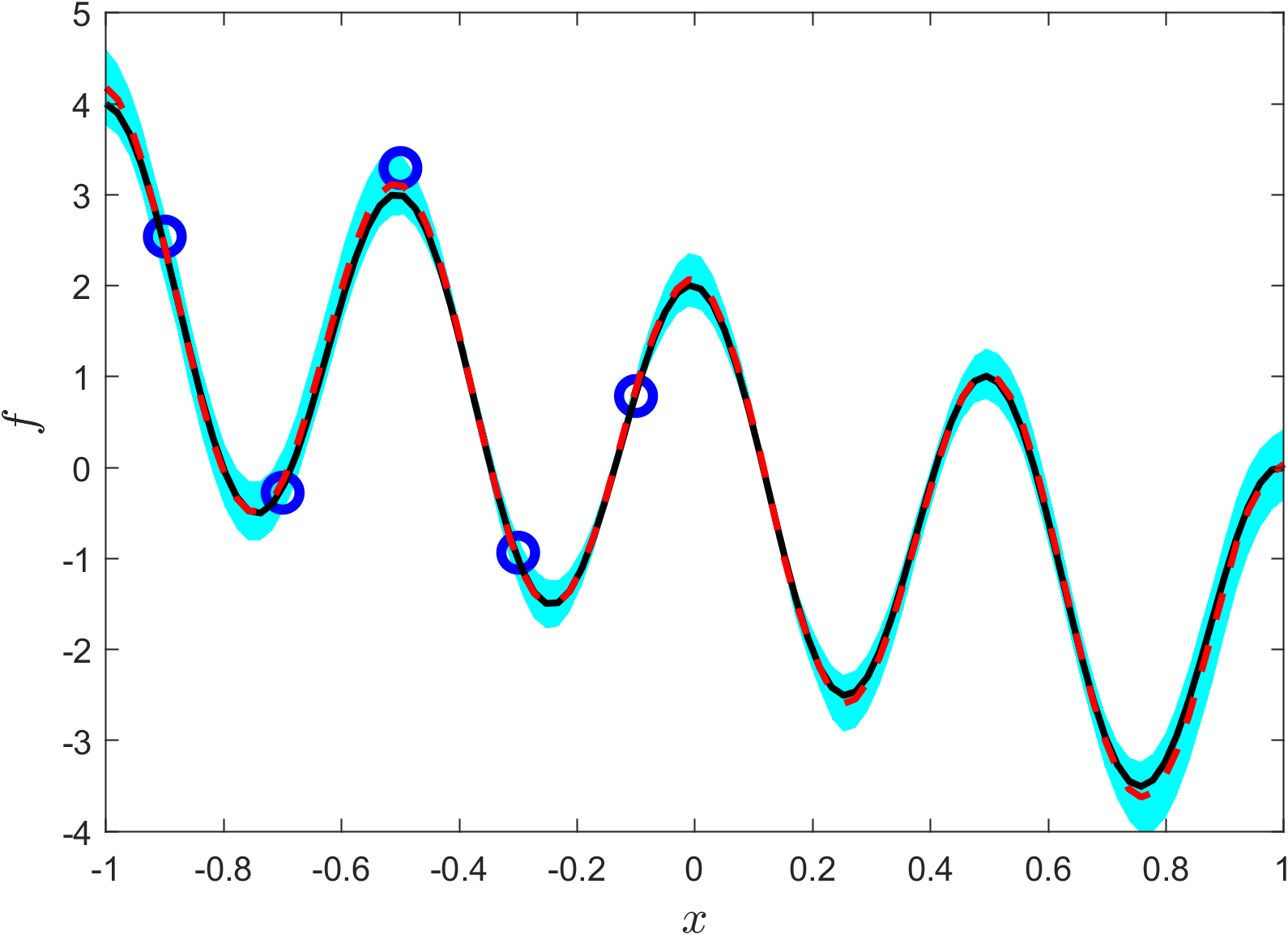}
    }
    \caption{Results for approximating the stochastic function defined in Eq.~\eqref{eq:example_1} and solving the downstream few-shot regression tasks. (a) Left: $1,000$ samples generated from the exact distribution; middle: $1,000$ samples generated from the learned generator; right: statistics computed from samples, in which we refer to the interval of mean $\pm$ 2 standard deviations as bound. (b)/(c) Results for the downstream tasks. Left: results for noiseless cases using our method, the transfer learning (TL) method in \cite{desai2021one}, and regular NN method; middle: results for noisy case using our method with HMC for posterior estimate; right: results for the same noisy case using our method with LA for posterior estimate.}
    \label{fig:example_1}
\end{figure}

For the training of MH-NNs and NFs, $1,000$ $f$ subject to Eq.~\eqref{eq:example_1} are sampled, each of which forms a regression task with $40$ measurements sampled equidistantly on $[-1, 1]$ as data. Samples of $f$ for training are displayed in Fig.~\ref{fig:example_1}(a). Both noiseless and noisy data cases are considered in the few-shot regression tasks. As described in Sec.~\ref{subsec:downstream}, the former is solved by fine-tuning the head using gradient descent, while the latter is solved by estimating the posterior distribution~(Eq.~\eqref{eq:posterior}) using HMC and LA. The noise $\varepsilon$ is assumed to be independent additive Gaussian noise with scale $0.2$, i.e., $\varepsilon\sim N(0, 0.2^2)$. In the downstream few-shot regression tasks we compare our method with two other approaches, the transfer learning (TL) method from \cite{desai2021one}, which only transfers the body, $\Phi$, and the regular NN method, in which no prior knowledge is employed and all parameters of NN are trained.

Results for approximating $f$ and solving the downstream tasks are presented in Fig.~\ref{fig:example_1}. As shown in Fig.~\ref{fig:example_1}(a), our method approximates the stochastic function $f$ well, demonstrating the capability of MH-NNs in generative modeling. In solving the downstream tasks with noiseless data, $L_2$ regularization is imposed in the TL method and regular NN method, to prevent over-fitting when only $4$ or $5$ measurements are available. As we can see from Figs.~\ref{fig:example_1}(b) and (c), our approach yields accurate predictions and performs significantly better than the other two in both tasks, particularly in the region where there are no measurements. By comparing our approach with the NN method, we can see that prior knowledge of $f$ is learned from $\{\T_k\}_{k=1}^M$ and transferred successfully to new downstream tasks. By comparing our approach with the TL method, we can see that the prior knowledge is stored in both the body and (the distribution of) the head. 
For the noisy cases, it is shown that, for both tasks and both posterior estimating methods, the predictions are accurate and trustworthy: the predicted means agree with the references and the errors are bounded by the predicted uncertainties. It is worth noting that the predicted uncertainties do not develop in the interval $[0, 1]$ and show periodic patterns, even if there are no measurements. That is because an informative prior, which is learned by MH-NNs and NFs, is imposed on the head in Bayesian inference.

\begin{figure}[ht]
    \centering
    \includegraphics[scale=.25]{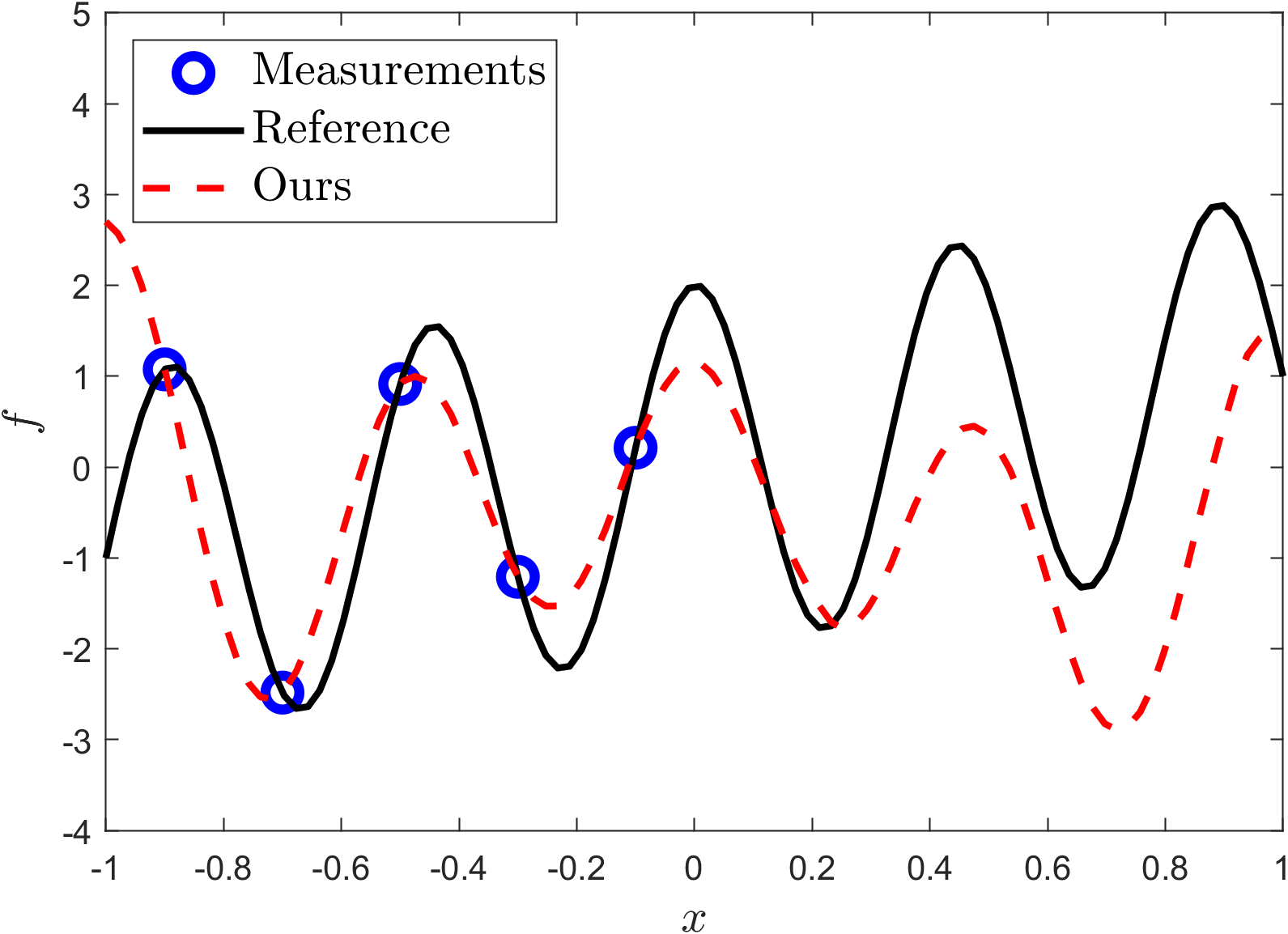}
    \includegraphics[scale=.25]{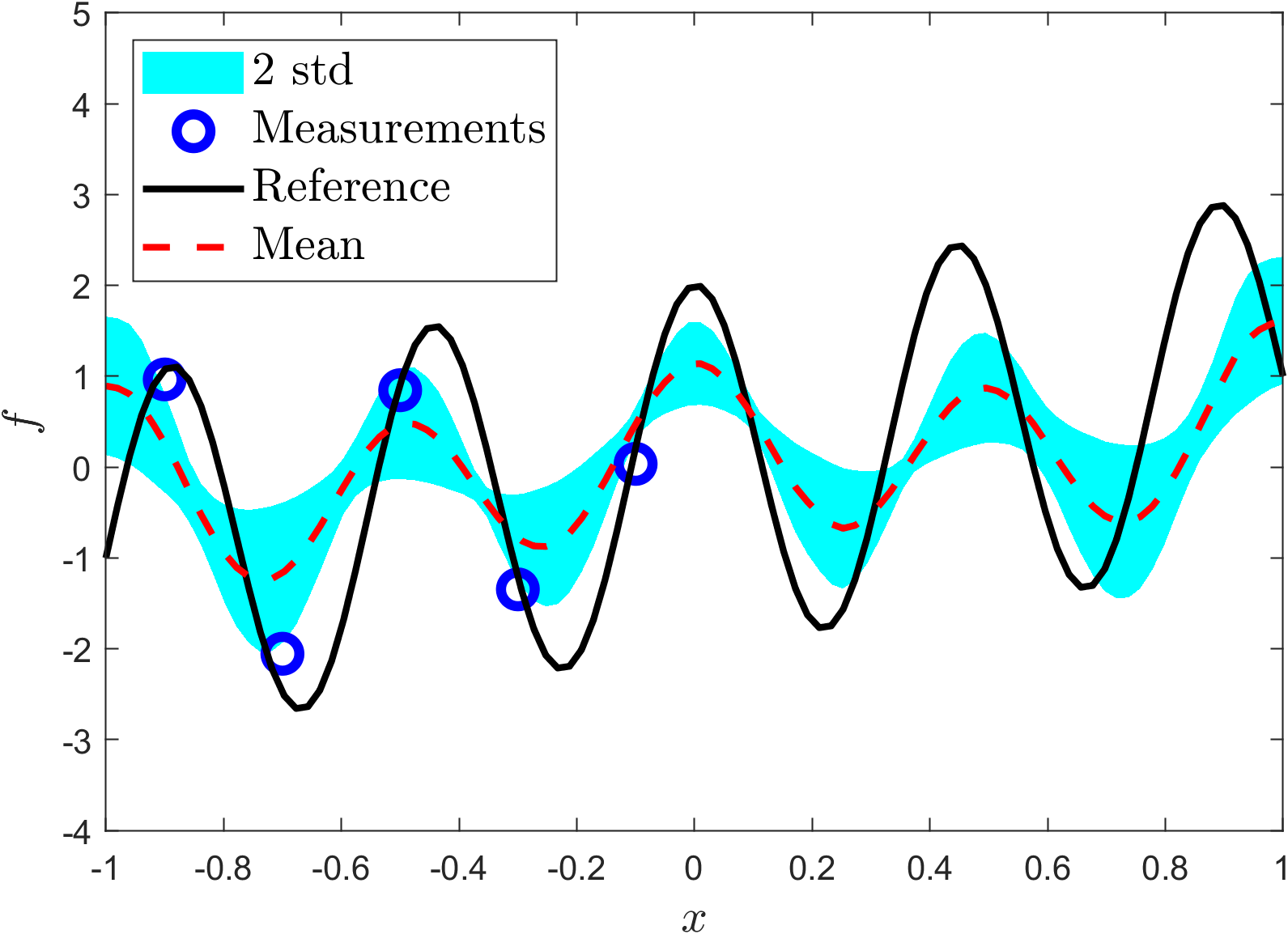}
    \includegraphics[scale=.25]{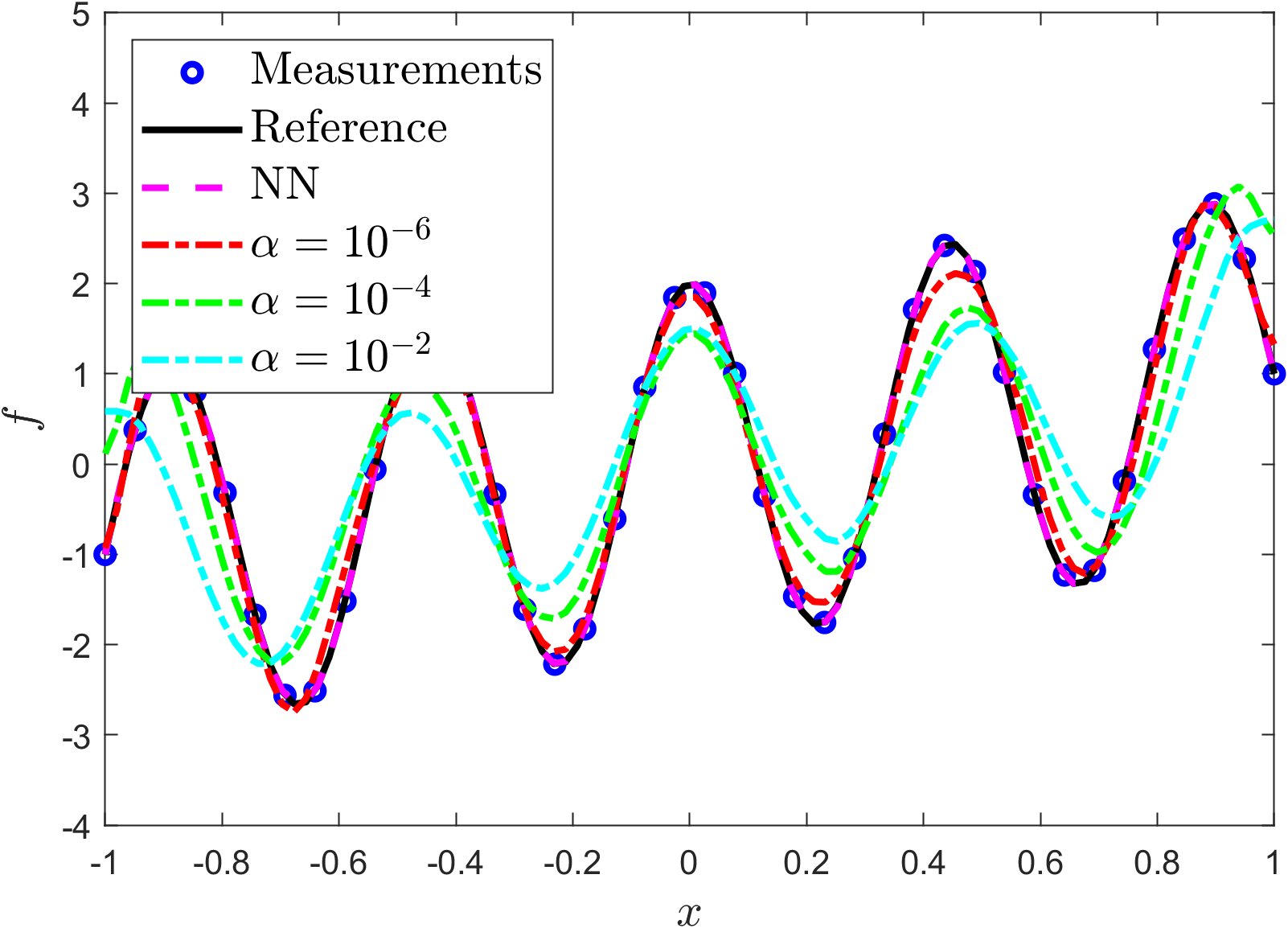}
    \caption{Results for regression on an out-of-distribution function. Left: few-shot regression with clean data using our approach; middle: few-shot regression with noisy data using our approach with HMC for posterior estimate; right: regression with sufficient clean data using regular NN method and our approach with different regularization terms, $\alpha$ in Eq.~\eqref{eq:optimization}.}
    \label{fig:example_1:ood}
\end{figure}

The target functions in downstream tasks considered previously are chosen to be in-distribution. They are regressed well with insufficient data, mainly because they belong to the space of functions, on which the generator is trained. However, when functions in the downstream tasks are out-of-distribution (OOD), our approach fails to produce good predictions, even if the data is sufficient, as shown in Fig.~\ref{fig:example_1:ood}. Here, the target function is chosen to be $2\cos(4.5\pi) + x$ with both $\omega$ and $\beta$ being OOD. Fluctuations are predicted but do not match the reference. In Fig.~\ref{fig:example_1:ood}, we can further see that when data is sufficient, a NN trained from scratch significantly outperforms our approach, showing that, for OOD functions, the more we rely on the learned regularization, which is indicated by the value of $\alpha$ in Eq.~\eqref{eq:optimization}, the more erroneous the prediction is.

\subsection{Nonlinear ODE system}
\label{subsec:example_2}
In this example, we consider the following ODE system \cite{lu2021learning}, which describes the motion of a pendulum with an external force:
\begin{equation}\label{eq:ode}
    \begin{aligned}
        \frac{du_1}{dt} &= u_2,\\
        \frac{du_2}{dt} &= -\lambda \sin(u_1) + f(t),
    \end{aligned}
\end{equation}
with initial condition $u_1(0) = u_2(0) = 0$. In Eq.~\eqref{eq:ode}, $f$ is the external force and $\lambda$ is a constant. 
Here, to demonstrate and study the capability of our method in generative modeling, we first consider the case where $f$ is a Gaussian process and $\lambda=1$ is known, which is referred to as the forward problem. Different from previous studies \cite{yang2020physics, zhong2022pi, guo2022normalizing}, in which a stochastic process is approximated directly by the output of NNs, in this example we place the differential operator right after NNs and approximate the stochastic process $f$ as the source term in Eq.~\eqref{eq:problem}. We also test our method on the inverse problem, where the values of $\lambda$ in Eq.~\eqref{eq:ode} are unknown in $\{\T_k\}_{k=1}^M$ and $\tilde{\T}$. The forward problem corresponds to Eq.~\eqref{eq:problem} with $\F_k, b_k$ being the same for all tasks and $u_k, f_k$ being task-specific, while the inverse problem corresponds to Eq.~\eqref{eq:problem} with $b_k$ being the same and $u_k, f_k$, and the differential operator $F_k$ being different as a consequence of task-specific $\lambda$.

\subsubsection{Forward problem}
We first assume $\lambda=1$ in Eq.~\eqref{eq:ode} is known, and the data on $f$ is available, i.e., $\D_k = \{(x_k^i, f_k^i)\}_{i=1}^{N_k}$, $k=1,...,M$. As described before, we employ MH-PINNs to solve $\{\T_k\}_{k=1}^M$ all at once and then employ NFs to learn the distribution of the head. Consequently, we obtain generators of $f$ and $u$. In this case, $f$ is assumed to be a Gaussian process with squared kernel function:
\begin{equation}\label{eq:gp}
\begin{aligned}
     f(t) \sim GP(0, K), t \in [0, 1], K(x, x^\prime) = \exp(-\frac{|x-x^\prime|^2}{2l^2}),
\end{aligned}
\end{equation}
where the correlation length $l$ is set to $0.1, 0.25, 0.5$. As discussed in Sec.~\ref{subsec:nf}, many types of NFs have been developed in the past decade for generative modeling and density estimate. MH-PINNs, when used as generators, are compatible with all NFs. In this regard, we compare three popular NFs, RealNVP \cite{dinh2016density}, MAF \cite{papamakarios2017masked} and IAF \cite{kingma2016improved}, and eventually compare MH-PINNs (with NFs) against PI-GAN \cite{yang2020physics} in approximating the Gaussian process defined in Eq.~\eqref{eq:gp} with different correlation lengths.

For the training of MH-PINNs and NFs as well as PI-GANs, $2,000$ $f$ are sampled with respect to Eq.~\eqref{eq:gp}, each of which forms a physics-informed regression task with $65$ measurements of $f$ equidistantly sampled on $[0, 1]$ as data. Notice that the ODE system in Eq.~\eqref{eq:ode} can be rewritten in a simpler format as follows:
\begin{equation}\label{eq:ode2}
\begin{aligned}
    u_{tt} = -\lambda \sin(u) + f(t), t\in[0, 1],
\end{aligned}
\end{equation}
with initial conditions $u(0) = u_t(0) = 0$. Hence, we choose to use Eq.~\eqref{eq:ode2} to build the loss function for physics-informed learning. 

\begin{figure}[ht]
    \centering
    \includegraphics[scale=.25]{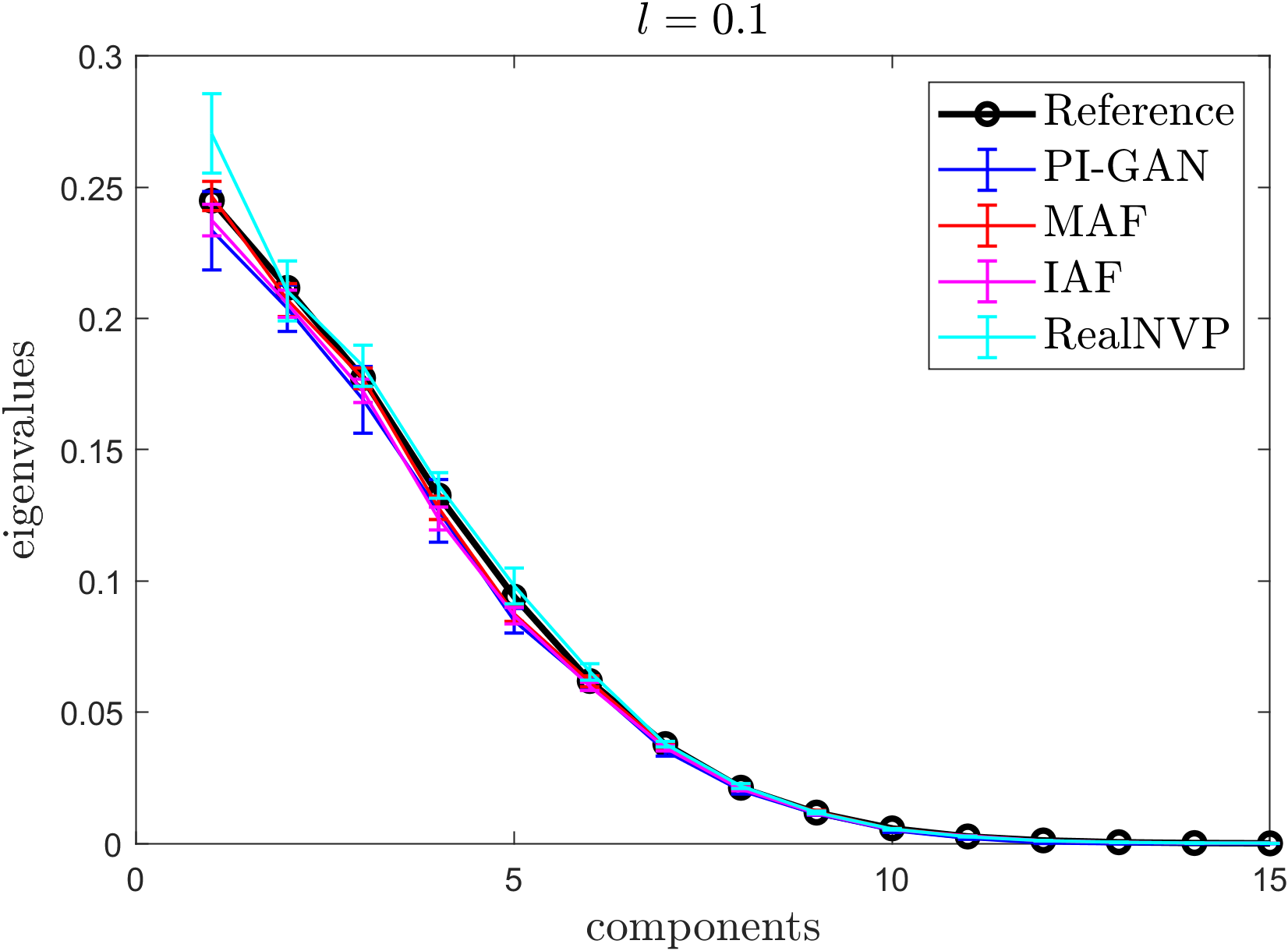}
    \includegraphics[scale=.25]{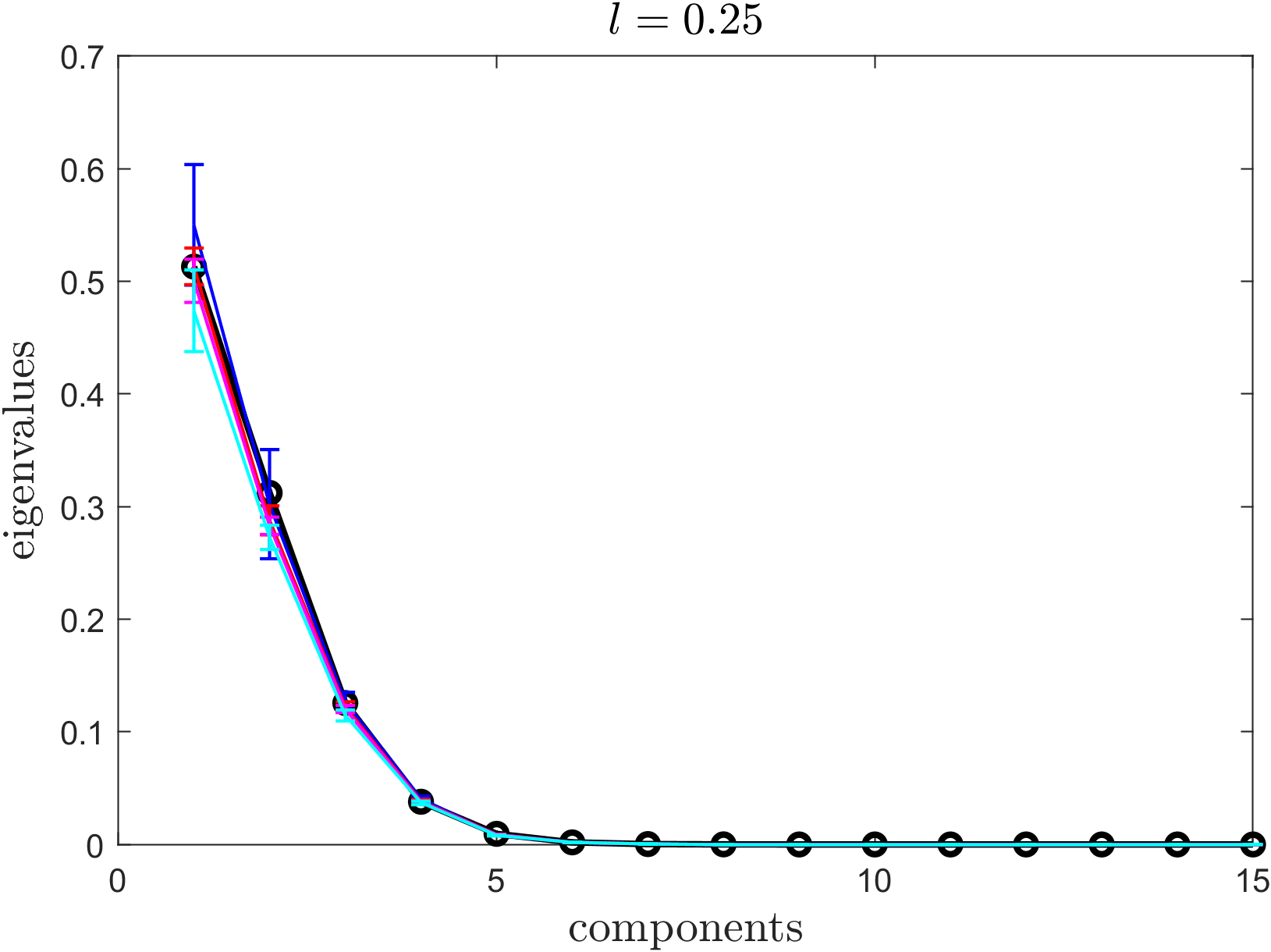}
    \includegraphics[scale=.25]{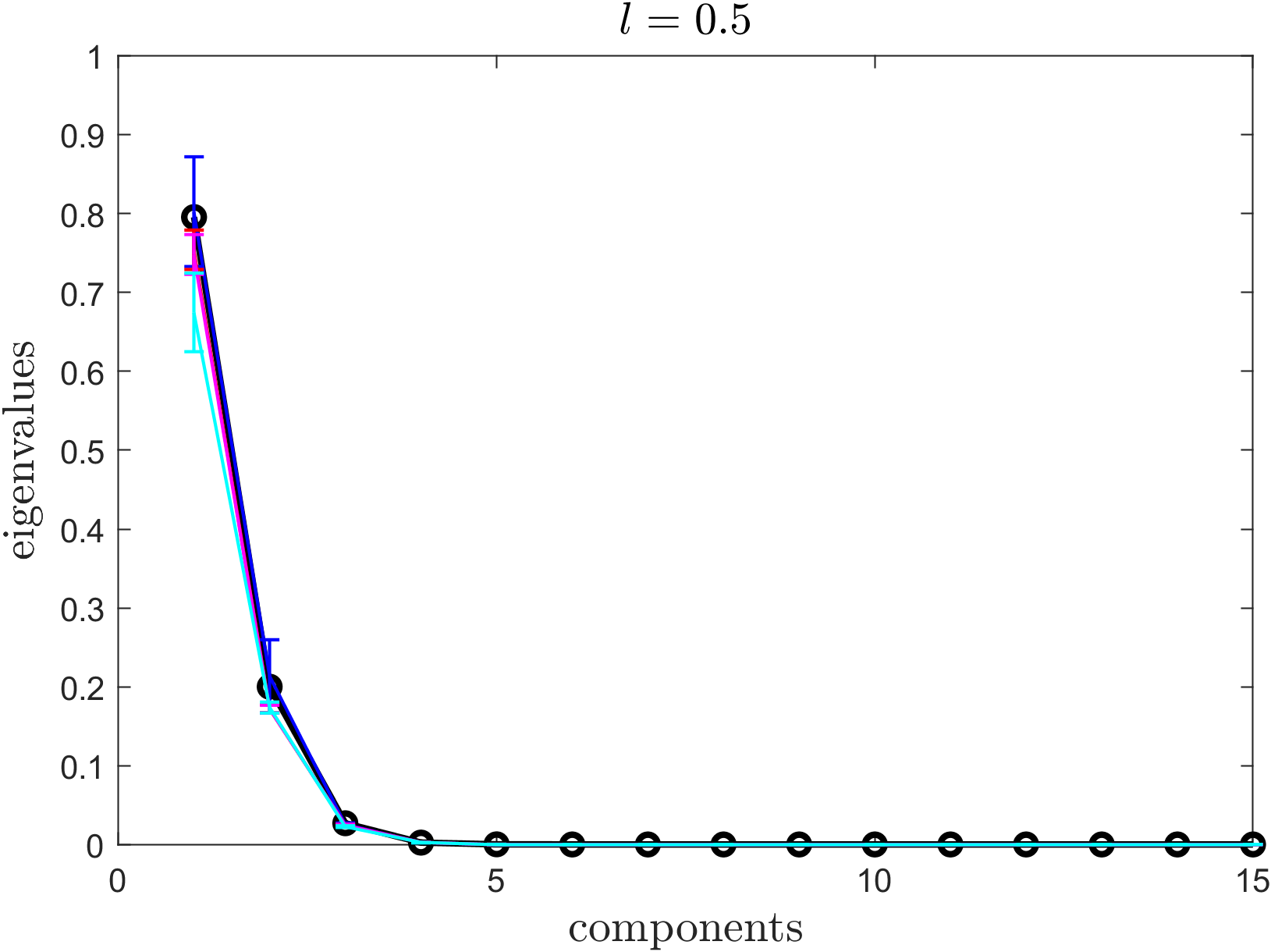}
    \caption{Approximating Gaussian processes as the source term in Eq.~\eqref{eq:ode} using different models: spectra of the correlation structure for the learned generators, for different correlation lengths, $l$. The covariance matrix is constructed using $10,000$ generated samples, and eigen-values are averaged over 10 generators trained independently.}
    \label{fig:example_2:gp}
\end{figure}

Results for comparisons are shown in Fig.~\ref{fig:example_2:gp} and Table~\ref{tab:training_cost}. From the spectral analysis of the approximated Gaussian processes shown in Fig.~\ref{fig:example_2:gp}, we can see that MH-PINNs with MAF and IAF are comparable with PI-GANs while MH-PINNs with RealNVP fall marginally behind. As shown in Table~\ref{tab:training_cost}, the computational costs of MH-PINNs with MAF and RealNVP are significantly lower than PI-GANs, while MH-PINNs with IAF is more expensive than PI-GANs. We note that in this example we also record the computational cost for sampling using different generators. As shown in Table~\ref{tab:training_cost}, PI-GANs are significantly faster in generating samples. That is because, generally, GANs require relatively shallow NNs as opposite to NFs, for which a deep architecture is needed to keep up the expressivity. 
Among three NFs, IAF is the fastest in sampling while MAF is the lowest, as opposed to training, which is consistent with the properties of those two NFs: MAF is slow for the forward pass, which is used to generate samples, and fast for the inverse pass, which is used to compute the density, while IAF is the opposite. Despite the fact that MAF is slow in sampling, considering its fast training and good performance, we equip MH-PINNs with MAF as the density estimator and the generator for all other examples in this paper.

\begin{table}[ht]
    \footnotesize
    \centering
    \begin{tabular}{|c|c|c|c|c|}
    \hline
         & MAF & IAF & RealNVP & PI-GAN\\
         \hline
       Phase 1 & $134$s & $134$s & $134$s & N/A\\
       \hline
       Phase 2 & $252$s & $3939$s & $245$s & N/A\\
       \hline
       Total & $386$s & $4073$s & $379$s & $3243$s \\
       \hline
       Sampling & $1.98\times 10^{-1}$s & $1.48\times 10^{-2}$s & $1.50\times 10^{-2}$s & $2.29\times 10^{-3}$s \\
       \hline
    \end{tabular}
    \caption{Computational time for different models to approximate Gaussian process with correlation length $l=0.1$. The MH-PINN method is a two-step method and hence its computation is decomposed into two parts: training MH-PINNs referred to as phase 1 and training NFs referred to as phase 2. Sampling time is defined to be the average time needed to generate $10,000$ samples of $u$.}
    \label{tab:training_cost}
\end{table}

\subsubsection{Inverse problem}
Next, we assume $\lambda$ in Eq.~\eqref{eq:ode} is unknown, and some measurements of $u$ are available, in addition to $f$, i.e., $\D_k = \{\{x_k^i, f_k^i\}_{i=1}^{N_k^f}, \{x_k^i, u_k^i\}_{i=1}^{N_k^u}\}$. MH-PINNs are first employed to infer $u_k$ as well as $\lambda_k$ from data $\D_k$ and physics, and NFs are employed afterwards to learn from samples of $H_k$ and $\lambda_k$. To this end, the generative model is for the joint distribution of $u, f$ and $\lambda$. Here, we assume $f$ follows a truncated Karhuen-Loeve (KL)-expansion, with $5$ leading terms, of the Gaussian process with squared kernel function and correlation length $0.1$, and for each task $\T_k$, $\lambda_k=\frac{1}{2}\exp(\int_{[0, 1]} f^2_k(t) dt)$. As for the downstream task $\tilde{\T}$, the target is to infer $u$ and $\lambda$ from insufficient data of $u$ and $f$.  

\begin{figure}[ht]
    \centering
    \subfigure[]{
        \includegraphics[scale=.25]{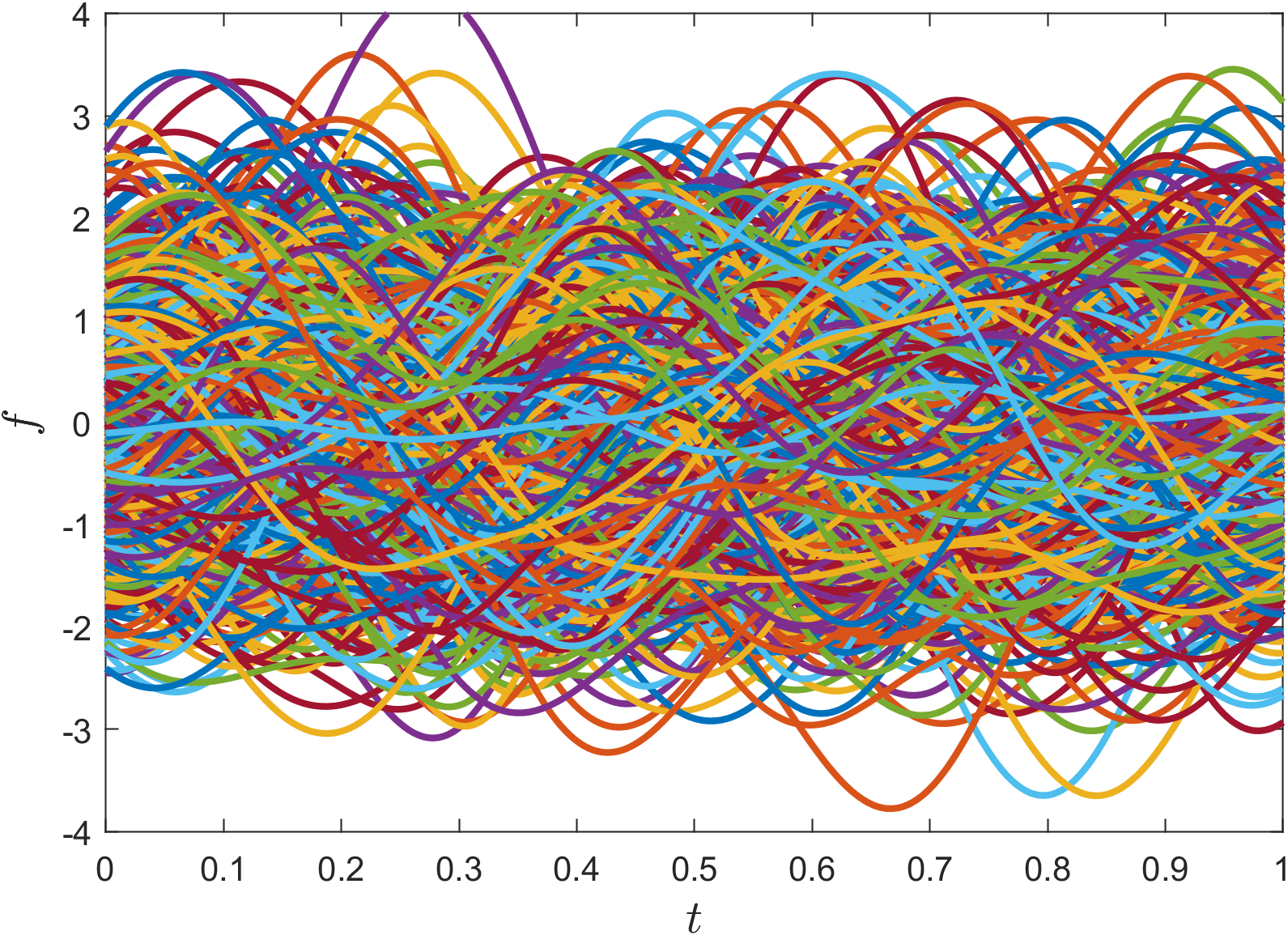}
        \includegraphics[scale=.25]{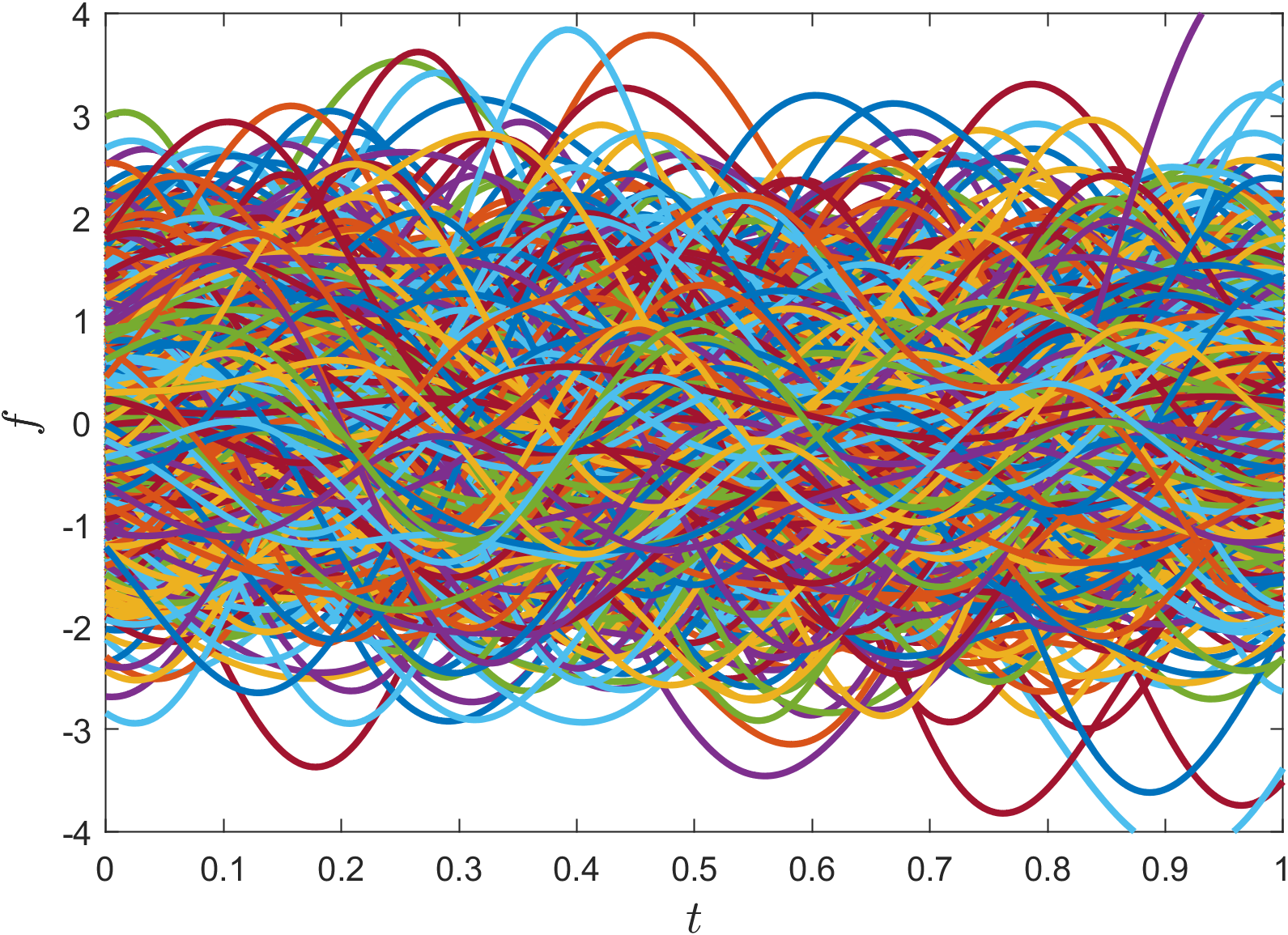}
        \includegraphics[scale=.25]{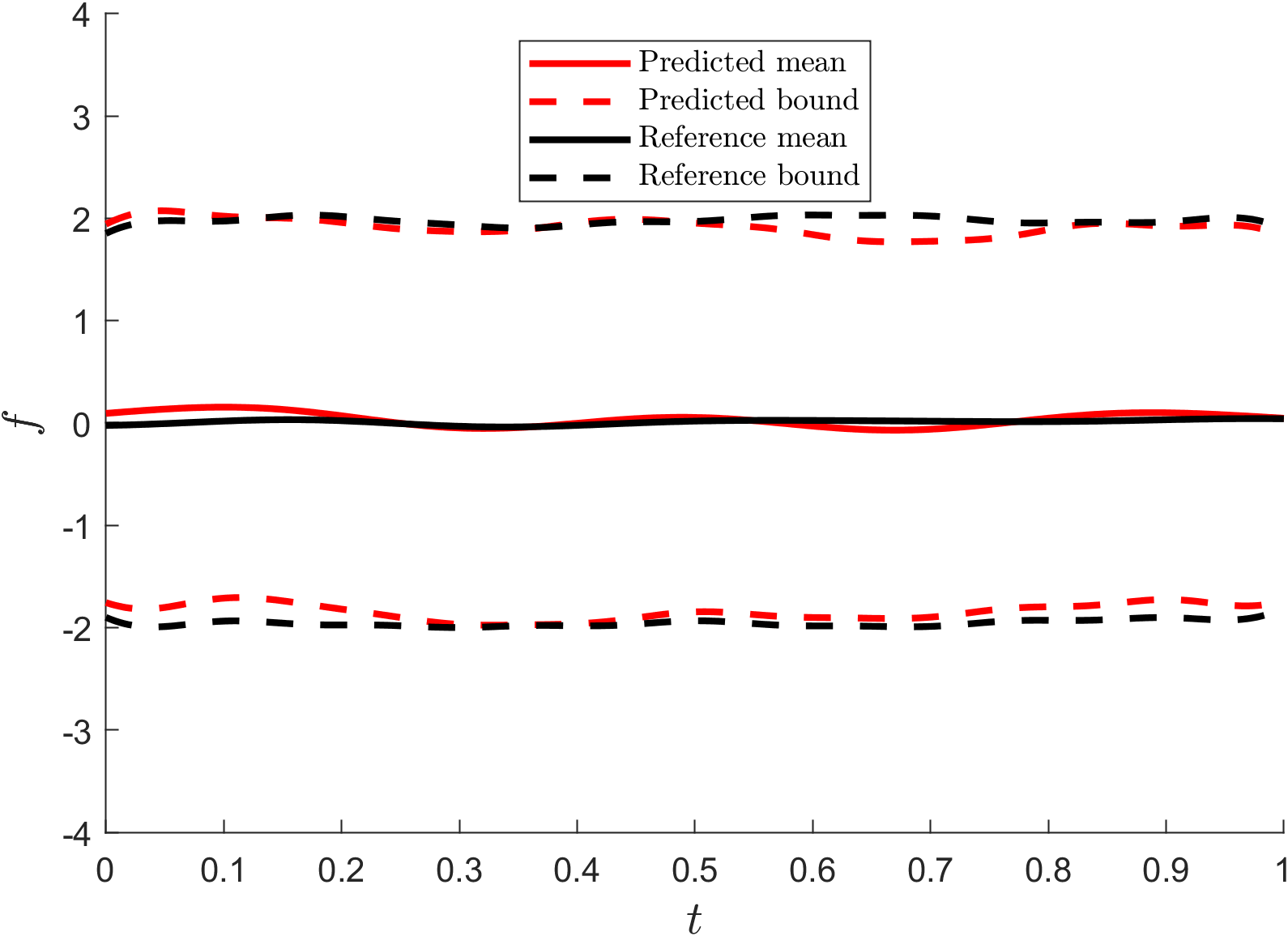}
    }
    \subfigure[]{
        \includegraphics[scale=.25]{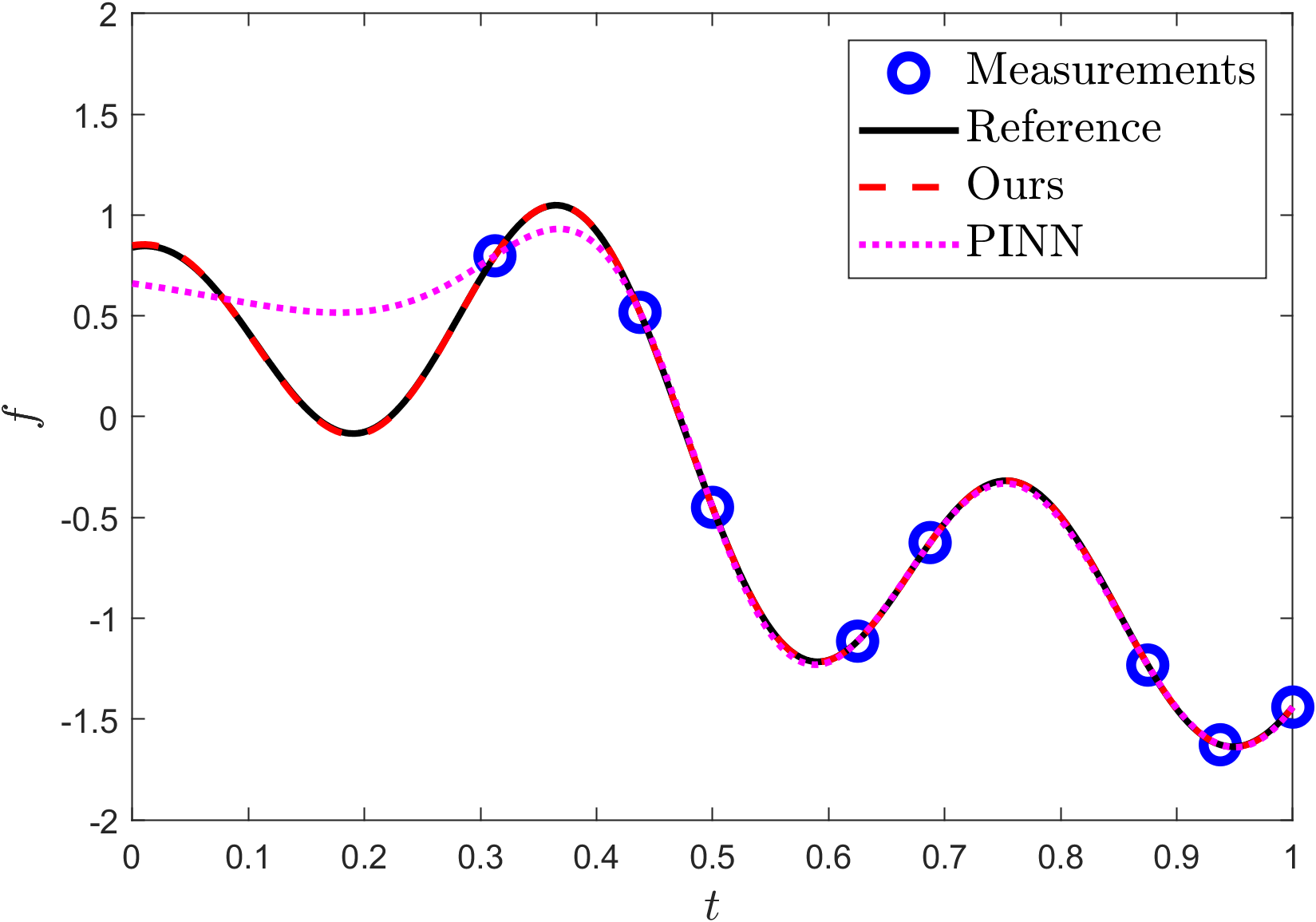}
        \includegraphics[scale=.25]{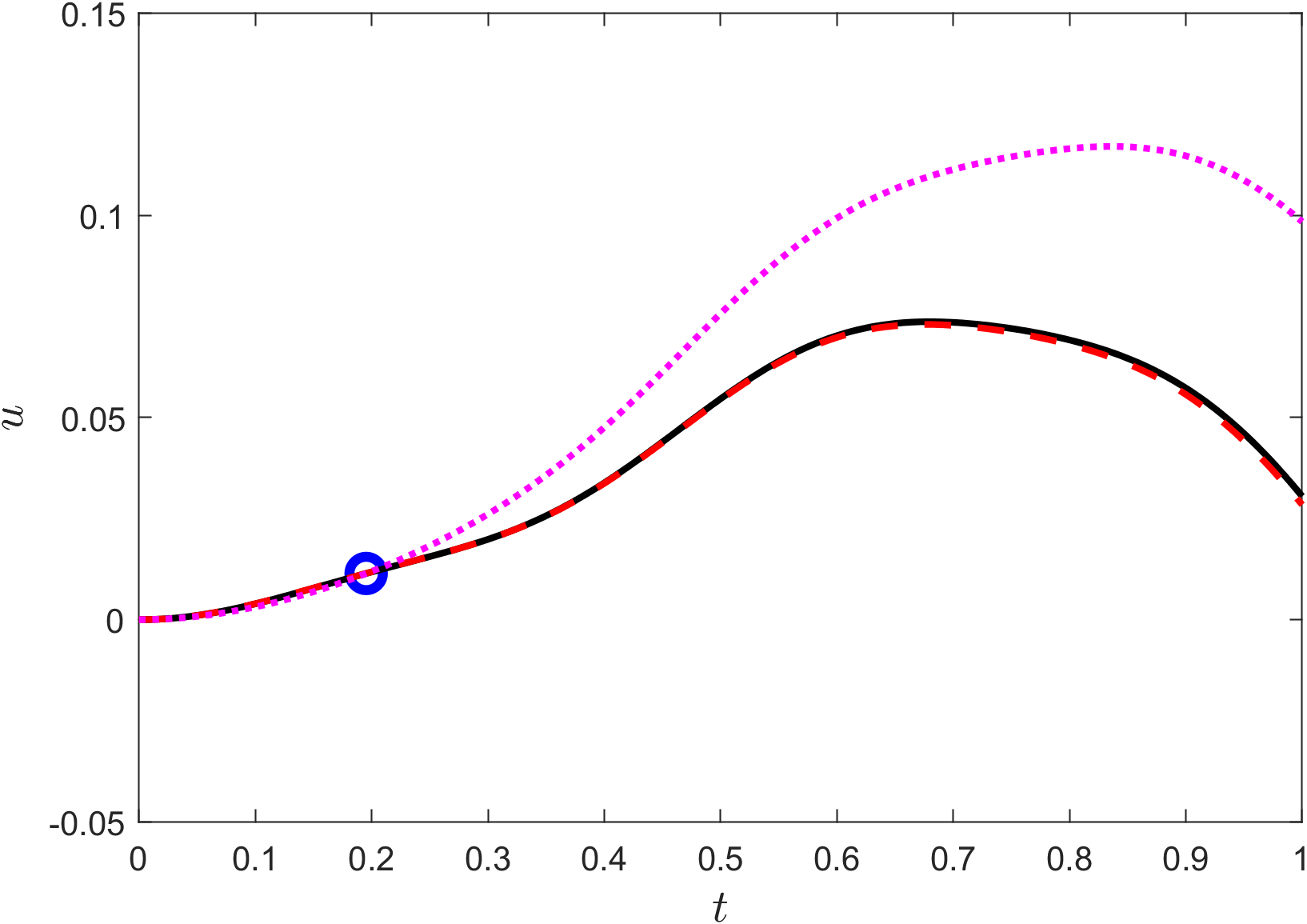}
    }
    \subfigure[]{
        \includegraphics[scale=.25]{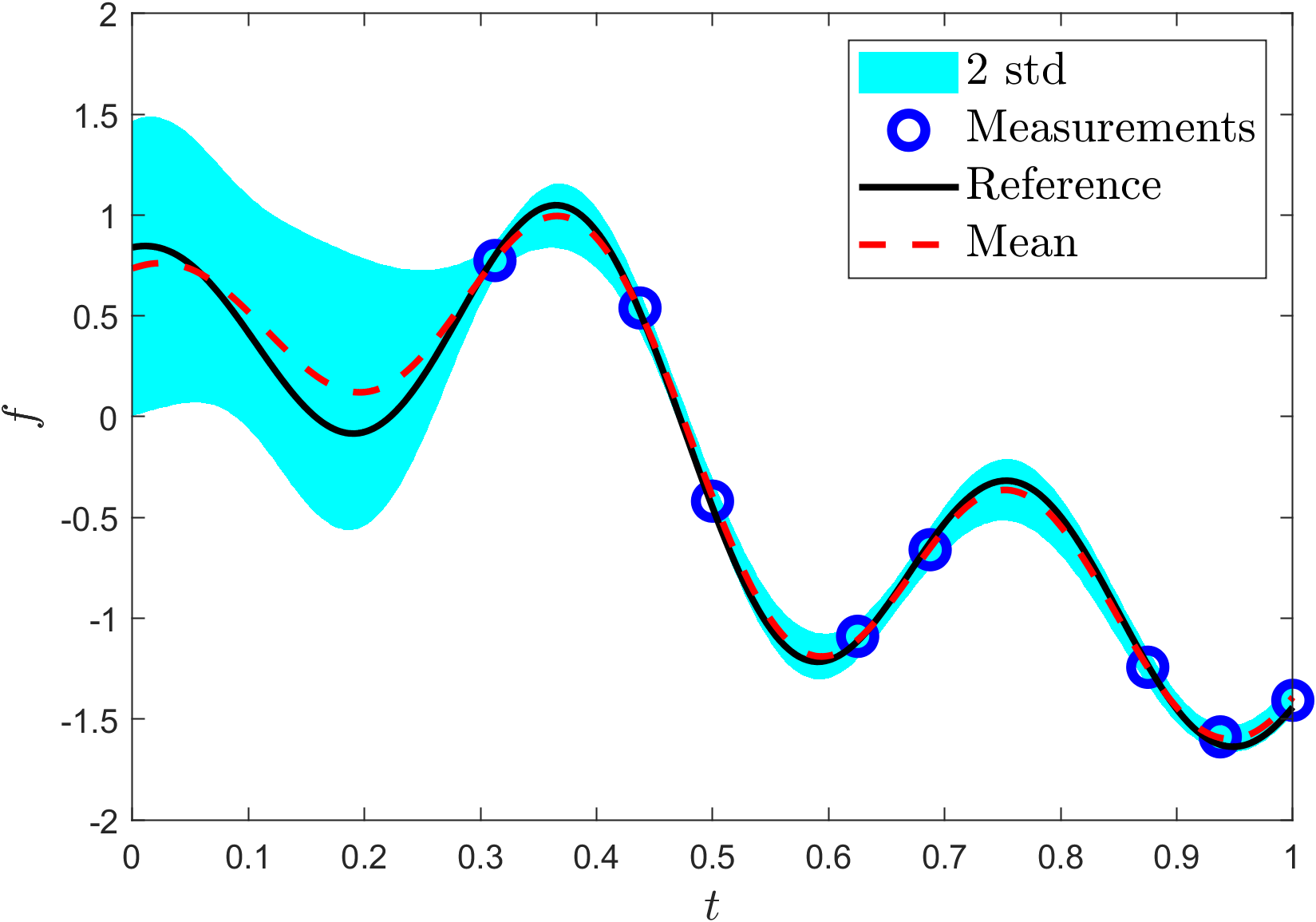}
        \includegraphics[scale=.25]{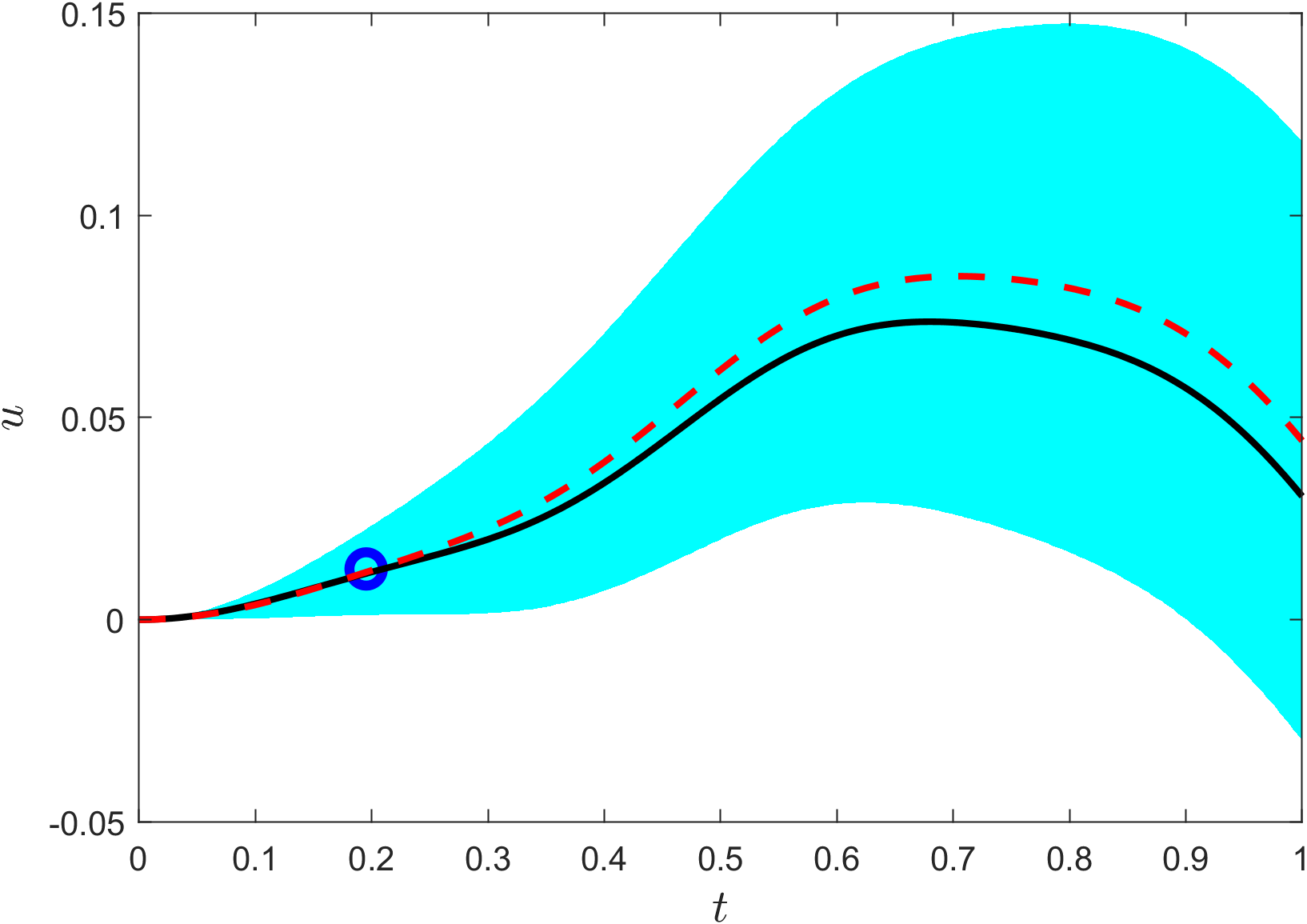}
        \includegraphics[scale=.25]{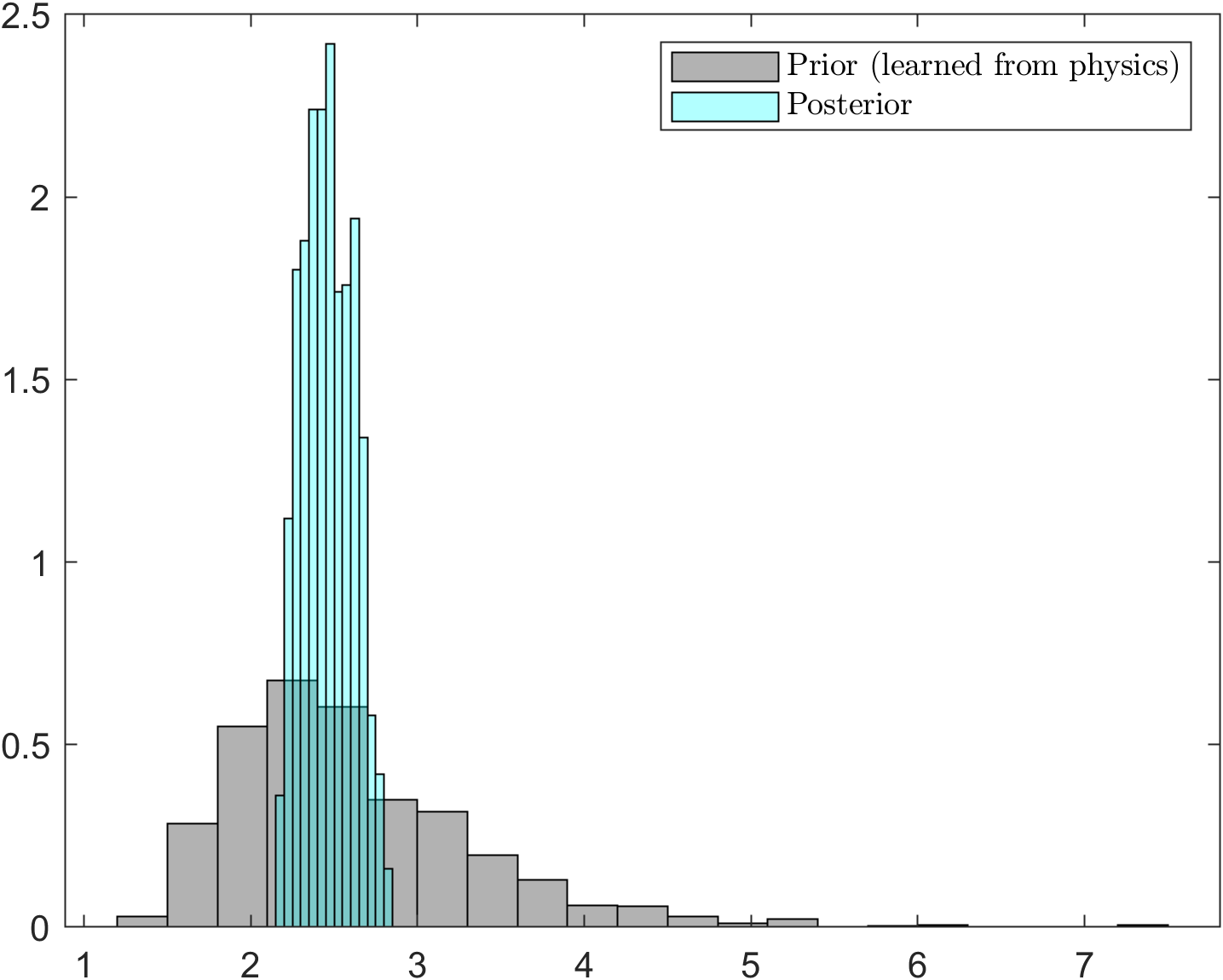}
    }
    \caption{Results for the inverse problem of the ODE system \eqref{eq:ode}, with initial conditions hard-encoded in NN modeling. (a) Left: $1,000$ samples of $f$, generated from the exact distribution; middle: $1,000$ samples of $f$, generated from the learned generator; right: statistics computed from samples. The bound is defined as the same as in the caption of Fig.~\ref{fig:example_1}.  (b) Predicted $f$ and $u$ using PINNs and our approach, for the downstream inverse problem with noiseless data. (c) Predicted $f$, $u$ and $\lambda$ with uncertainties using our approach, for the downstream inverse problem with noisy data. The predicted mean and standard deviation of $\lambda$ is $2.4663$ and $0.1501$, while the reference value is $2.3609$.}
    \label{fig:example_2:inverse}
\end{figure}

For the training of MH-PINNs and NFs, $2,000$ samples of $f$ are generated and displayed in Fig.~\ref{fig:example_2:inverse}(a). For each task, we assume $33$ measurements of $f_k$ and $9$ measurements of $u_k$, equidistantly distributed on $[0, 1]$, are available, and initial conditions are hard-encoded in NN modeling. For the downstream task, we assume $1$ random measurement of $u$ and $8$ random measurements of $f$ are available with hard-encoded initial conditions as well. For the case with noisy measurements, we assume the noises $\varepsilon_f$ and $\varepsilon_u$ to be additive Gaussian, with $0.05$ noise scale for measurements of $f$ and $0.005$ noise scale for measurements of $u$, respectively, i.e. $\varepsilon_f\sim N(0, 0.05^2)$ and $\varepsilon_u\sim N(0, 0.005^2)$. The reference solution as well as the clean data of $u_k$ are generated by solving Eq.~\eqref{eq:ode} for each task $\T_k$, with corresponding $f_k$ and $\lambda_k$ using Matlab \textit{ode45}.

Results are shown in Fig.~\ref{fig:example_2:inverse} and Table~\ref{tab:inverse}, from which we can see our method is able to approximate the stochastic process as a source term well and produce accurate and trustworthy predictions, for $u, f$ and also $\lambda$ in the downstream task with limited data, in both noiseless and noisy cases. As shown, the PINN method yields unacceptable estimate over both $u$ and $\lambda$ due to lack of data, while our approach is of much higher accuracy by integrating prior knowledge from $\{\T_k\}_{k=1}^M$ with MH-PINNs.

\begin{table}[ht]
    \footnotesize
    \centering
    \begin{tabular}{|c|c|c|}
    \hline
         & PINN & MH-PINN \\
         \hline
       $\lambda$ & $0.8440$ & $2.5428$\\
       \hline
       Error (\%) & $63.99$ & $1.21$ \\
       \hline
    \end{tabular}
    \caption{Estimate of $\lambda$ and $L_2$ relative error of $u$ for the downstream inverse problem on Eq.~\eqref{eq:ode} with clean data, using our approach and the regular PINN method. The reference value for $\lambda$ is $2.3609$.}
    \label{tab:inverse}
\end{table}
\subsection{1-D nonlinear reaction-diffusion equation}
\label{subsec:example_3}
We now test our method on a 1-D nonlinear time-dependent reaction-diffusion equation, which is commonly referred to as Fisher's equation \cite{ablowitz1979explicit}:
\begin{align}\label{eq:time_dependent}
    &u_t = Du_{xx} + k u(1-u), t \in [0, 1], x\in[-1, 1],\\
    &u(t, -1) = u(t, 1) = 0, t \in [0, 1],\\
    &u(0, x) = u_0(x), x\in[-1, 1],
\end{align}
where $D=0.1, k=0.1$ and $u_0(x)$ is the initial condition function. In this example, we assume that the initial condition function is a stochastic process with the following distribution:
\begin{equation}
    \begin{aligned}
        u_0(x) = \frac{(x^2 - 1)}{5}\sum_{j=1}^5\xi_j(\cos^2(j x) - 1), x\in[-1, 1], 
    \end{aligned}
\end{equation}
where $\xi_j, j=1, ..., 5$ are independent and identically distributed (i.i.d.) random variables subject to uniform distribution on $[0, 1)$, i.e., $\xi_j \sim U[0, 1)$. Unlike previous examples, the stochasticity comes from the initial condition rather than the source term.
This example corresponds to Eq.~\eqref{eq:problem} with $\F_k$ and $f_k$ being the same for all tasks, and $u_k$ and $b_k$ being task-specific. In addition to measurements of $u_0$, points on which the PDE residuals are computed are also required in both $\{\T_k\}_{k=1}^M$ and $\tilde{\T}$. Hence, the data is $\D_k=\{\{(t_k^i, x_k^i), 0\}_{i=1}^{N_k^f}, \{(0, x_k^i),b_k^i\}_{i=1}^{N_k^b}\}.$

\begin{figure}[ht]
    \centering
    \subfigure[]{
        \includegraphics[scale=.25]{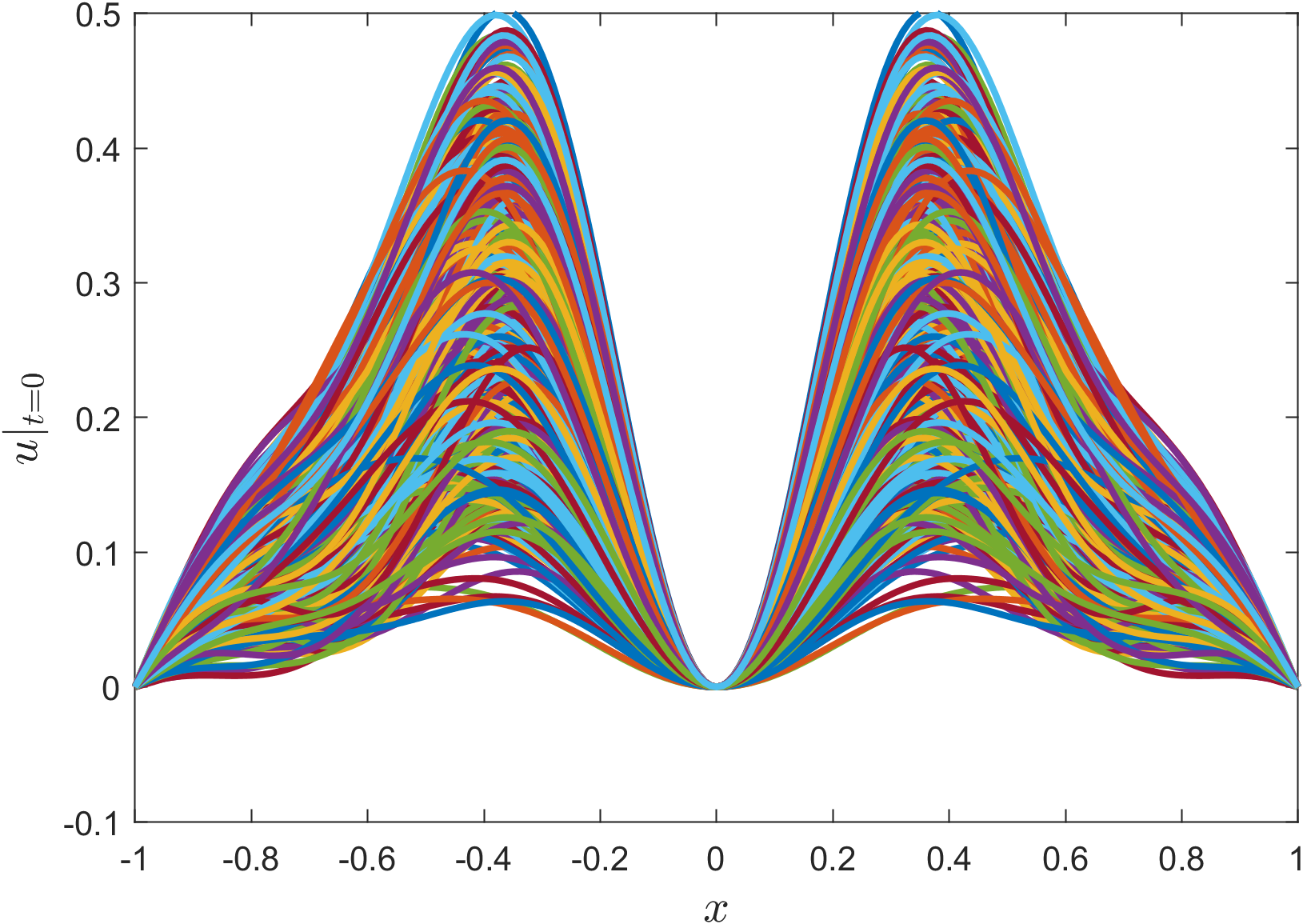}
        \includegraphics[scale=.25]{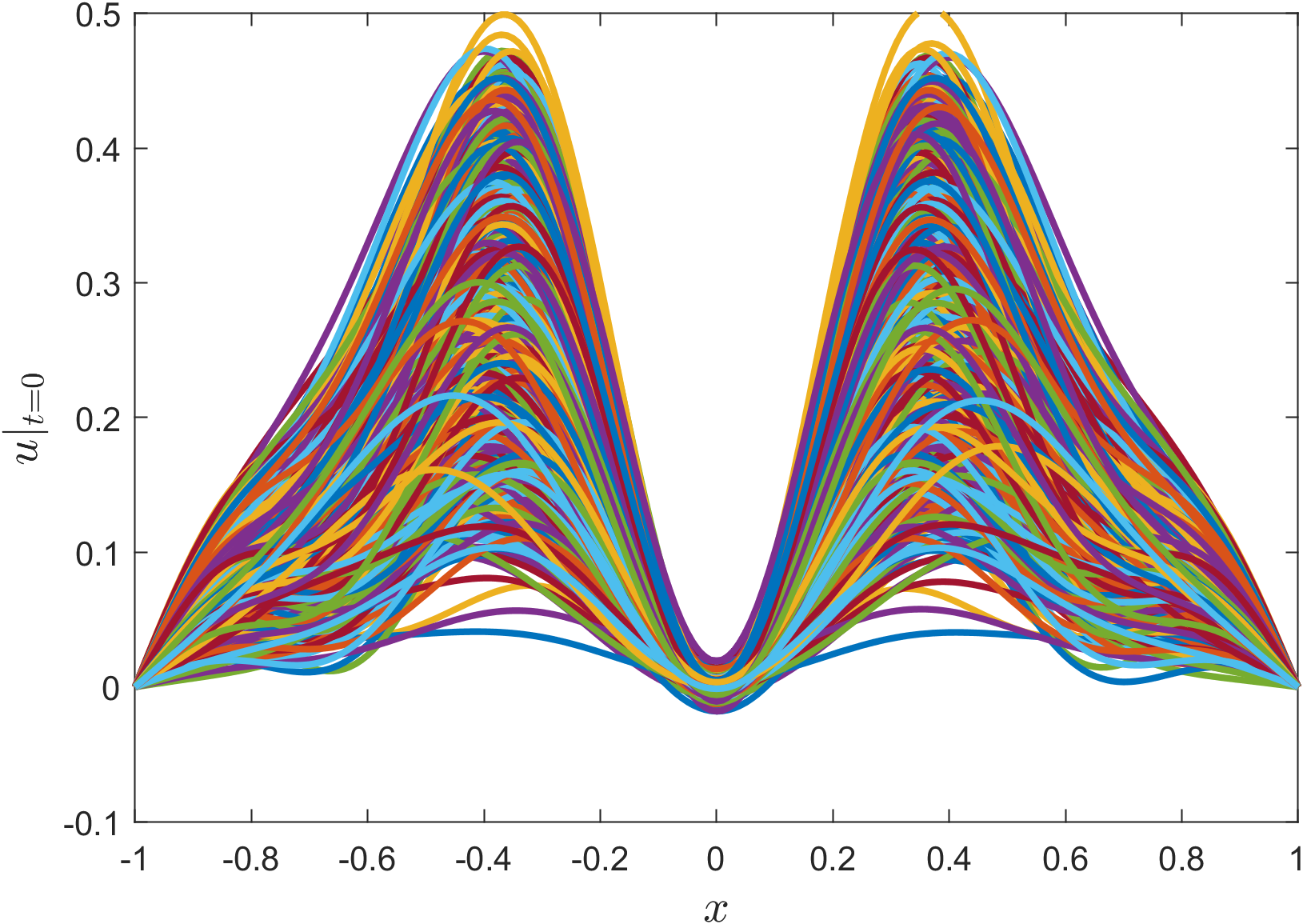}
        \includegraphics[scale=.25]{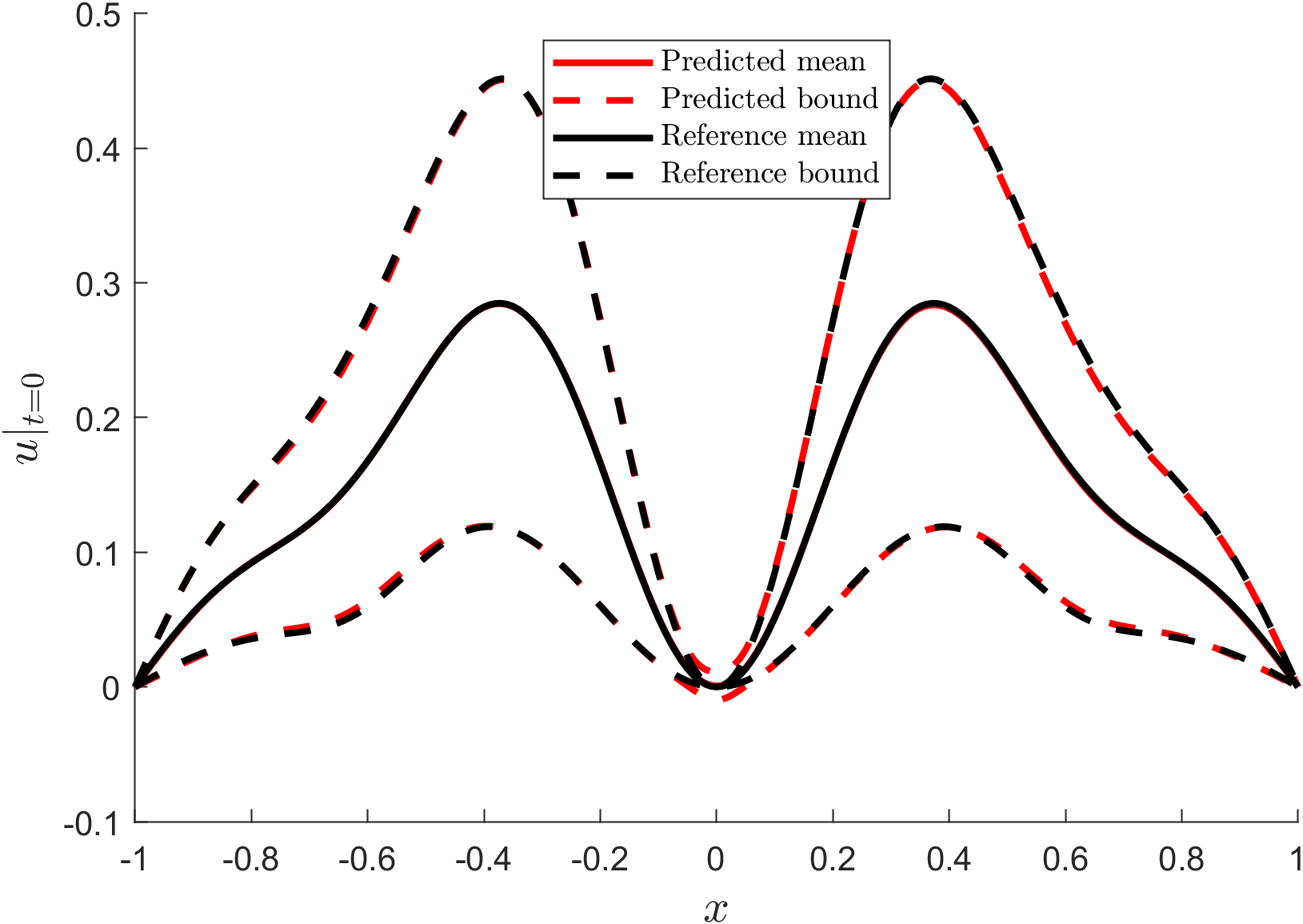}
    }
    \subfigure[]{
        \includegraphics[scale=.25]{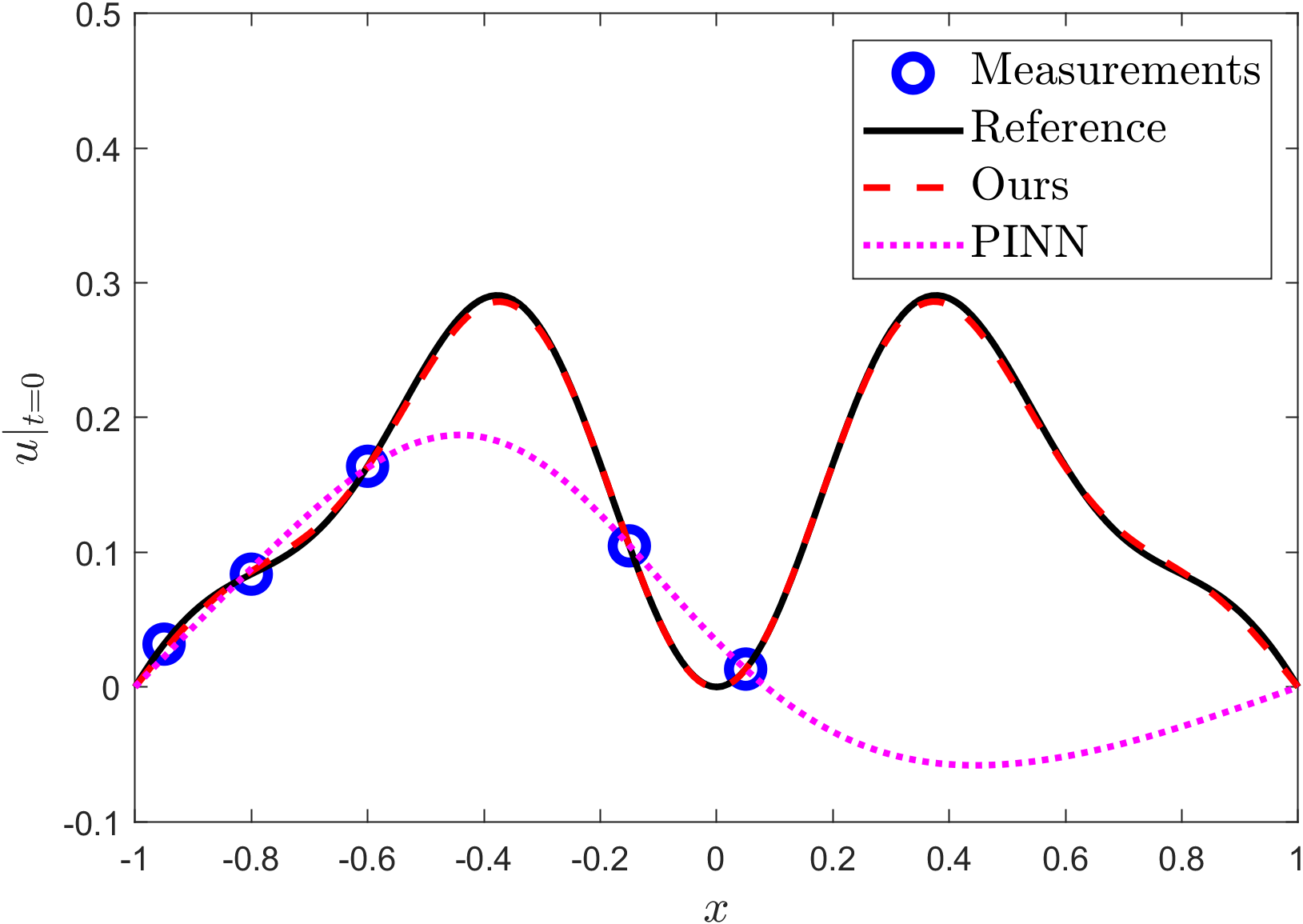}
        \includegraphics[scale=.25]{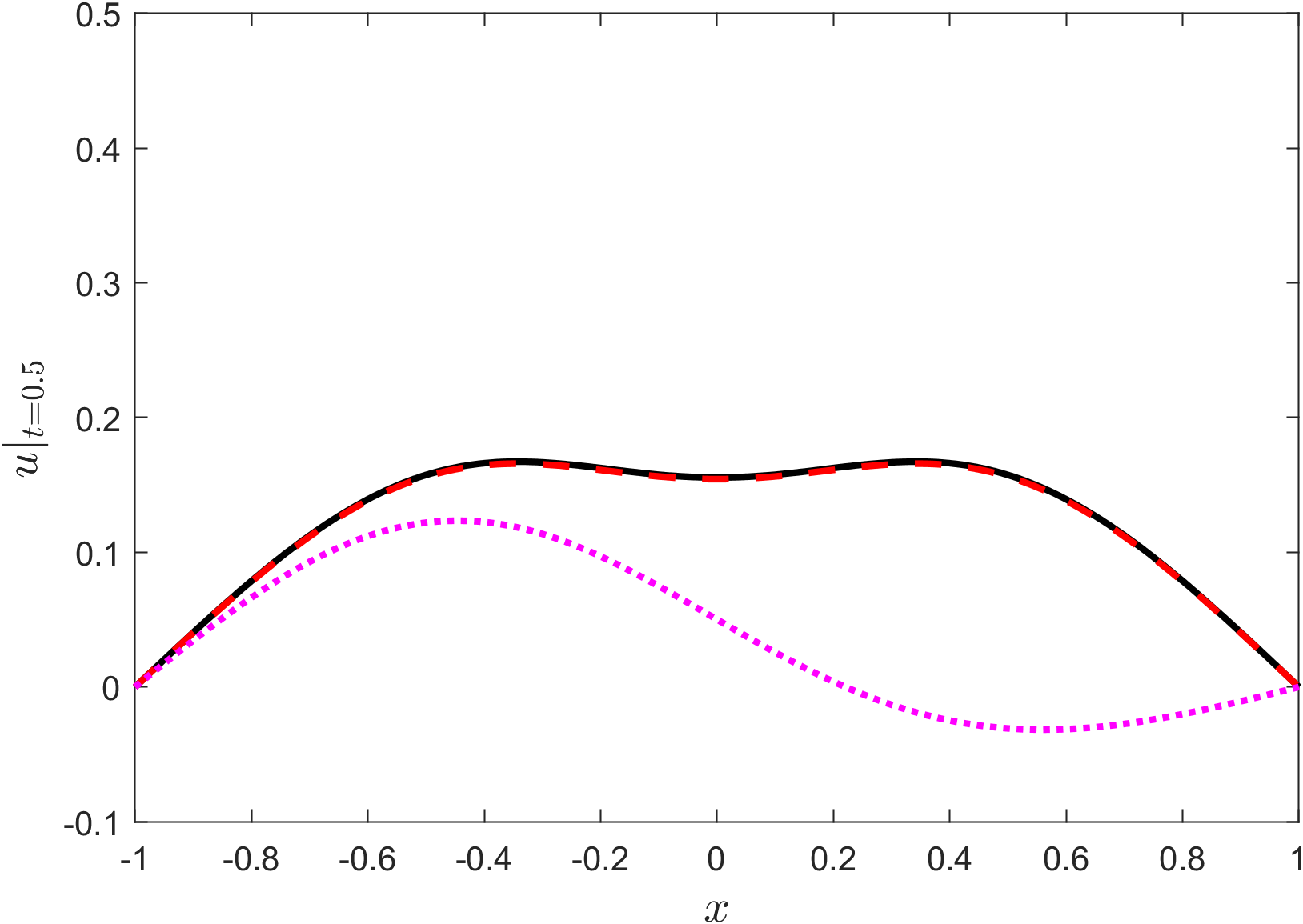}
        \includegraphics[scale=.25]{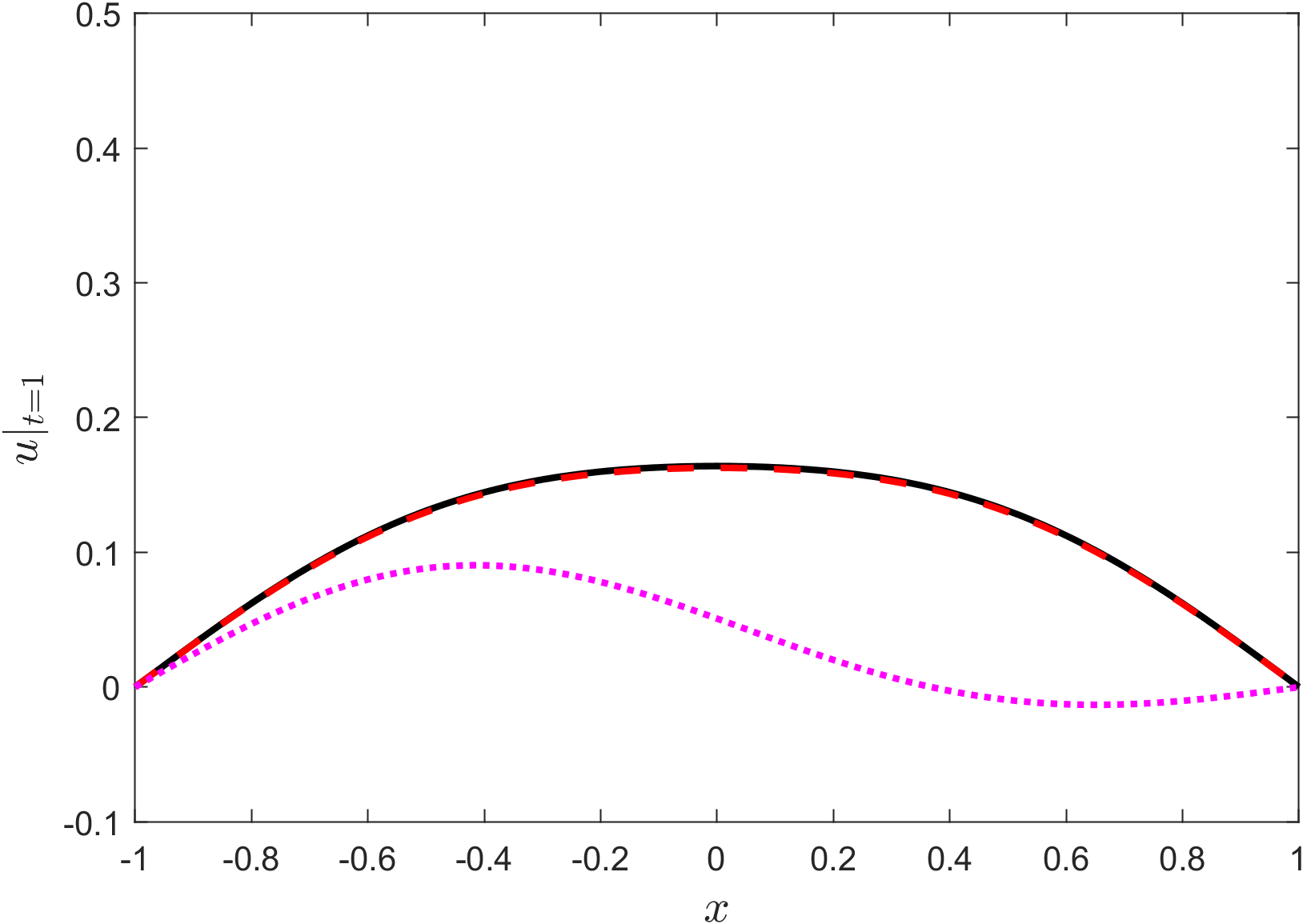}
    }
    \subfigure[]{
        \includegraphics[scale=.25]{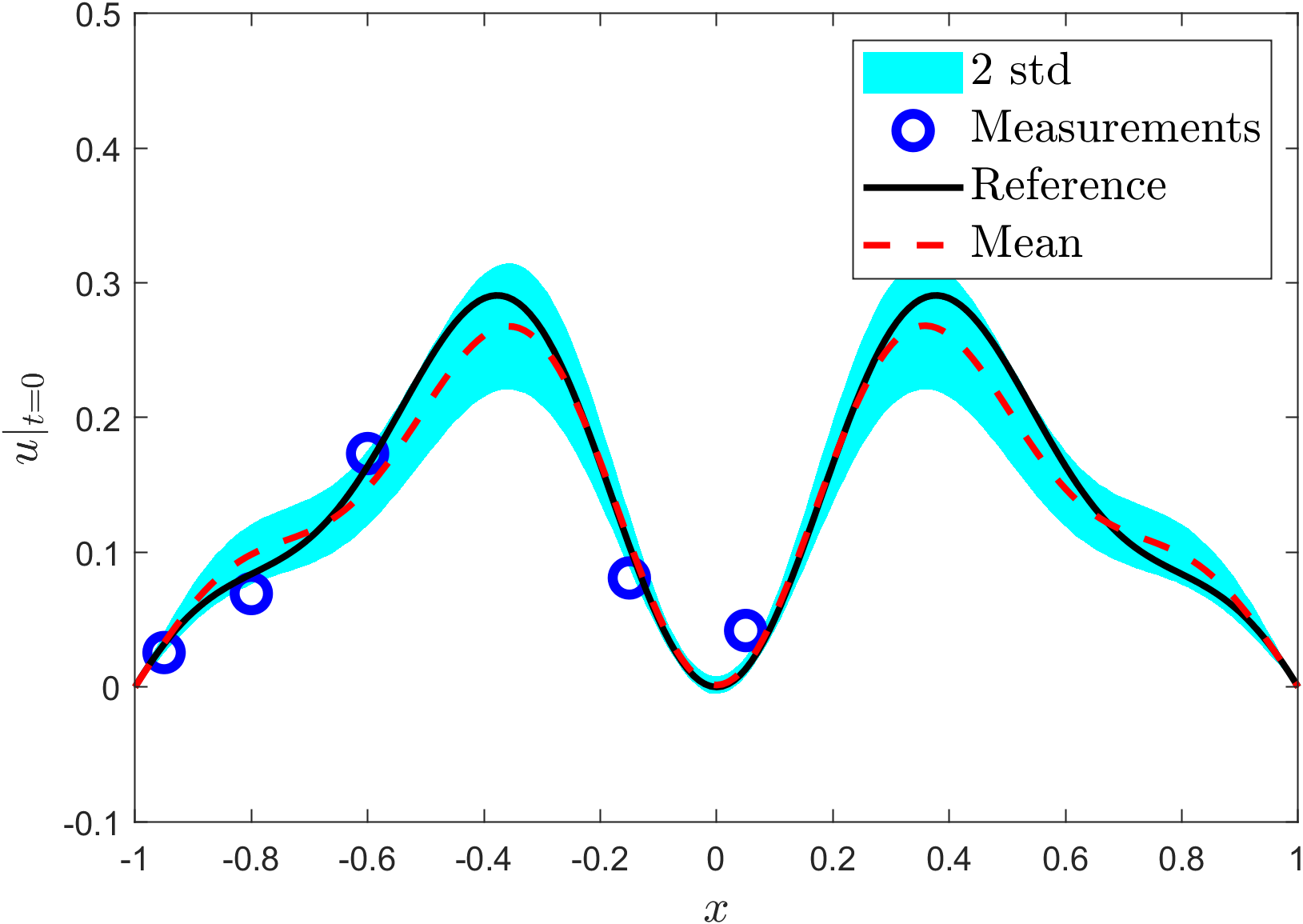}
        \includegraphics[scale=.25]{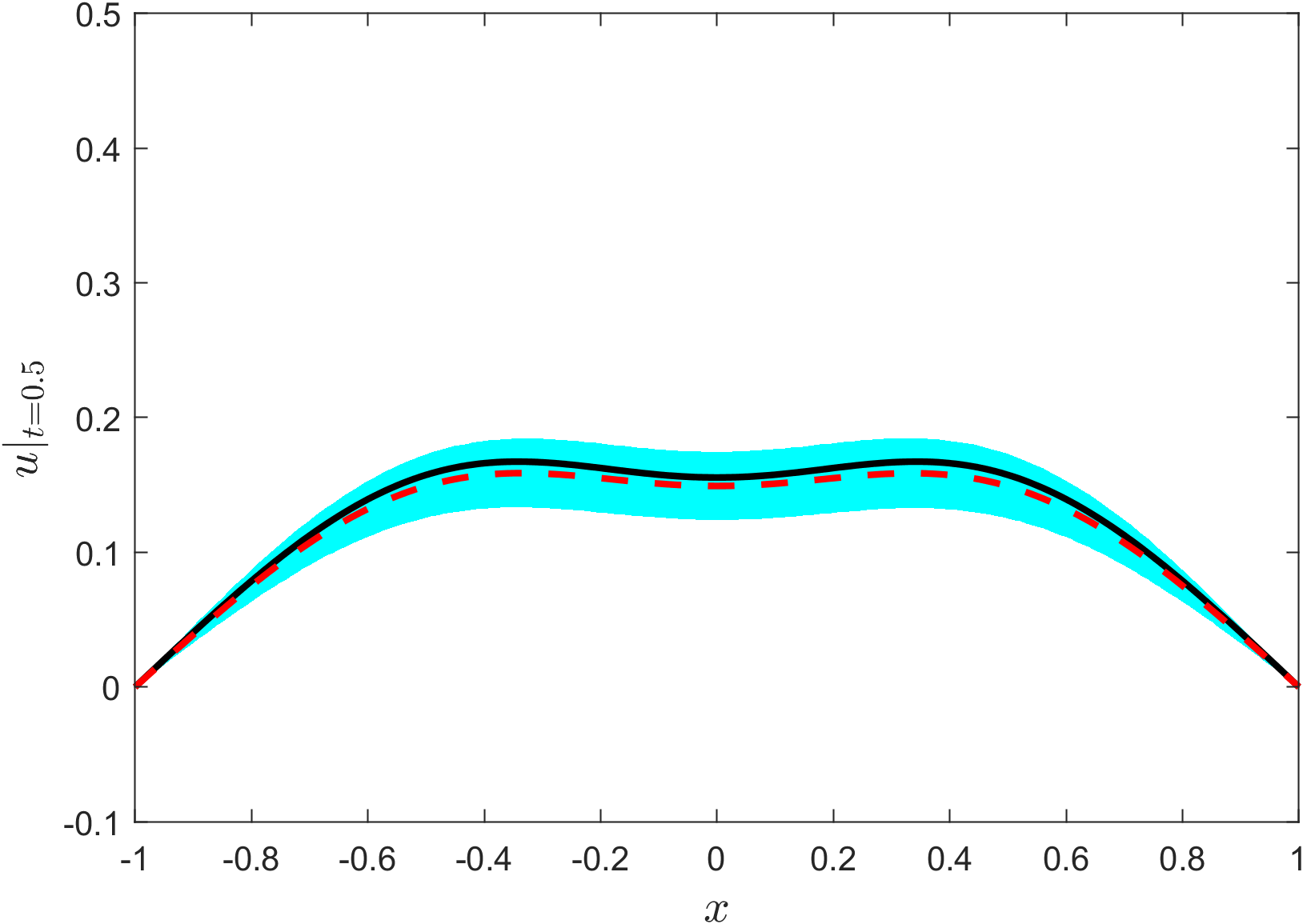}
        \includegraphics[scale=.25]{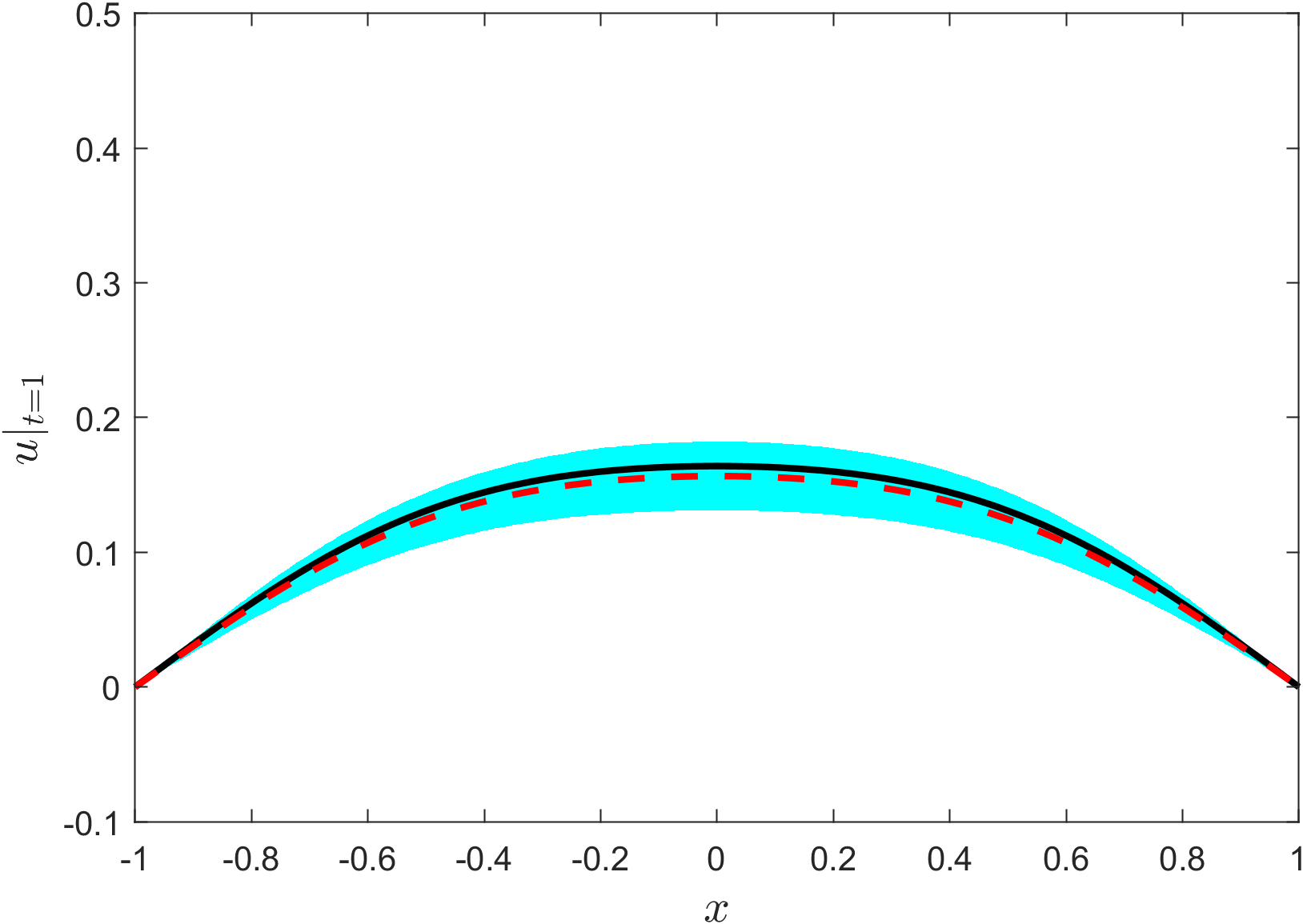}
    }
    \caption{Generator learning and few-shot physics-informed learning on 1-D time-dependent reaction-diffusion equation~\eqref{eq:time_dependent}, with boundary conditions hard-encoded in NN modeling. (a) Left: $1,000$ training samples of $u_0$; middle: $1,000$ samples of $u(0, \cdot)$ from the learned generator; right: statistics computed from samples. The bound is defined as the same as in the caption of Fig.~\ref{fig:example_1}. (b) Predicted $u$ at $t=0, 0.5, 1$ using our approach and the PINN method with noiseless measurements. (c) Predicted mean and uncertainty of $u$ at $t=0, 0.5, 1$ using our approach with HMC for posterior estimate, with noisy measurements.}
    \label{fig:example_3}
\end{figure}

For the training, $2,000$ samples of $u_0(x)$ are generated, displayed in Fig.~\ref{fig:example_3}(a), and each sample forms a physics-informed regression task with $41$ measurements of $u_0$ equidistantly sampled on $[-1, 1]$ as data for initial condition. Besides, for all tasks, a uniform mesh $21\times41$ on temporal-spatial domain $[0, 1]\times[-1, 1]$ is used to compute the PDE residual loss. For the downstream tasks $\tilde{\T}$, $5$ random measurements of $u_0$ are available and the same uniform mesh is applied. The boundary conditions are hard-encoded in NN modeling in both $\{\T_k\}_{k=1}^M$ and $\tilde{\T}$. For the noisy case, the noise $\varepsilon$ is assumed to be independent additive Gaussian noise with $0.02$ noise scale for both measurements of $u_0$ and the PDE residual, i.e., $\varepsilon\sim N(0, 0.02^2)$. Results are presented in Fig.~\ref{fig:example_3} and Table~\ref{tab:example_3}. We can see that our method estimates a good generator of the stochastic processes from data and physics, which provides informative prior knowledge in the downstream few-shot physics-informed regression tasks. The prediction is accurate in both noiseless and noisy cases, and the errors in the noisy case are bounded by the predicted uncertainty. The $L_2$ error of $u$, shown in Table~\ref{tab:example_3}, indicates that our approach outperforms the PINN method by a significant amount, hence demonstrating the effectiveness of bringing prior knowledge into solving similar tasks.

\begin{table}[ht]
    \footnotesize
    \centering
    \begin{tabular}{|c|c|c|}
    \hline
         & PINN & MH-PINN \\
         \hline
       Error (\%) & $78.77$ & $0.22$ \\
       \hline
    \end{tabular}
    \caption{$L_2$ relative error of $u$ for the downstream few-shot physics-informed learning task on Eq.~\eqref{eq:time_dependent} with clean data of $u_0$ using our approach and the PINN method.}
    \label{tab:example_3}
\end{table}

\subsection{2-D nonlinear Allen-Cahn equation}
\label{subsec:example_4}
We now move to a 2-D steady nonlinear Allen-Cahn equation with Dirichlet boundary conditions \cite{yang2021b}:
\begin{align}\label{eq:allen_cahn}
    \lambda \Delta u + u(u^2 - 1) &= f, x, y \in [0, 1],\\
    u(x, 0) = u(x, 1) &= u(0, y) = u(1, y) = 0,
\end{align}
where $\lambda = 0.1$ is a constant and $f$ is the source term. Here, we impose a distribution to $f$, which is derived from Eq.~\eqref{eq:allen_cahn} and the following distribution of the solution $u$:
\begin{equation}
\begin{aligned}
    u(x, y) = \frac{1}{5}\sum_{j=1}^5 \xi_j \frac{\sin(j\pi x) \sin(j\pi y)}{j^2\pi^2}, x,y\in[0, 1],
\end{aligned}
\end{equation}
with i.i.d. random variables $\xi_j, j=1,...,5,$ subject to uniform distribution, i.e. $\xi_j\sim U[0, 1)$.
In this example, we wish to use our method to learn generators of both $u$ and $f$ from data of $f$ and physics in Eq.~\eqref{eq:allen_cahn}, and use it to solve the downstream task $\tilde{\T}$ with insufficient data $\tilde{\D}$. This example corresponds to Eq.~\eqref{eq:problem} with $\F_k, b_k$ being the same among tasks and $f_k, u_k$ being task-specific, and the data is $\D_k=\{(x_k^i, y_k^i),f_k^i\}_{i=1}^{N_k^f}$.

\begin{figure}[ht]
    \centering
    \subfigure[]{
        \includegraphics[scale=.25]{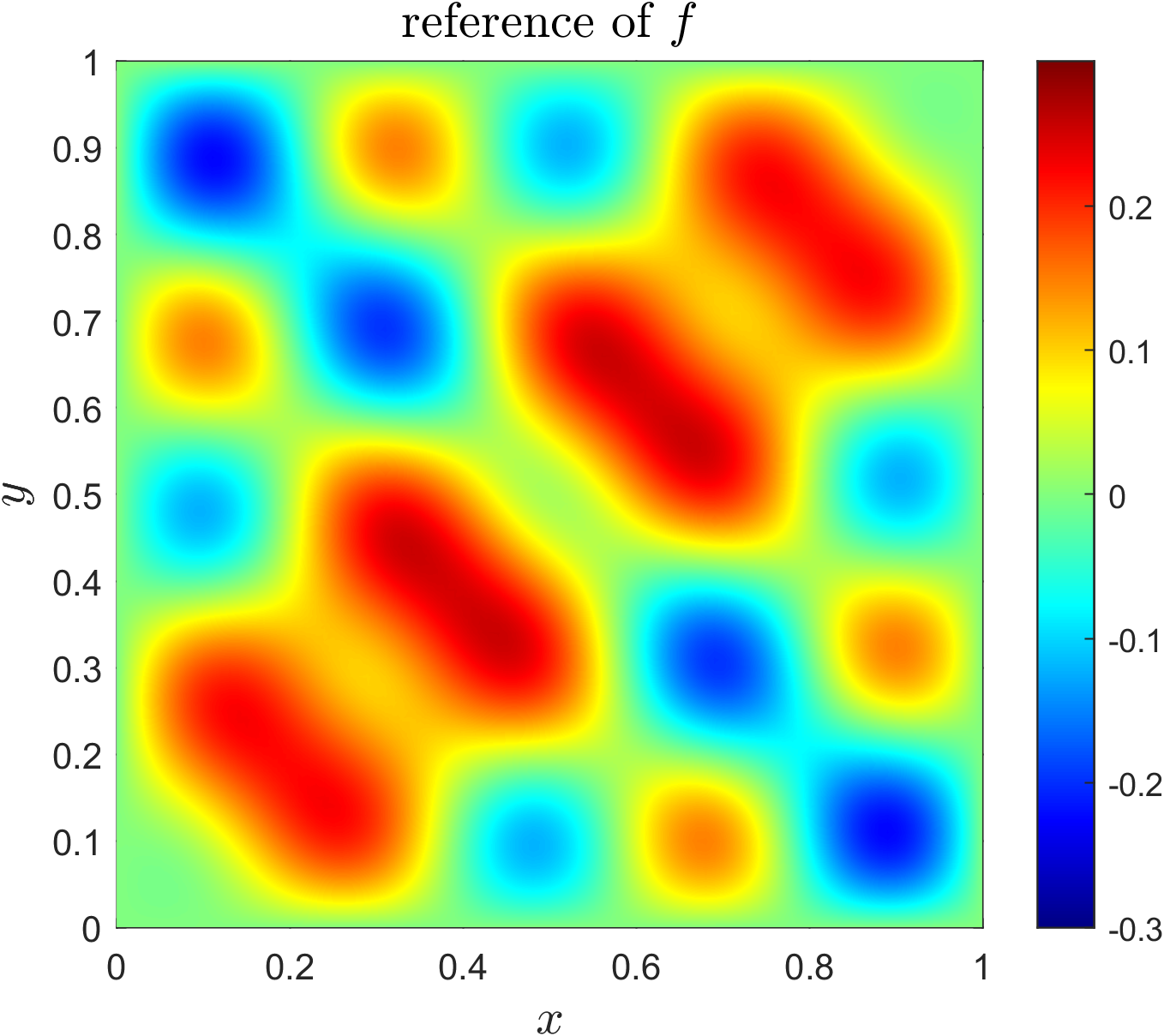}
        \includegraphics[scale=.25]{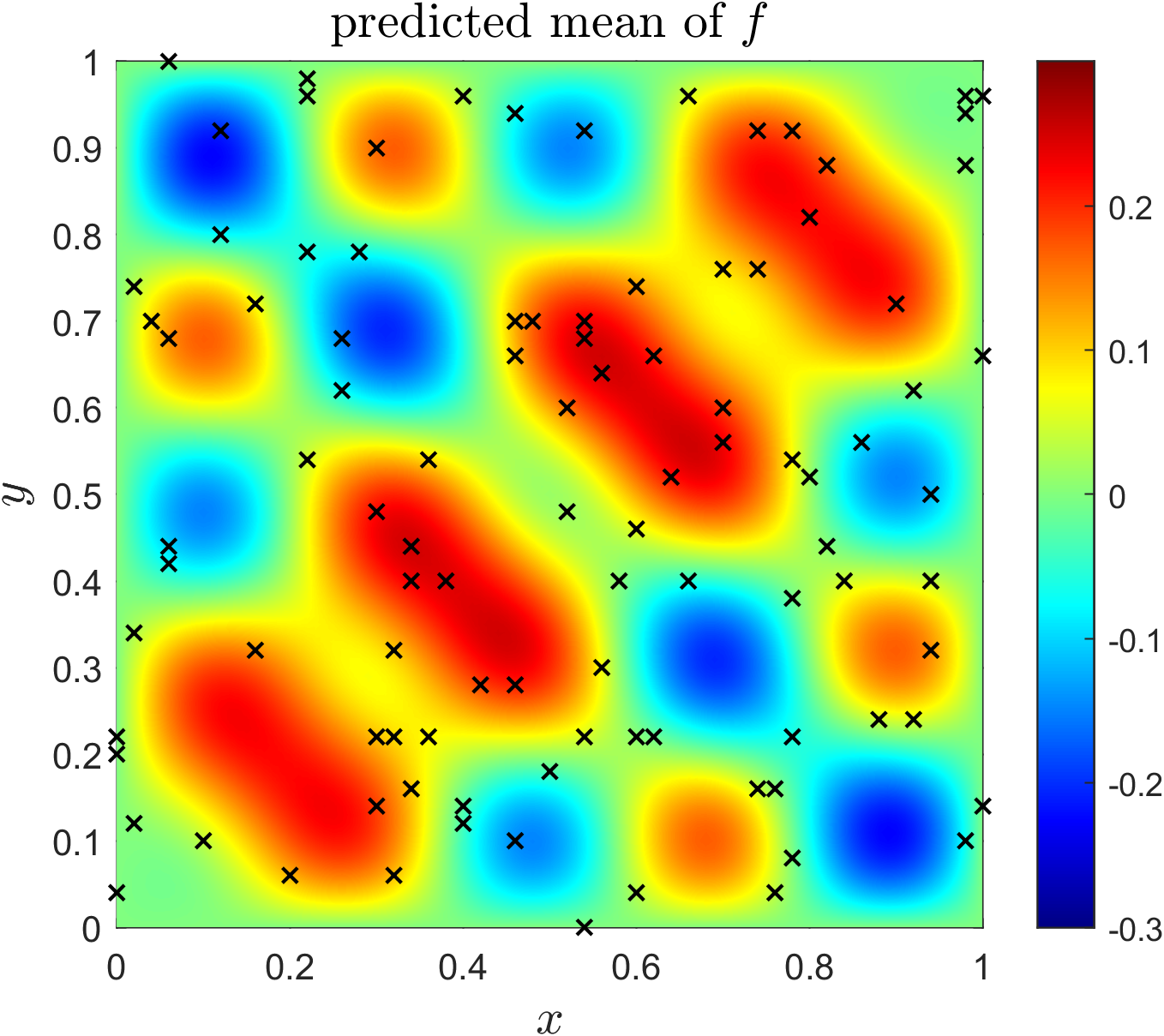}
        \includegraphics[scale=.25]{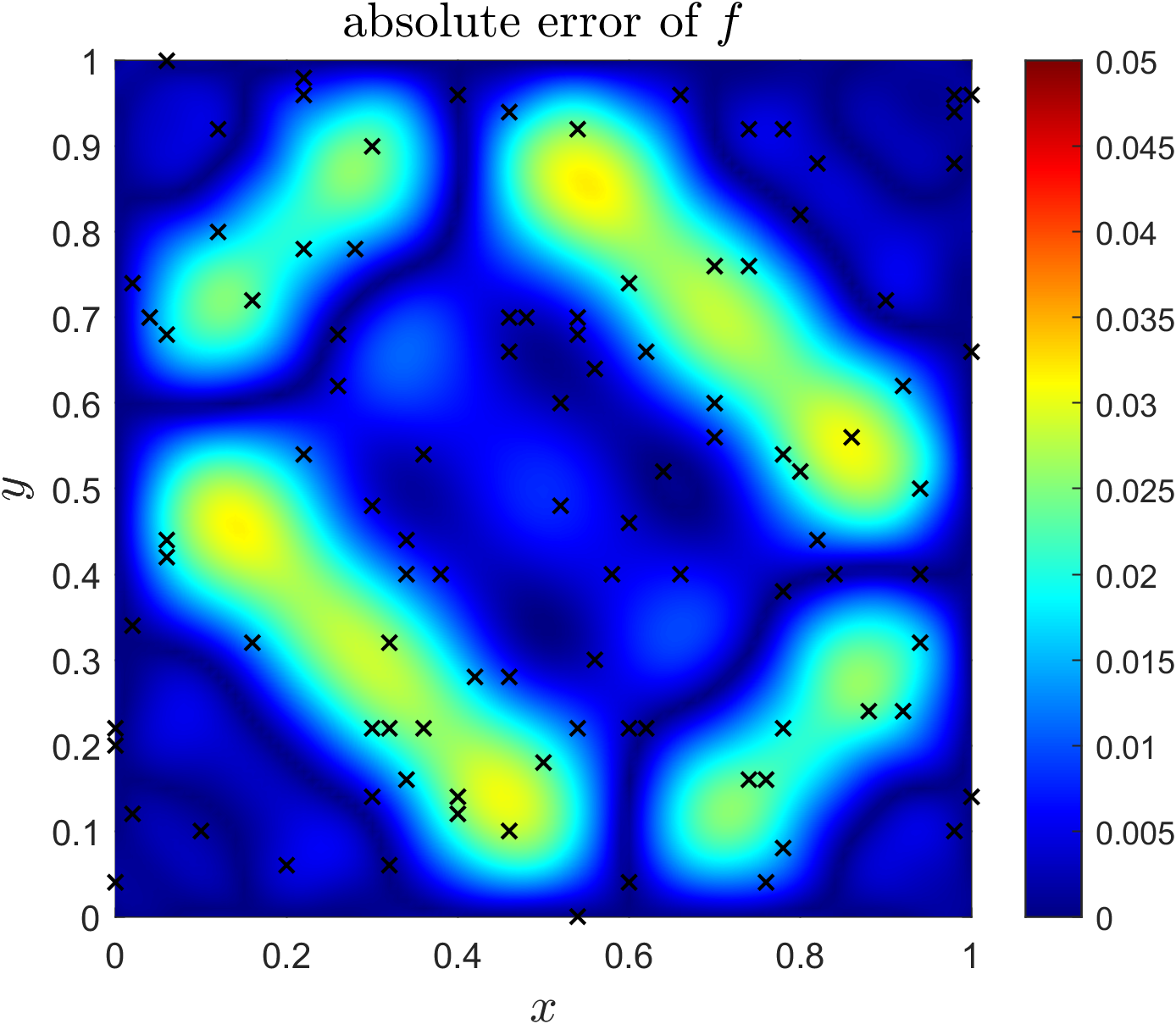}
        \includegraphics[scale=.25]{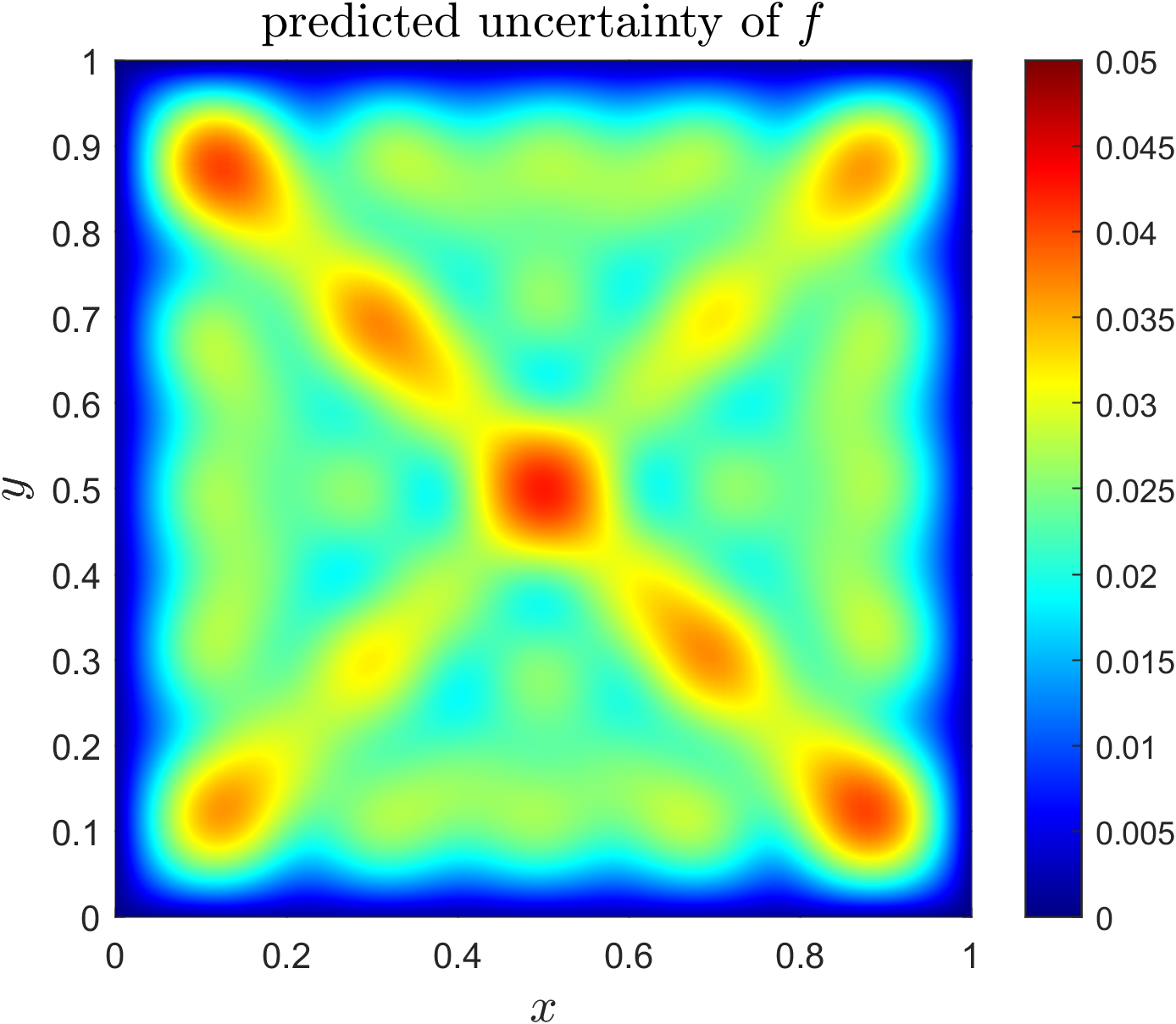}
    }
    \subfigure[]{
        \includegraphics[scale=.25]{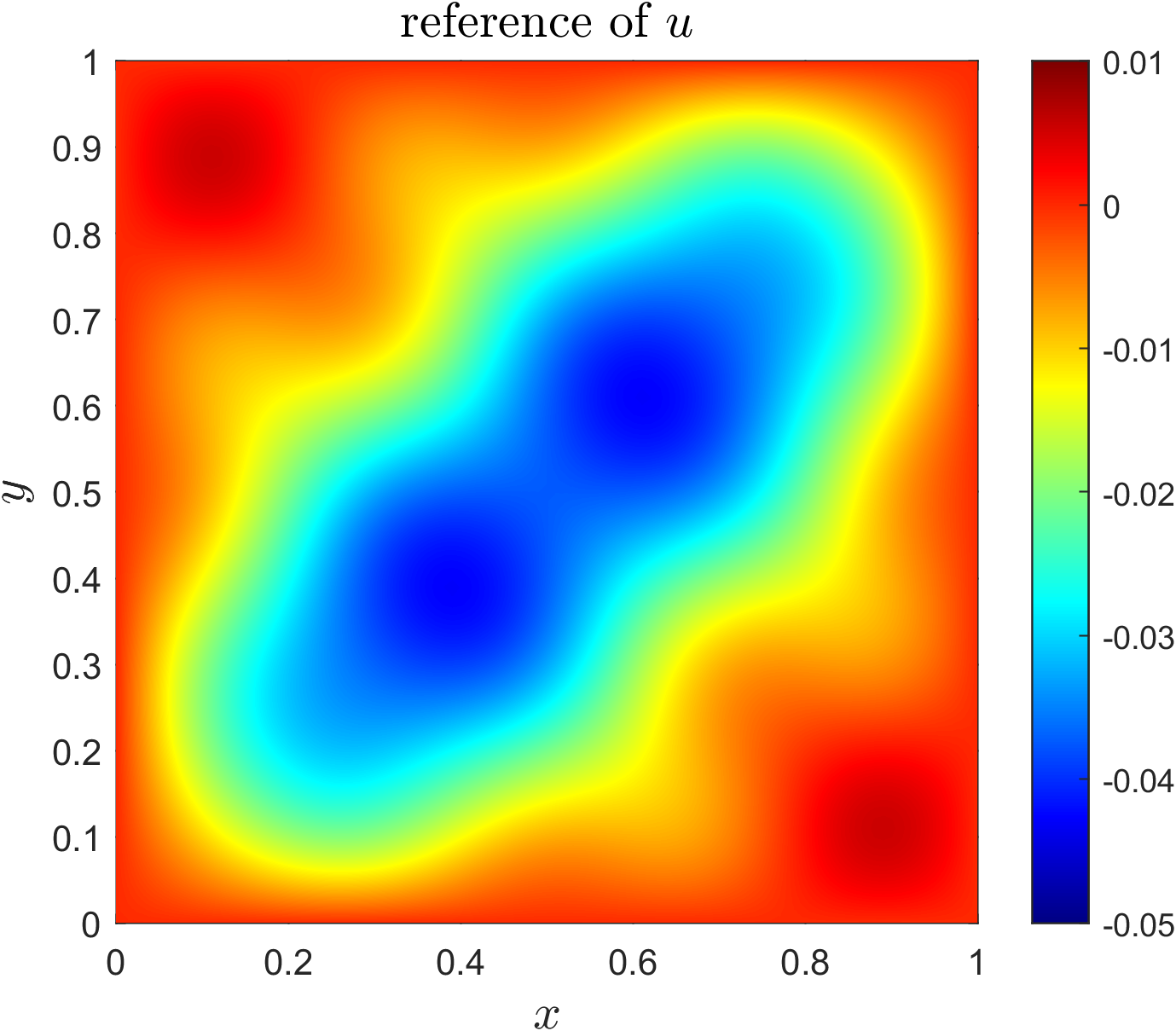}
        \includegraphics[scale=.25]{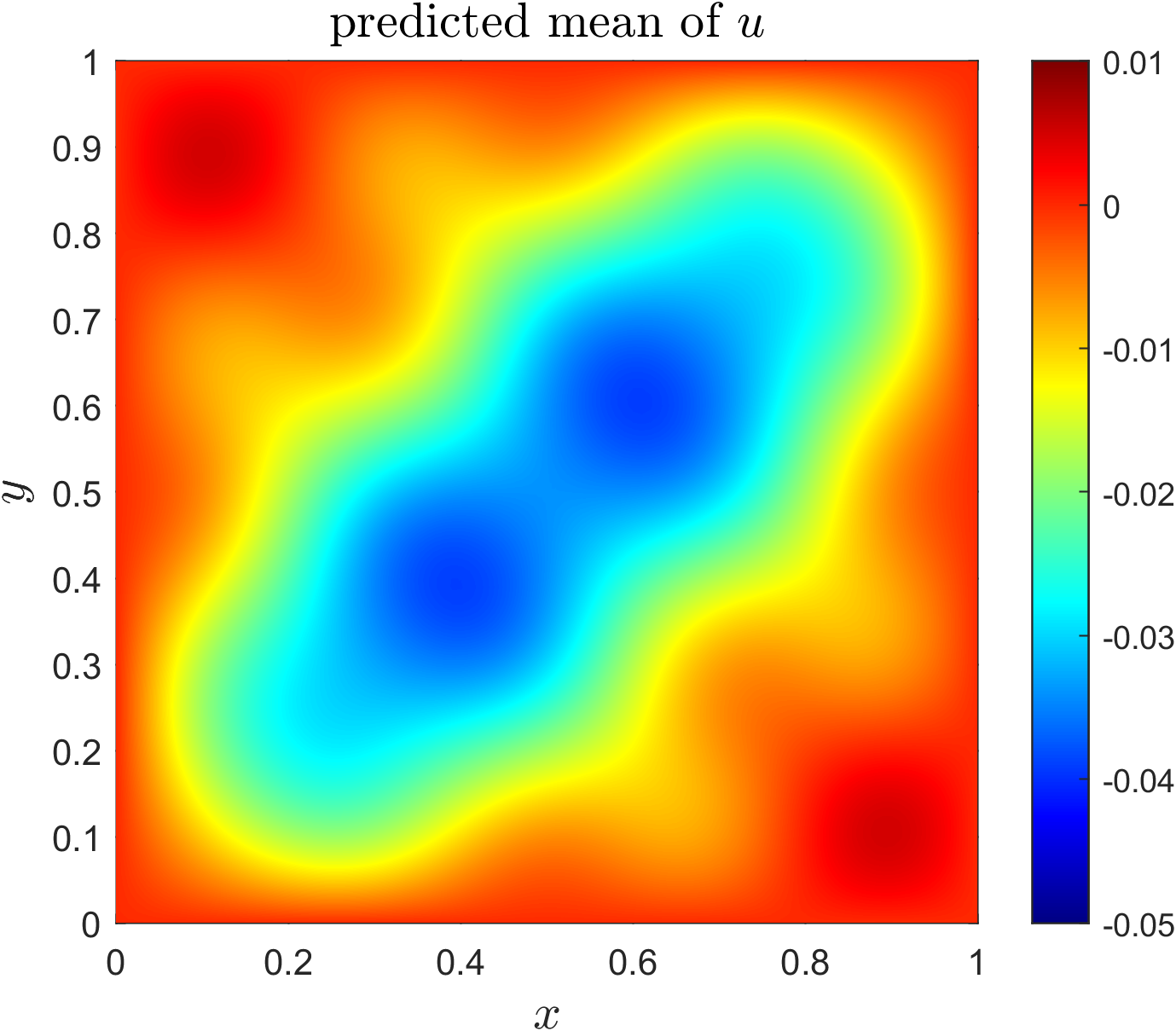}
        \includegraphics[scale=.25]{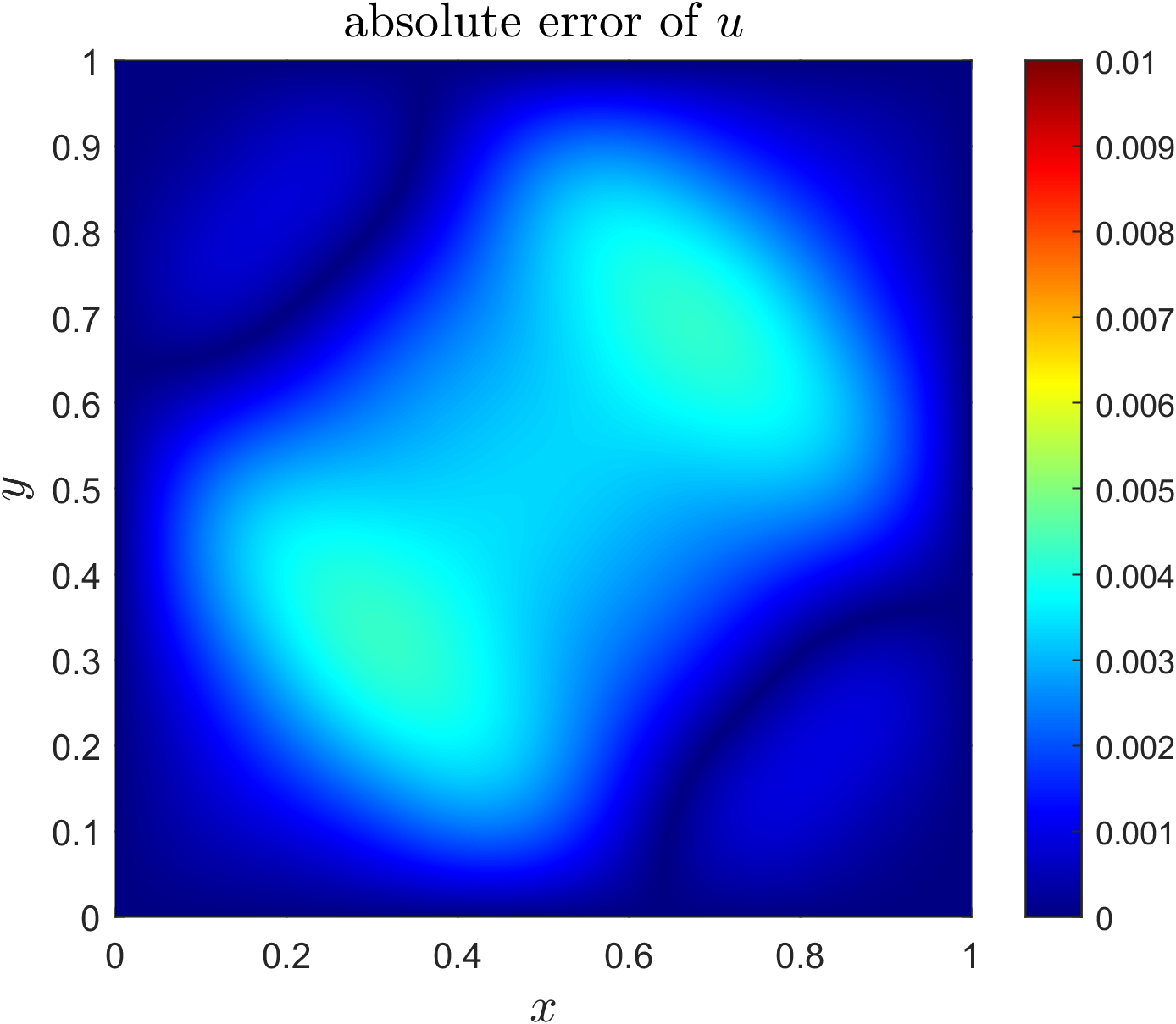}
        \includegraphics[scale=.25]{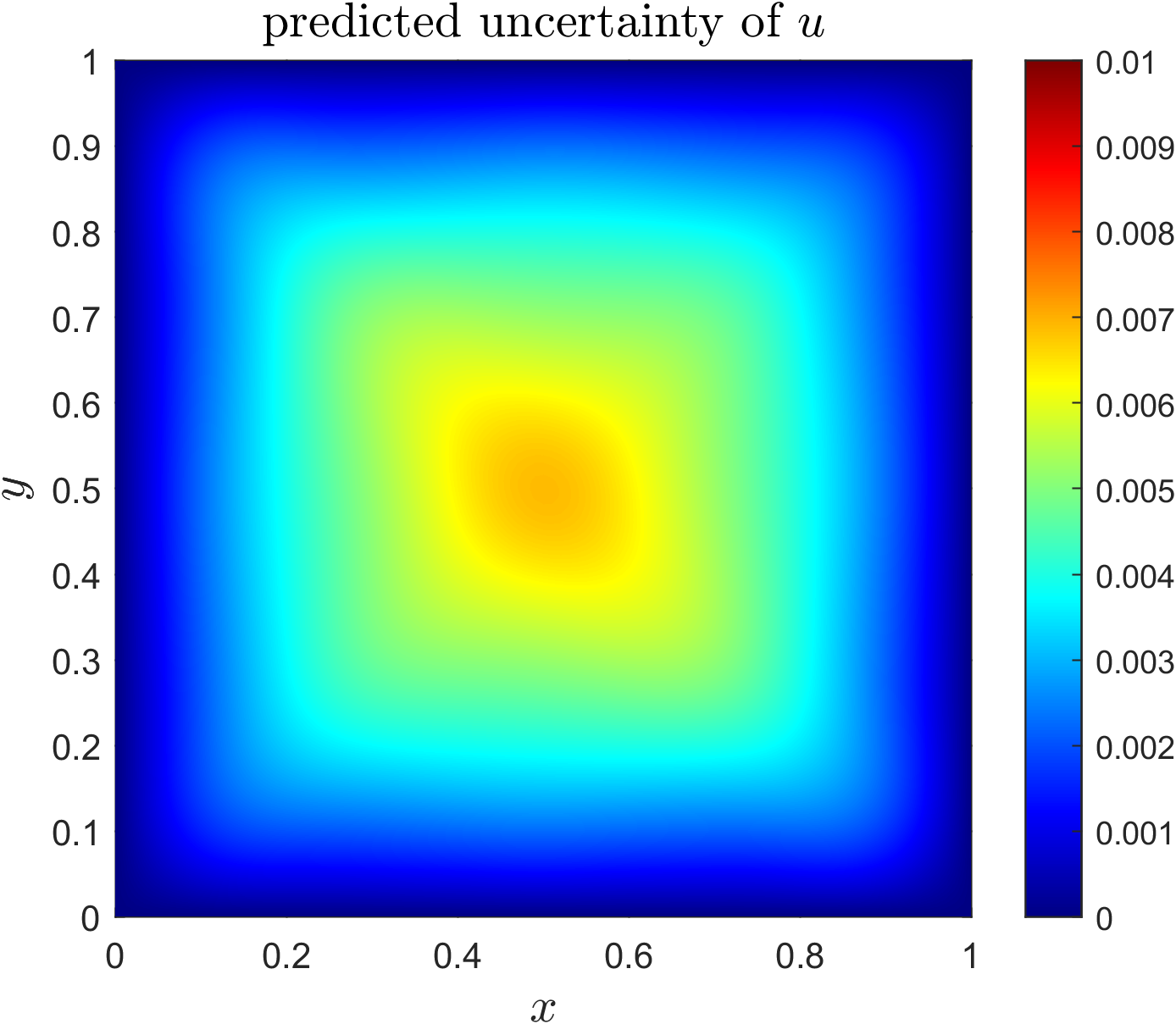}
    }
    \caption{Results for few-shot physics-informed learning on the 2-D nonlinear Allen-Cahn equation Eq.~\eqref{eq:allen_cahn} with noisy measurements of $f$. Predicted mean $\mu$ and standard deviation $\sigma$ are computed over $1,000$ posterior samples from HMC. The absolute error is defined as the absolute value of difference between the reference and $\mu$. Black crosses represent the locations of the measurements on $f$.}
    \label{fig:example_5}
\end{figure}

To train the MH-PINNs and NFs, we sample $2,000$ $f$ from its distribution, each of which is resolved with a $51\times51$ uniform mesh on 2-D spatial domain $[0, 1]\times[0, 1]$. As for the downstream task, $100$ random measurements of $f$ on the uniform mesh are assumed to be available. The noise is assumed to be independent additive Gaussian noise with $0.05$ noise scale. In both $\T_k$ and $\tilde{\T}$, the boundary conditions are hard-encoded in NN modeling. Results as well as the locations of the measurements are presented in Fig.~\ref{fig:example_5} and Table~\ref{tab:example_5}. Similar to all previous examples, our approach delivers accurate and trustworthy predictions, showing that prior knowledge is learned and transferred well in both deterministic and Bayesian inferences.

\begin{table}[ht]
    \footnotesize
    \centering
    \begin{tabular}{|c|c|c|}
    \hline
         & PINN & MH-PINN \\
         \hline
       Error (\%) & $12.82$ & $0.30$\\
       \hline
    \end{tabular}
    \caption{$L_2$ relative error of $u$ for the downstream few-shot physics-informed learning task on Eq.~\eqref{eq:allen_cahn} with clean data of $f$, using our approach and the PINN method.}
    \label{tab:example_5}
\end{table}
\subsection{2-D Stochastic Helmholtz equation}
\label{subsec:example_5}
The last example we test in this paper is the 2-D Helmholtz equation with stochastic source term and Dirichlet boundary conditions \cite{perdikaris2016multifidelity}:
\begin{align}\label{eq:helmholtz}
    (\lambda^2 - \nabla^2) u &= f, x, y \in [0, 2\pi],\\
    u(x, 0) = u(x, 2\pi) &= u(0, y) = u(2\pi, y) = 0,
\end{align}
where $\lambda^2$ is the Helmholtz constant and $f$ is defined as follows:
\begin{equation}\label{eq:example_6:source_term}
\begin{aligned}
    f(x, y) = \frac{2}{d}\{\sum_{i=1}^{d/4} \xi_i \sin(ix) + \xi_{i+d}\cos(ix) + \xi_{i+2d} \sin(iy) + \xi_{i+3d}\cos(iy)\},
\end{aligned}
\end{equation}
where $\xi_j, j=1,...,d$ are i.i.d. random variables subject to uniform distribution $U[0, 1)$ and $d$ represents the dimension of the randomness. For demonstration purposes, we consider the case where $d=20$ in this paper, unlike the one in \cite{perdikaris2016multifidelity} with $d=100$.

The first case we study is the forward problem with $\lambda^2 =1$ known. This setup corresponds to Eq.~\eqref{eq:problem} with $\F_k, b_k$ being shared among tasks and $u_k, f_k$ being task-specific. Next, we study the inverse problem with unknown $\lambda$, where data on $u$ and $f$ are available, which corresponds to Eq.~\eqref{eq:problem} with only $b_k$ being the same and $u_k, f_k$ and operator $\F_k$ being task-specific. The downstream tasks are defined as the same as $\{\T_k\}_{k=1}^M$ in both cases, but with fewer measurements. 

\begin{figure}[ht]
    \centering
    \subfigure[]{
        \includegraphics[scale=.25]{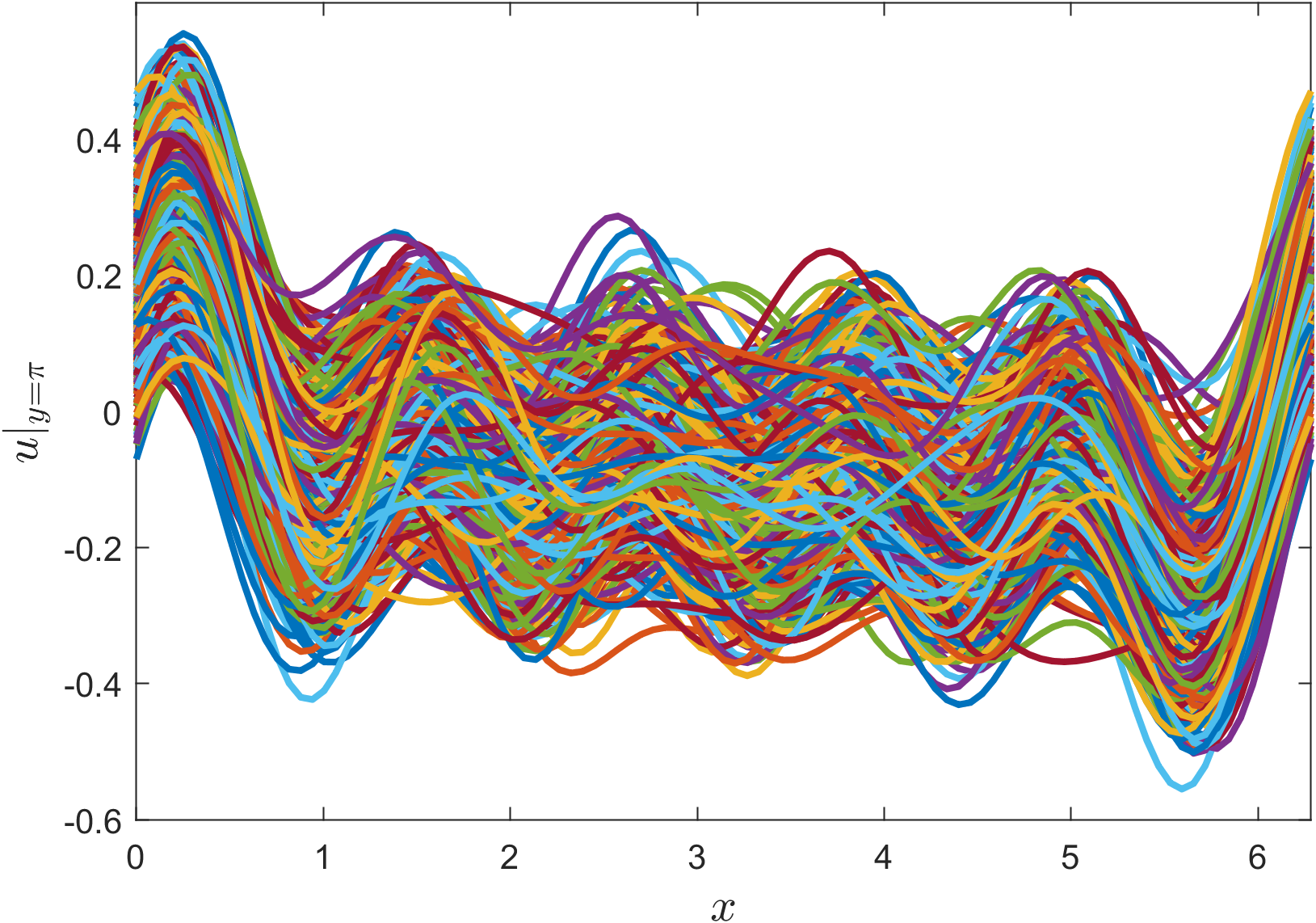}
        \includegraphics[scale=.25]{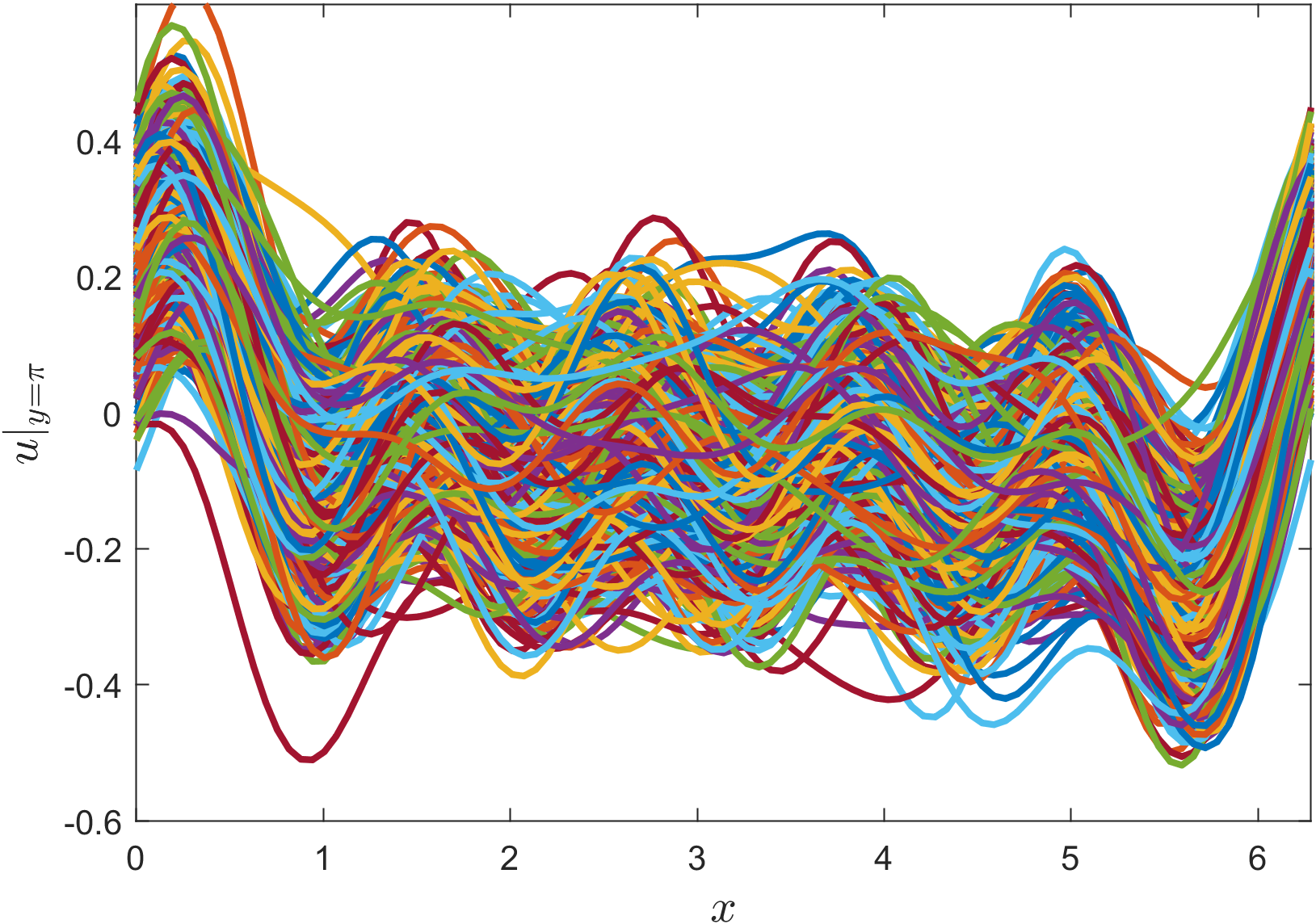}
        \includegraphics[scale=.25]{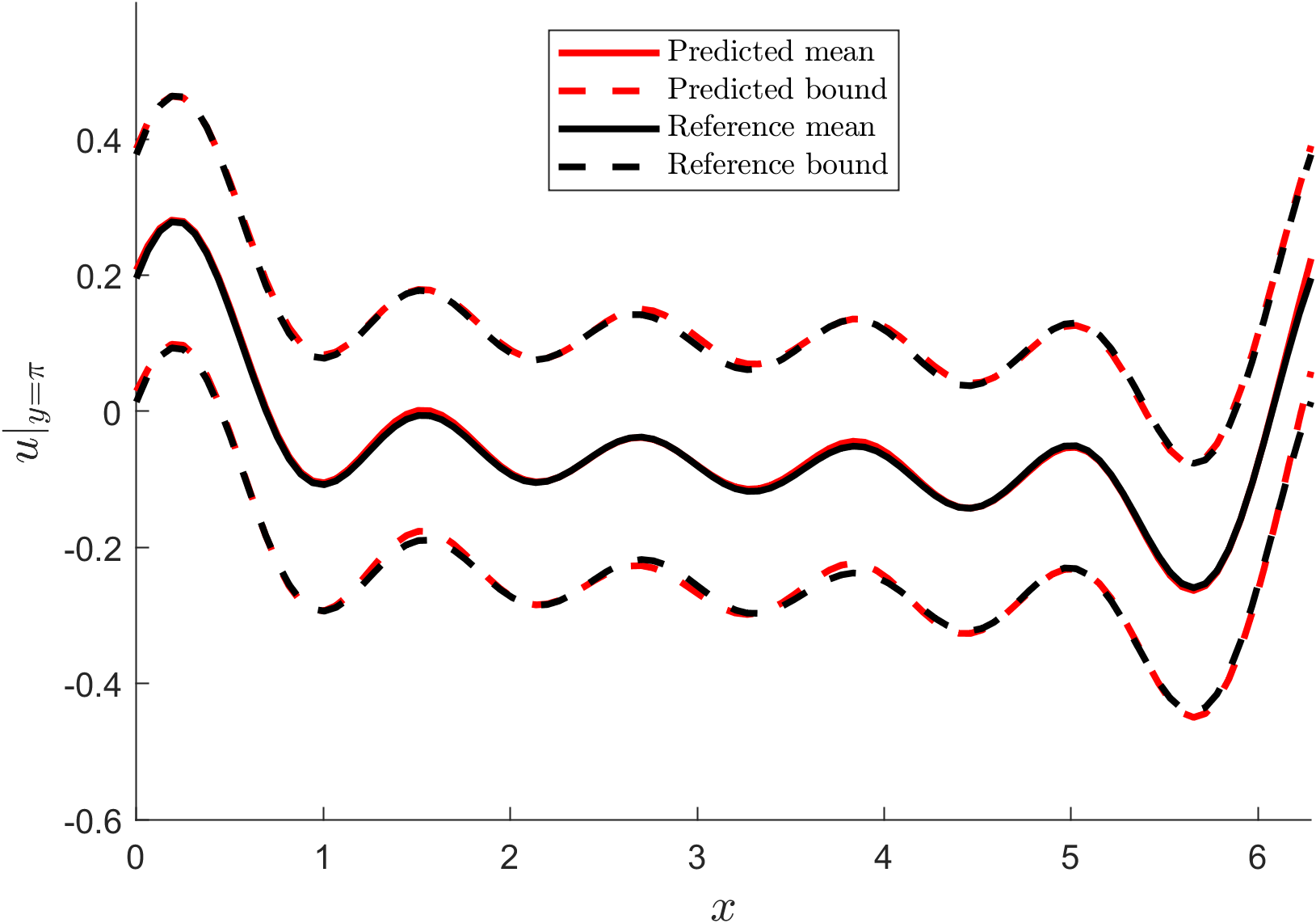}
    }
    \subfigure[]{
        \includegraphics[scale=.25]{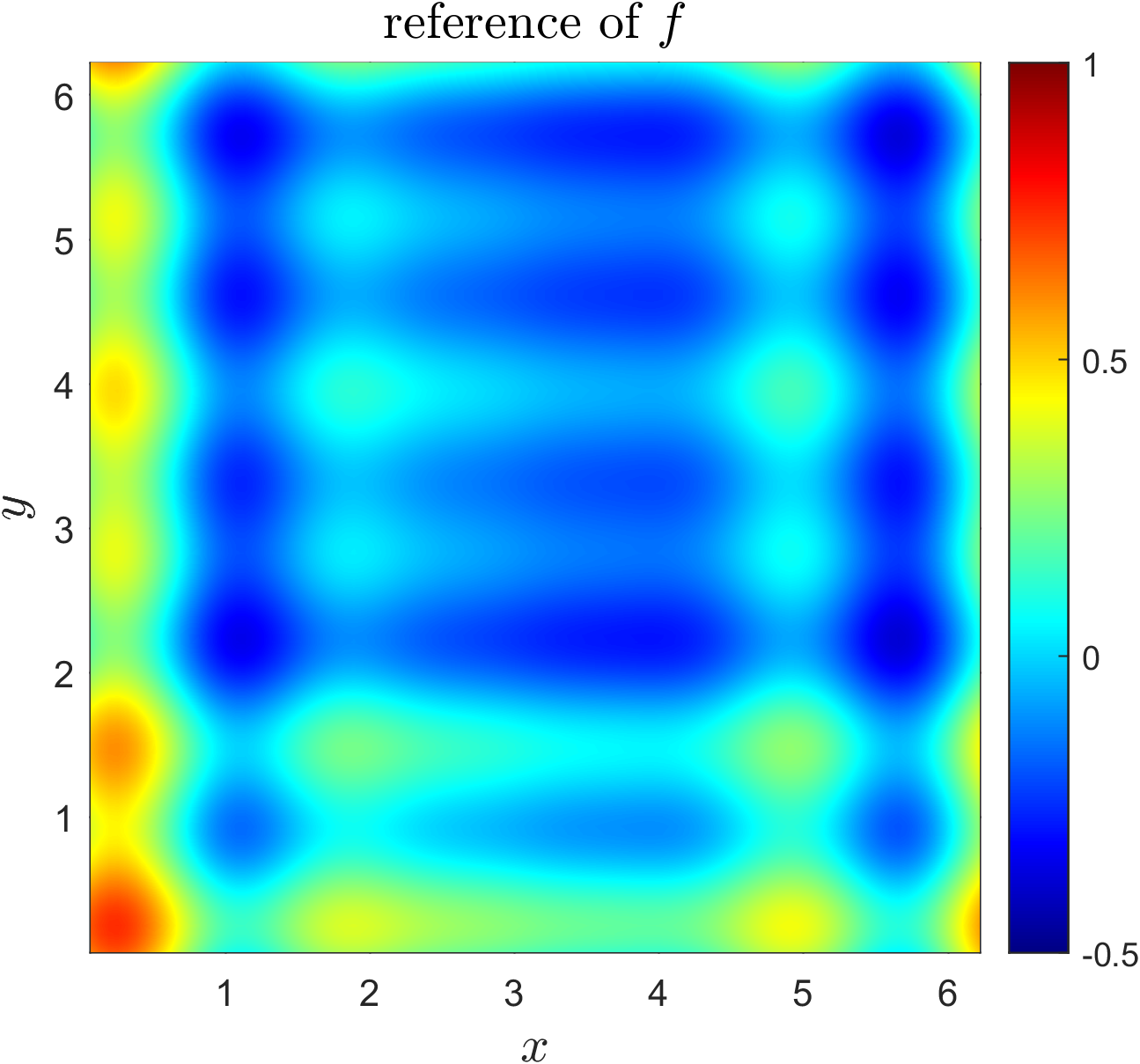}
        \includegraphics[scale=.25]{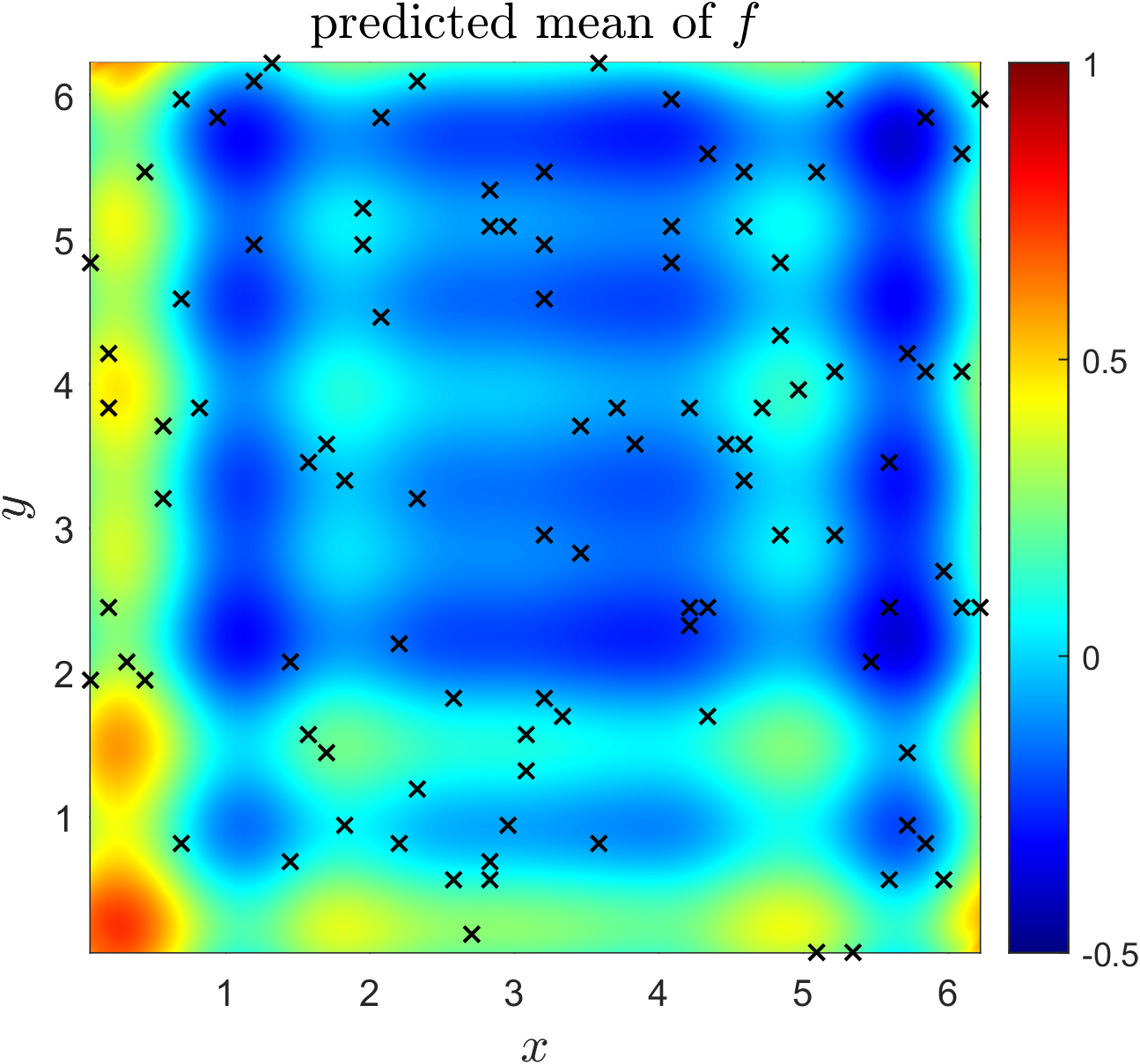}
        \includegraphics[scale=.25]{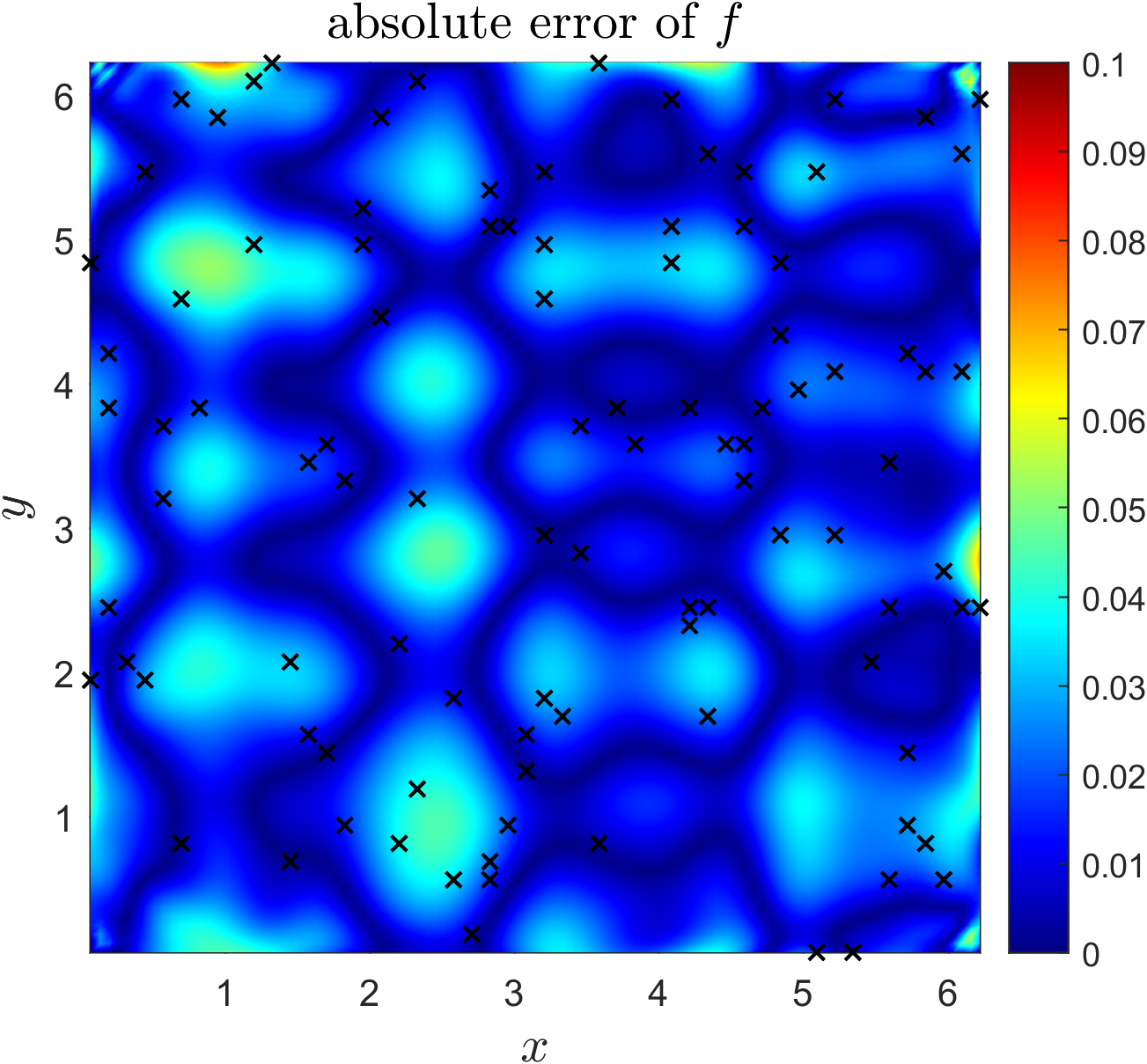}
        \includegraphics[scale=.25]{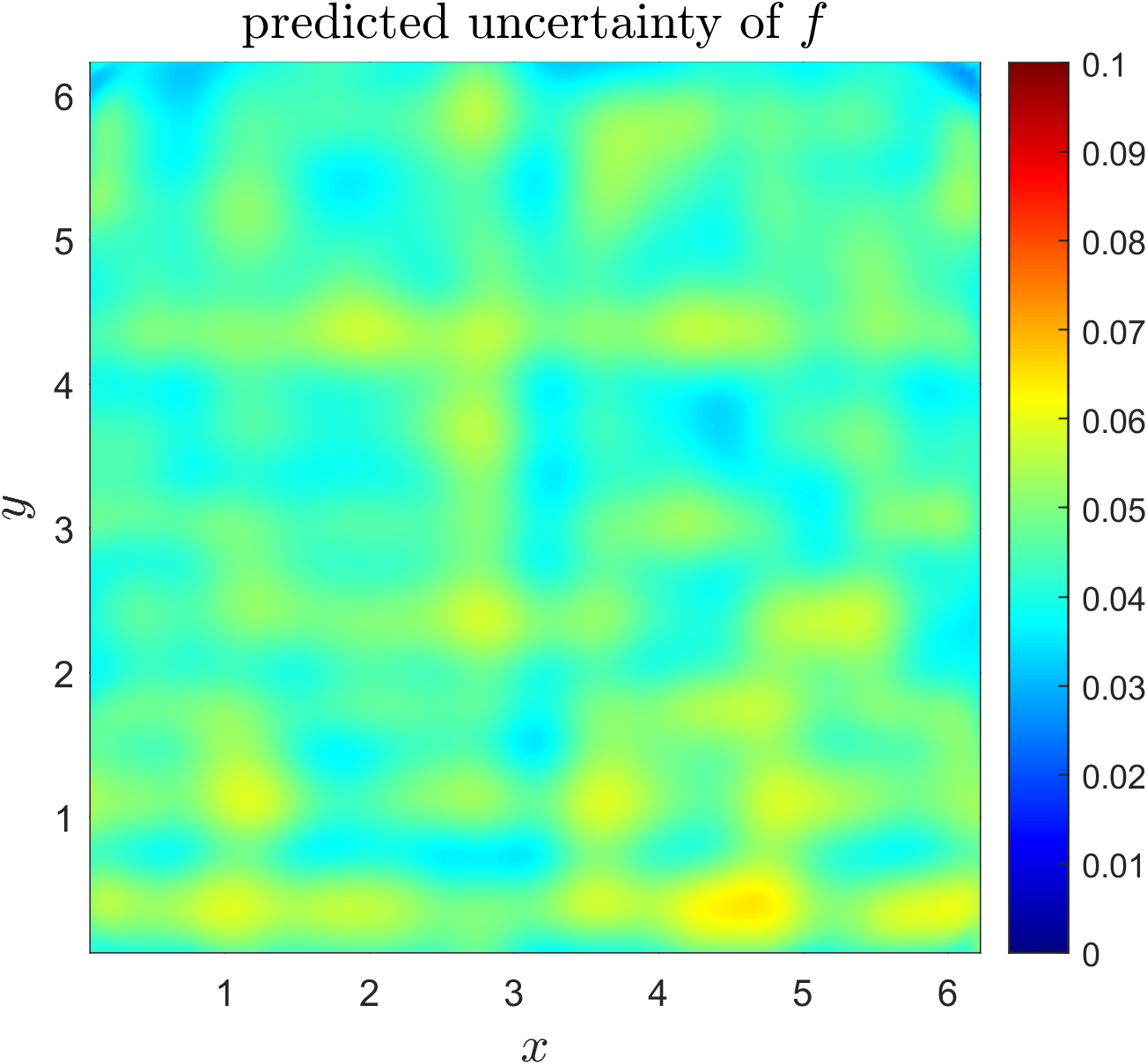}
    }
    \subfigure[]{
        \includegraphics[scale=.25]{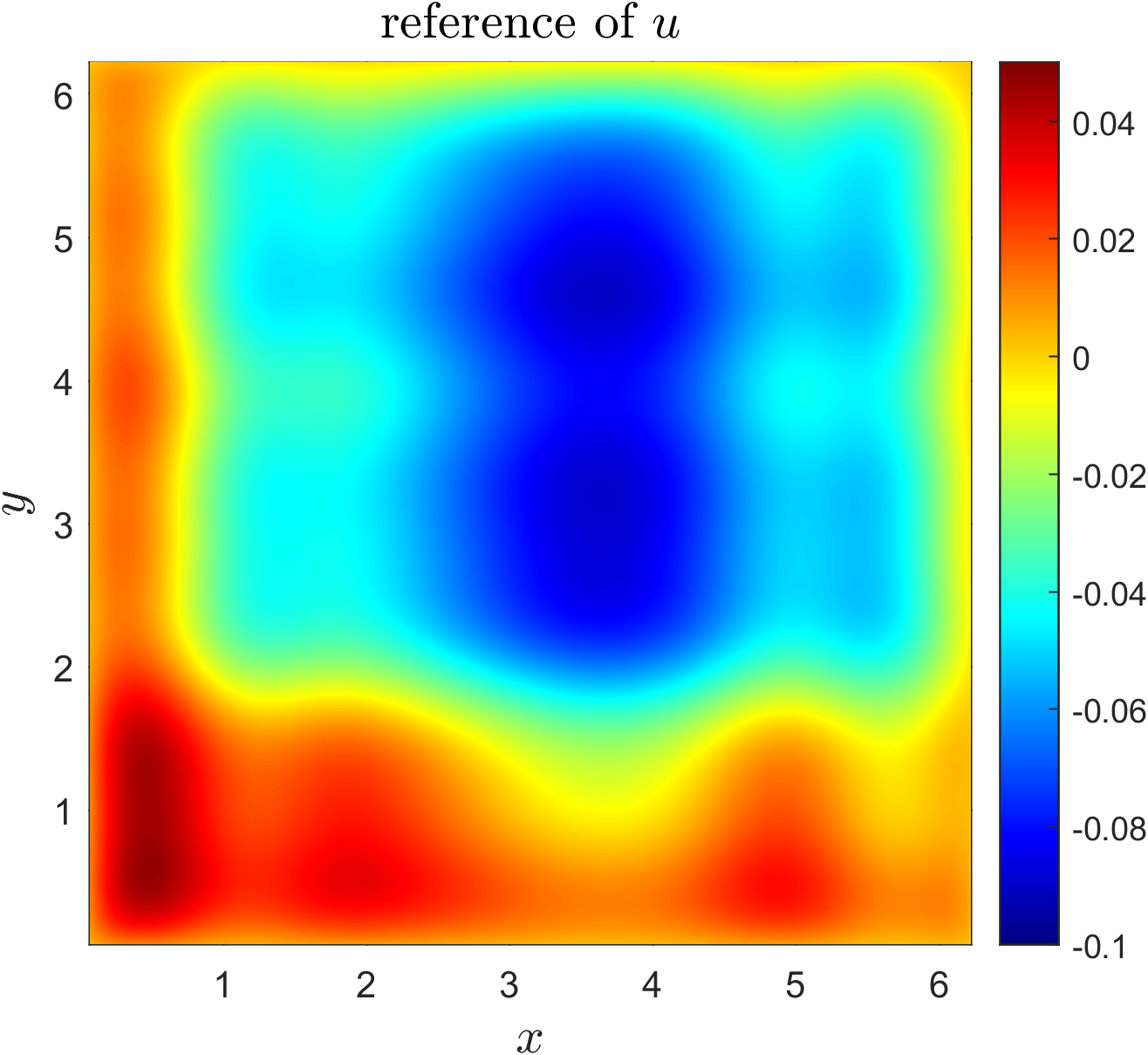}
        \includegraphics[scale=.25]{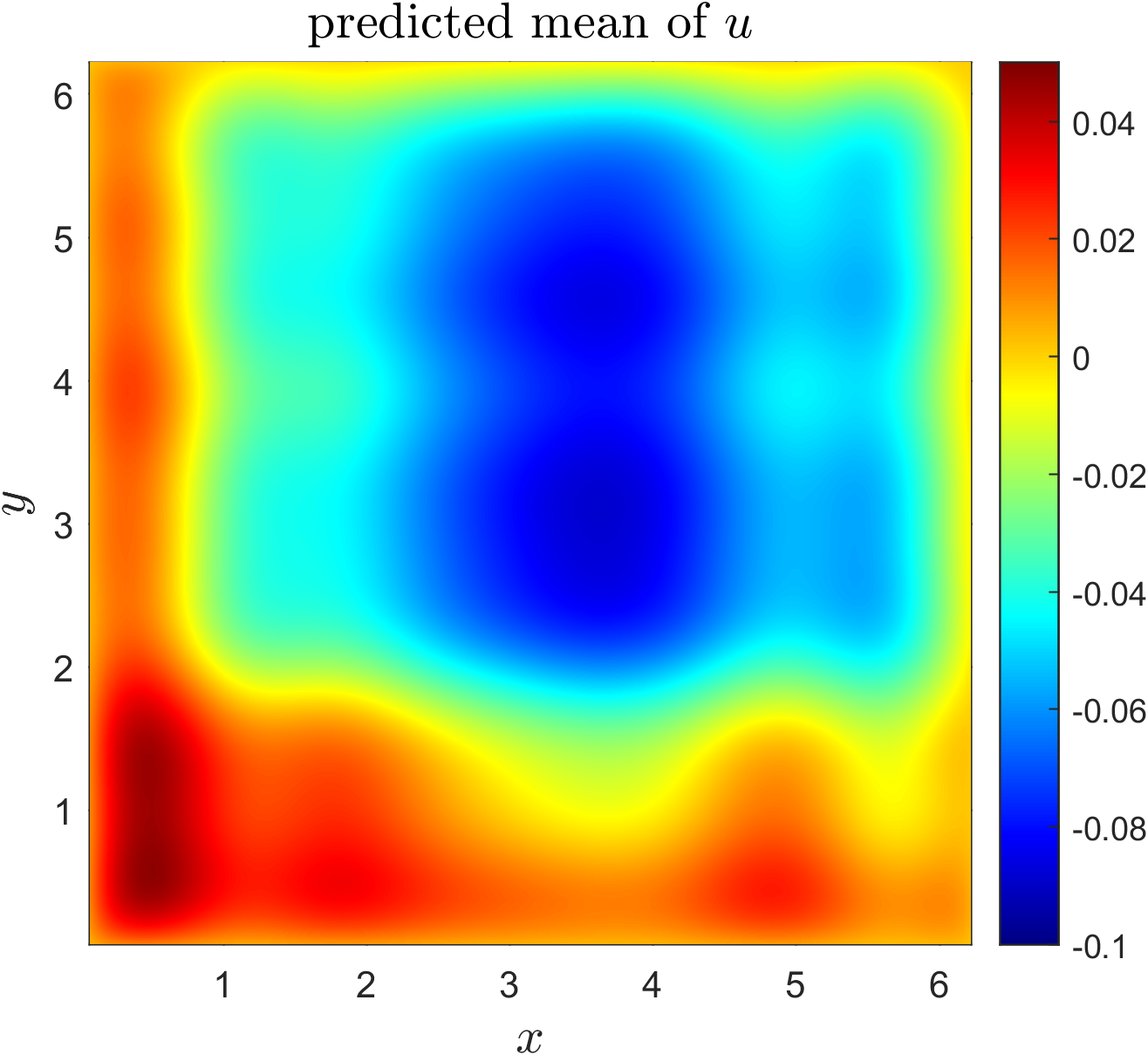}
        \includegraphics[scale=.25]{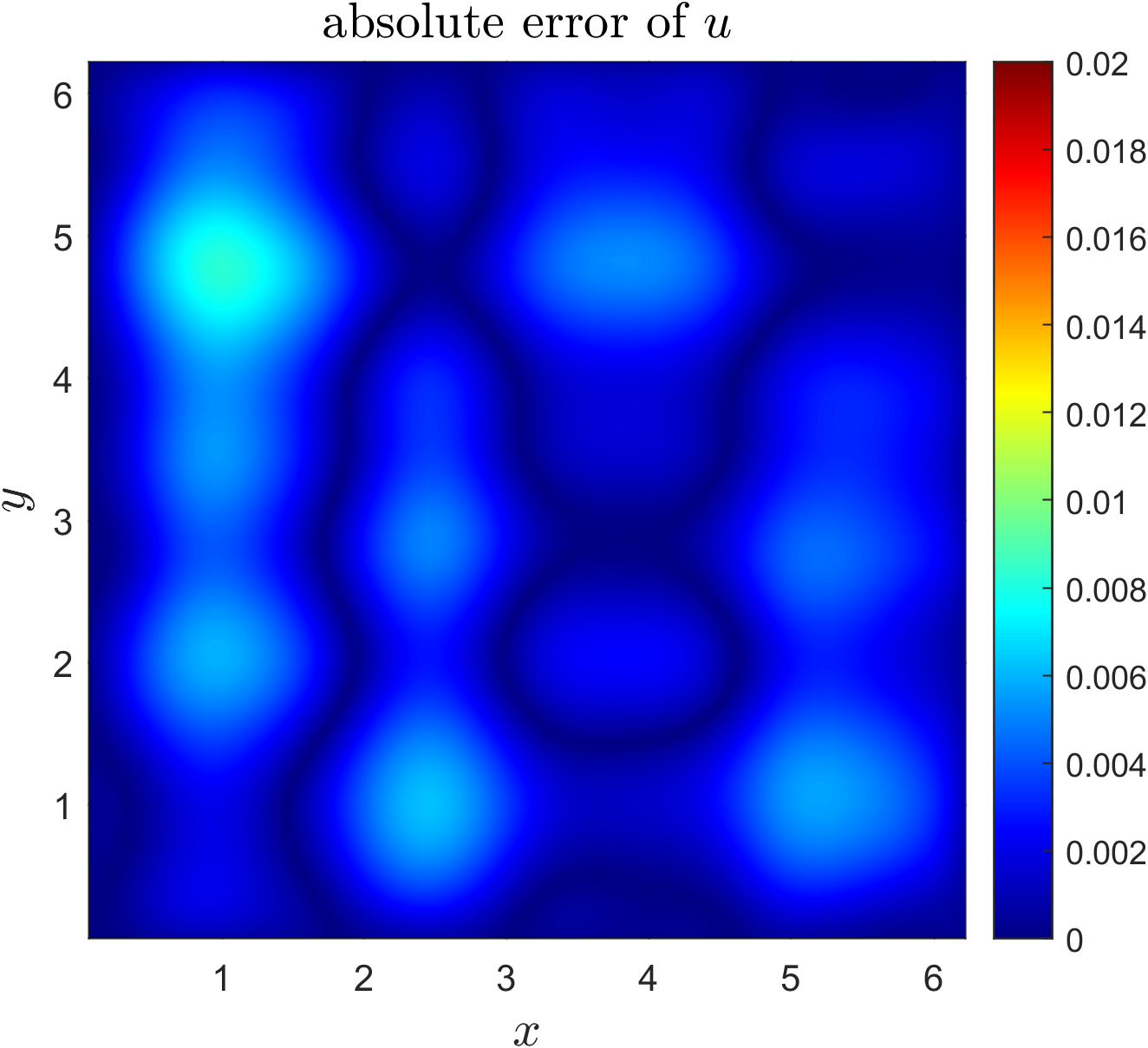}
        \includegraphics[scale=.25]{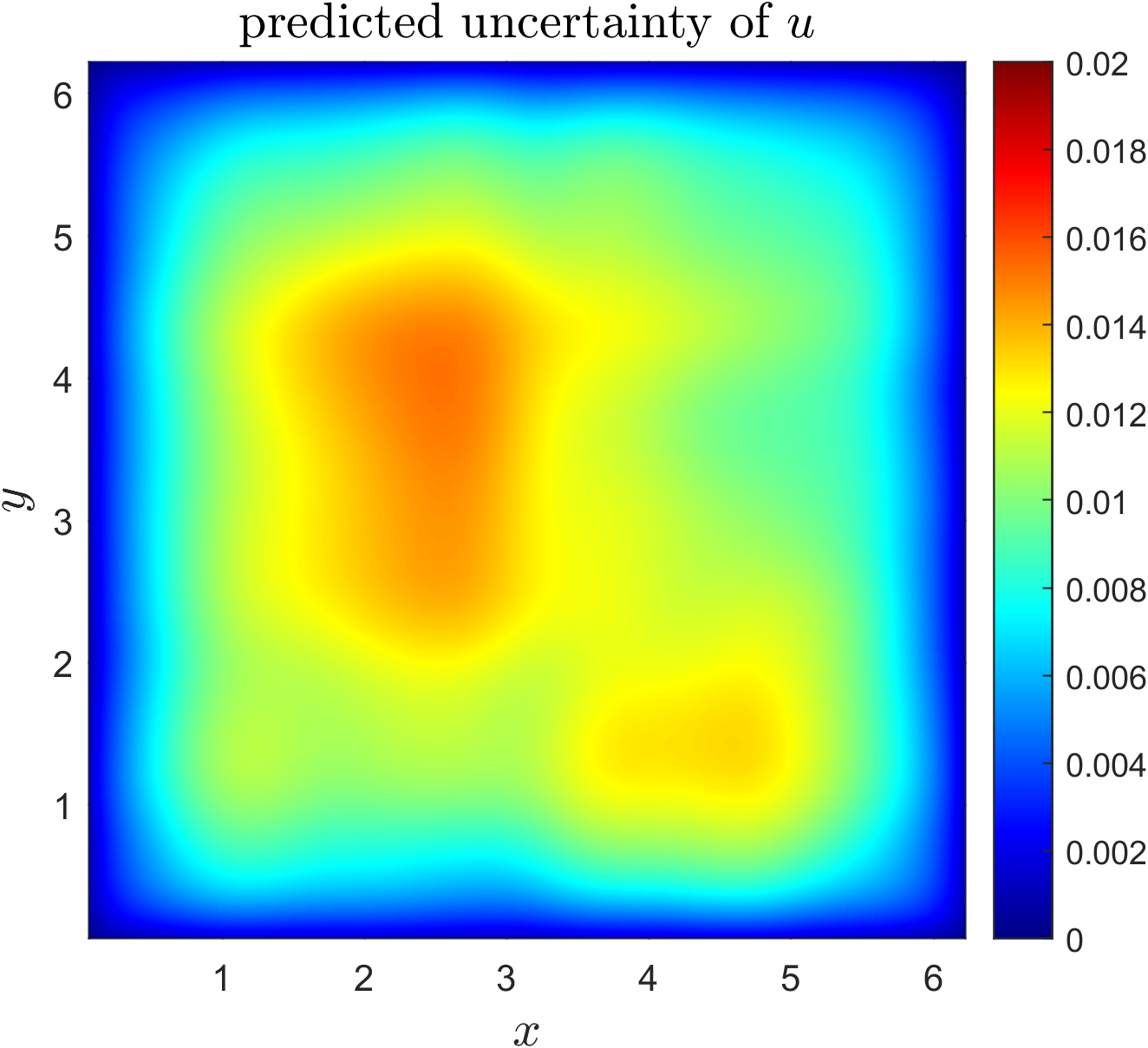}
    }
    \caption{Generator learning and few-shot physics-informed learning on the stochastic Helmholtz equation~\eqref{eq:helmholtz}. (a) Left: $1,000$ training samples of a slice of $f$ at $y=\pi$; middle: $1,000$ samples of a slice of $f$ at $y=\pi$ from the learned generator; right: statistics computed from samples. (b)/(c) Results for the downstream forward problem with $100$ random noisy measurements on $f$, using our approach with HMC. From left to right are reference, predicted mean, absolute error, and predicted uncertainty of $f$/$u$. Black crosses represent the locations of the measurements of $f$.}
    \label{fig:example_6:forward}
\end{figure}

For both the forward and inverse problems, $10,000$ $f$ are sampled from its distribution, and hence $10,000$ tasks are solved with MH-PINNs with boundary conditions hard-encoded in NN modeling. We display the samples of a slice of $f$ in Fig.~\ref{fig:example_6:forward}(a). For the forward problem, $\D_k$ only contains measurements of the source term $f_k$, i.e. $\D_k=\{\{(x_k^i, y_k^i),f_k^i\}_{i=1}^{N_k^f}\}$, while for the inverse problem $\D_k$ also contains measurements of the sought solution $u_k$: $\D_k=\{\{(x_k^i, y_k^i),f_k^i\}_{i=1}^{N_k^f}, \{(x_k^i, y_k^i),u_k^i\}_{i=1}^{N_k^u}\}$. For the training in the forward problem, each sample of $f$ is resolved by a $50\times50$ uniform mesh on 2-D spatial domain $(0, 2\pi)\times(0, 2\pi)$ with boundary excluded. For the inverse problem, the same $10,000$ samples of $f$ are used, but this time they are resolved with a $21\times21$ uniform mesh. In addition, for each task $\T_k$, measurements of $u_k$ on a $6\times6$ uniform mesh are available.
The reference solution and measurements of $u$ are generated by solving Eq.~\eqref{eq:helmholtz} with $\lambda_k^2 = \int_{[0, 2\pi]^2} f_k^2(x, y)dxdy$ using the finite difference method with five-point stencil.
For the downstream tasks, $100$ random measurements of $f$ are available for the forward problem, and $50$ random measurements of $f$ and $10$ random measurements of $u$ are available for the inverse problem. The noise is assumed to be independent additive Gaussian noise with $0.05$ noise scale. 

\begin{figure}[ht]
    \centering
    \subfigure[]{
        \includegraphics[scale=.25]{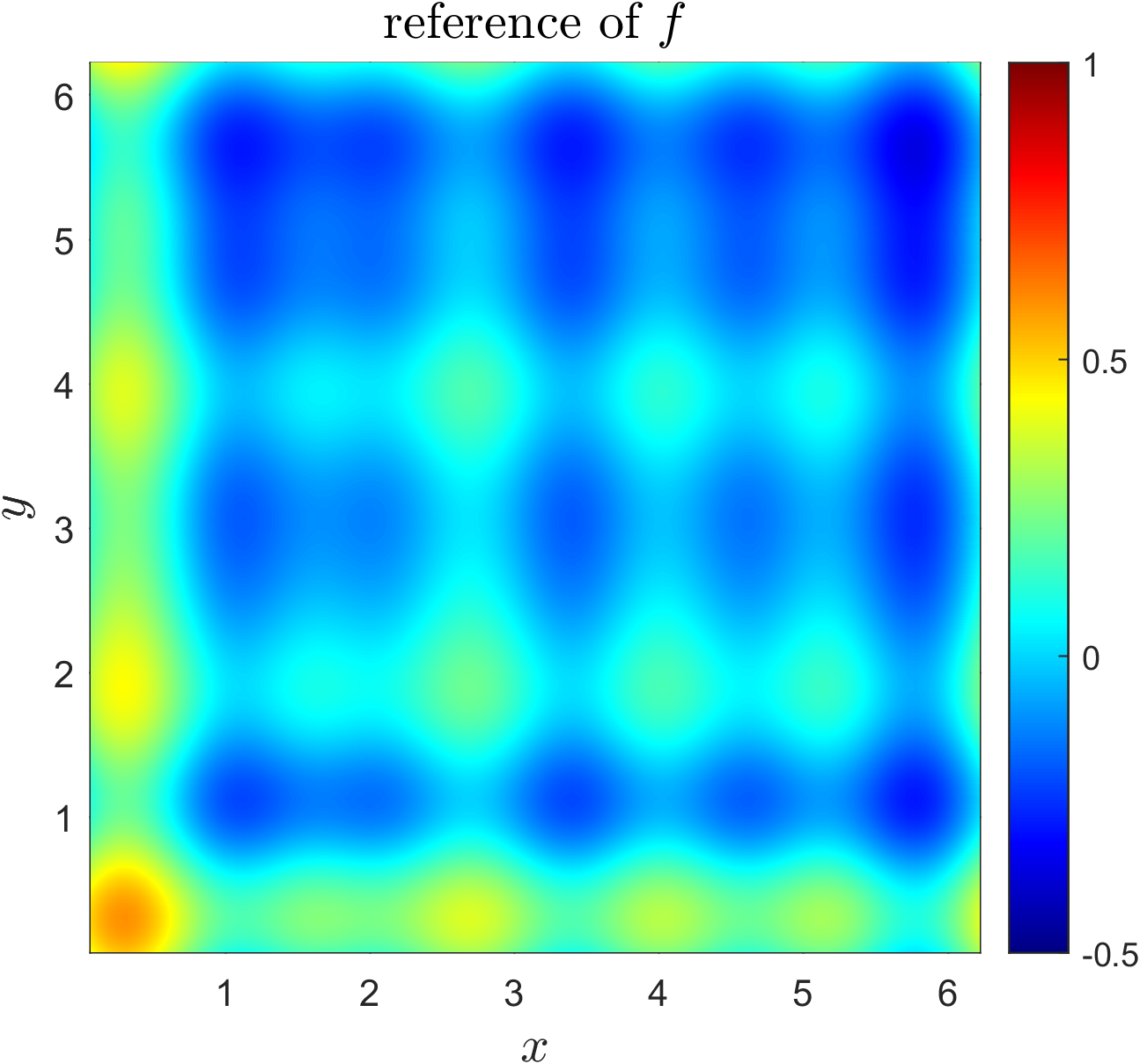}
        \includegraphics[scale=.25]{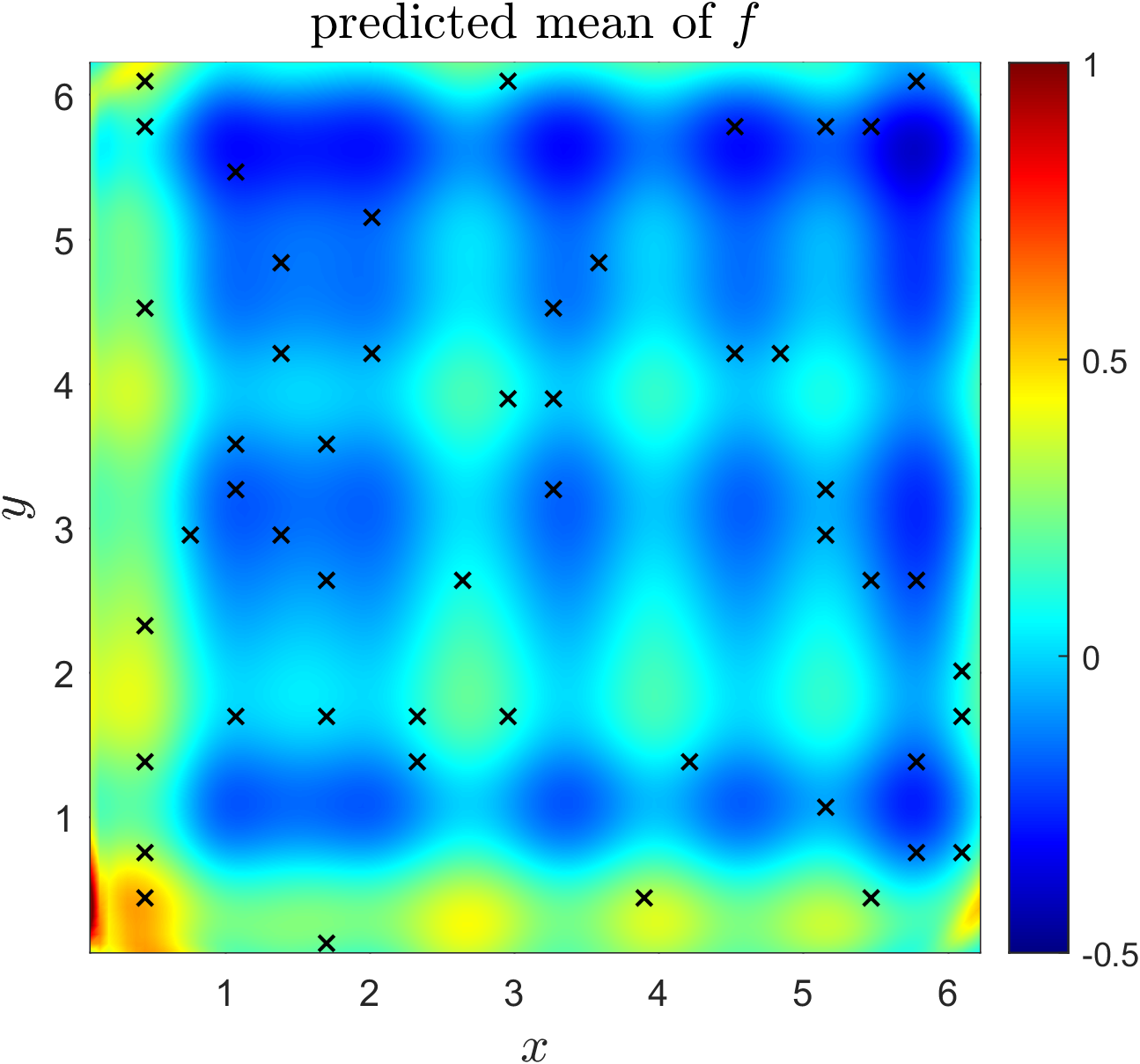}
        \includegraphics[scale=.25]{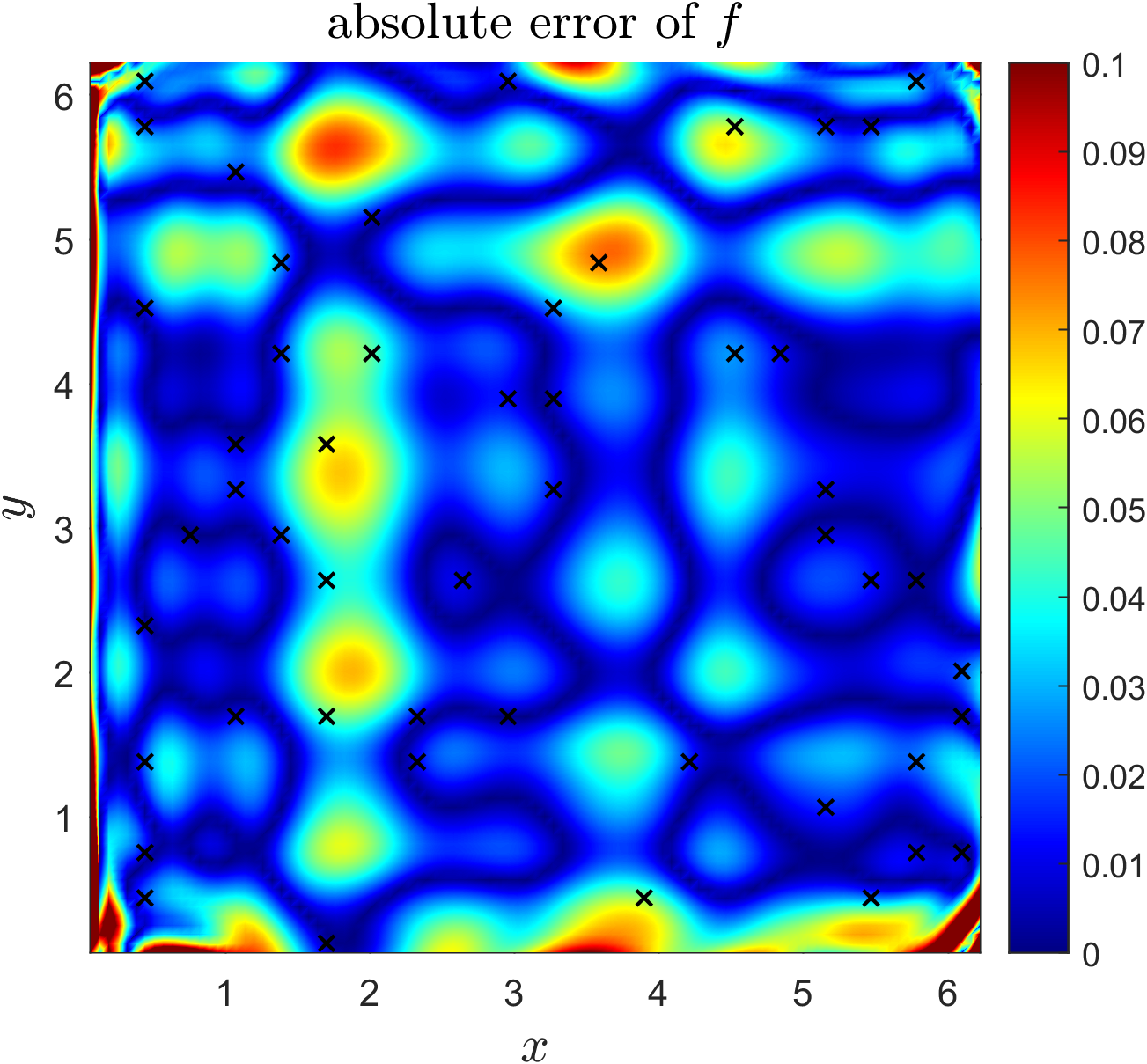}
        \includegraphics[scale=.25]{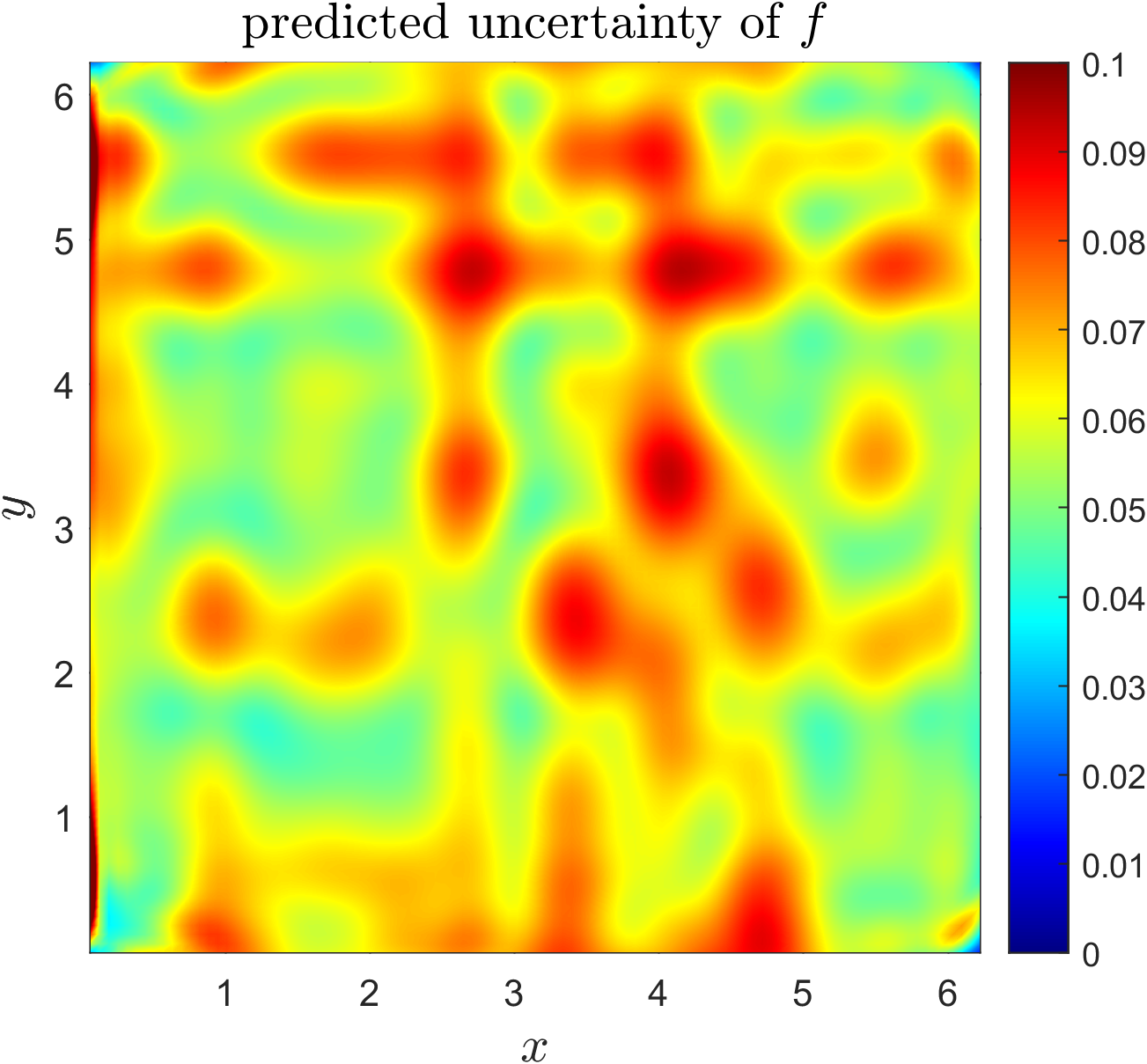}
    }
    \subfigure[]{
        \includegraphics[scale=.25]{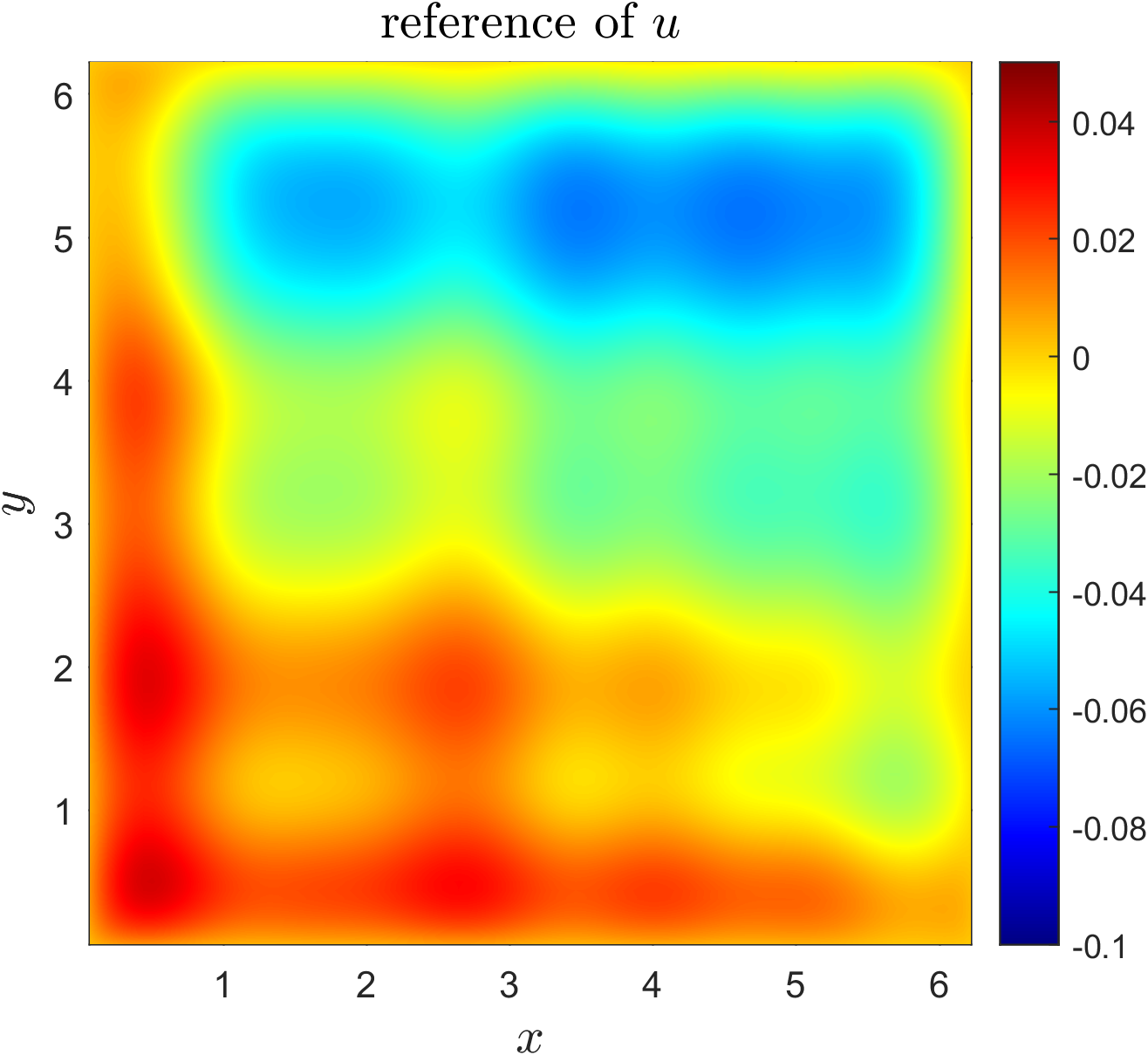}
        \includegraphics[scale=.25]{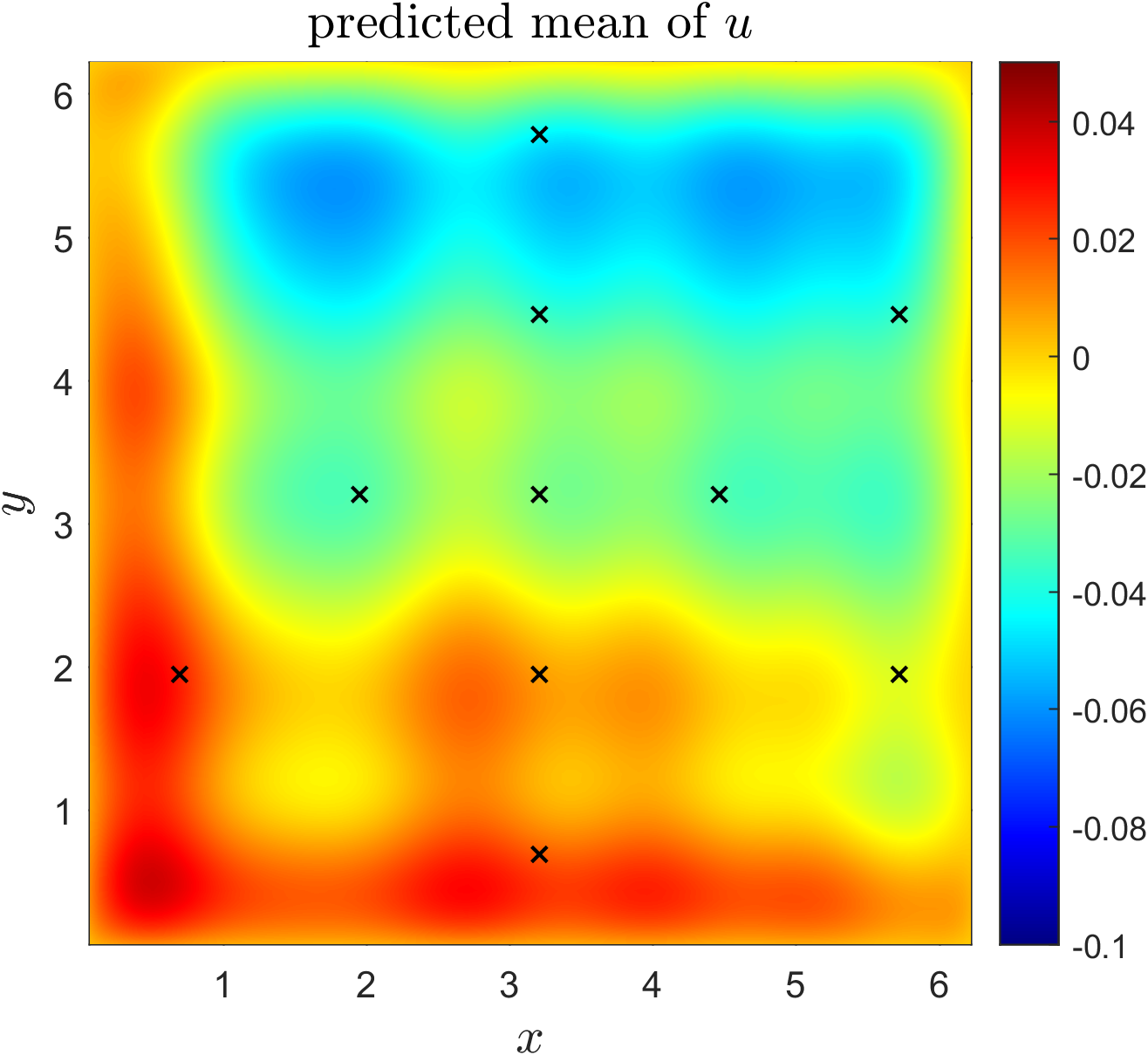}
        \includegraphics[scale=.25]{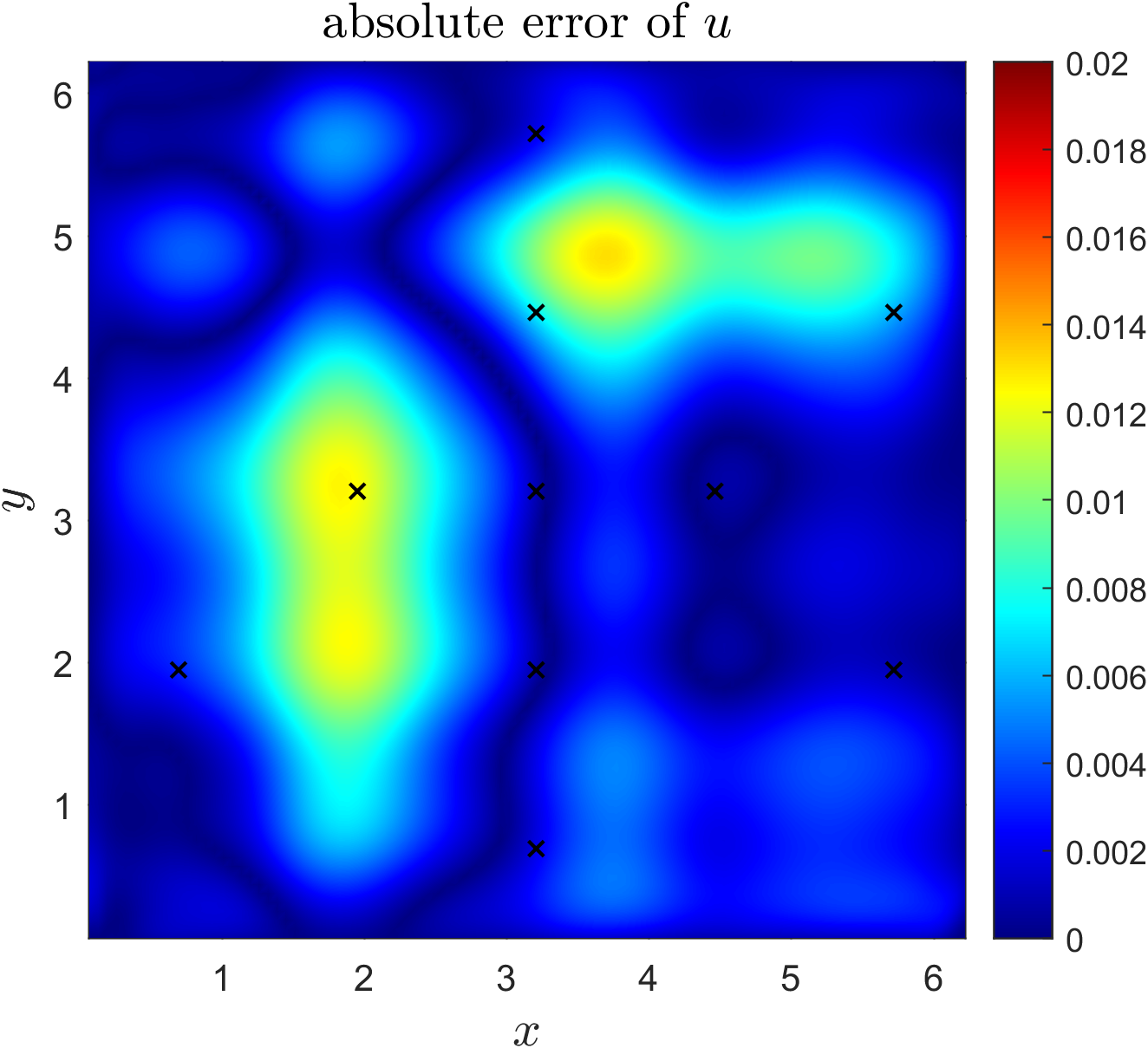}
        \includegraphics[scale=.25]{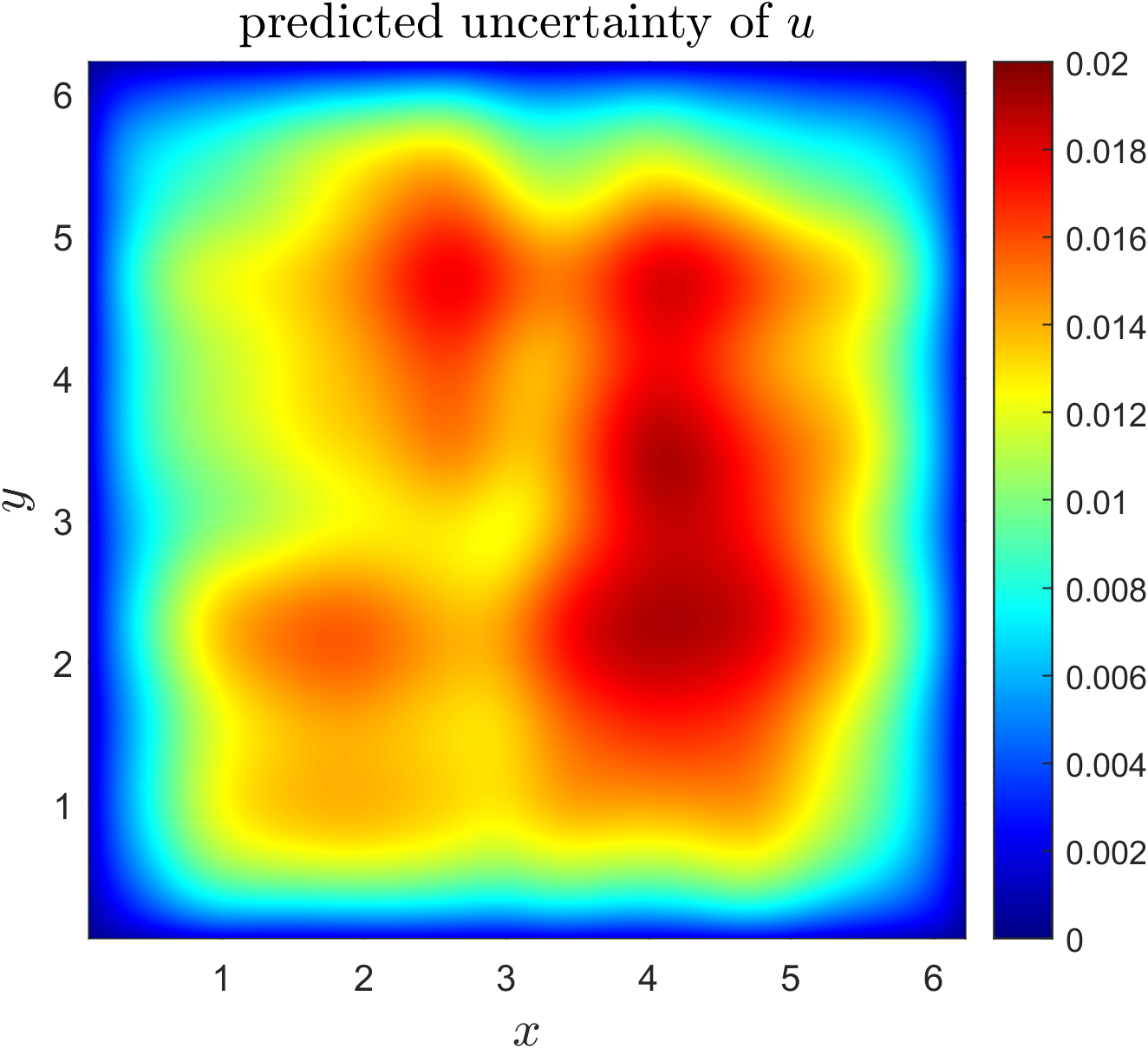}
    }
    \caption{Results for the downstream inverse problem on the stochastic Helmholtz equation~\eqref{eq:helmholtz}, with $50$ random noisy measurements of $f$ and $10$ random noisy measurements of $u$. $\lambda$ is estimated as $1.0785\pm0.0307$ in the format of predicted mean $\pm$ predicted standard deviation, while the reference value is $1.0042$. (a)/(b) From left to right are reference, predicted mean, absolute error, and predicted uncertainty of $f$/$u$. Black crosses represent locations of the measurements of $f$ or $u$.}
    \label{fig:example_6:inverse}
\end{figure}

Results are displayed in Tables~\ref{tab:example_6:forward} and~\ref{tab:example_6:inverse}, and Figs.~\ref{fig:example_6:forward} and~\ref{fig:example_6:inverse}. As shown, the learned generator is able to produce samples of $f$ with high quality as well as providing informative prior knowledge for the downstream tasks, in both the forward and inverse problems. As for the noisy case with Bayesian inference and UQ, the predicted means agree with the references and the absolute errors are mostly bounded by the predicted uncertainties. The effectiveness of our approach for few-shot physics-informed learning and the applicability to both deterministic optimization and Bayesian inference have been consistently demonstrated in the past five examples.

\begin{table}[ht]
    \footnotesize
    \centering
    \begin{tabular}{|c|c|c|c|}
    \hline
         & PINN & MH-PINN \\
         \hline
       Error (\%) & $21.14$ & $1.12$  \\
       \hline
    \end{tabular}
    \caption{$L_2$ relative error of $u$ for the downstream forward problem on Eq.~\eqref{eq:helmholtz} with clean data of $f$, using our approach and the PINN method.}
\label{tab:example_6:forward}
\end{table}

\begin{table}[ht]
    \footnotesize
    \centering
    \begin{tabular}{|c|c|c|}
    \hline
         & PINN & MH-PINN \\
         \hline
       $\lambda$  & $1.9328$ & $1.0170$ \\
       \hline
       Error (\%) & $59.92$ & $2.58$\\
       \hline
    \end{tabular}
    \caption{Estimate of $\lambda$ and $L_2$ relative error of $u$ for the downstream inverse problem on Eq.~\eqref{eq:helmholtz} with clean data. The reference value of $\lambda$ is $1.0042$.}
\label{tab:example_6:inverse}
\end{table}

\section{Multi-task learning with multi-head neural networks}
\label{sec:MTL}
So far we have mostly focused on using MH-NNs together with NFs to estimate stochastic generators and learn informative prior knowledge from $\{\T_k\}_{k=1}^M$. This was achieved by first training MH-NNs in a MTL fashion and then training NFs to estimate the PDF of the head. Intuitively, the capability of MH-NNs when trained in MTL in capturing shared information is the key to the success in generative modeling and few-shot learning. For physics-informed MTL with MH-PINNs, ODEs/PDEs are solved simultaneously, and assuming the solutions to share the same set of basis functions gives us samples of the set of coefficients, which enables the generative modeling, followed by few-shot learning, which is the whole point of the method proposed in this paper. However, the cost and/or the benefit of imposing the same set of basis functions to all solutions have not been explicitly discussed yet.
On one hand, the shared body relates the training of tasks, which may be helpful if tasks are similar in certain ways. On the other hand, forcing all solutions to share the same basis functions may also be harmful when they behave differently. In particular, for tasks with sufficient data and physics, forcing them to share the same body with all other tasks may act as a negative regularization, and single-task learning (STL) may outperform MTL in terms of the prediction accuracy in those specific tasks. In this section, we investigate the effect of MTL using MH-NNs and provide preliminary results and analysis by revisiting the simple function approximation example in Sec.~\ref{subsec:example_1}, which, hopefully, could provide useful information and insight for future more rigorous research.

\subsection{Basis function learning and synergistic learning}

As discussed before, the quality and behavior of basis functions learned in MTL are crucial to generative modeling and learning the relation and the representative information of tasks $\{\T_k\}_{k=1}^M$. We consistently noticed from numerical examples that the initialization of the head in MH-NNs has great impact on the average accuracy of MTL, the learning of the basis functions, and the distribution of the head. Here, we test three initialization strategies, random normal method with $0.05$ standard deviation referred to as \textit{RN (0.05)}, Glorot uniform method \cite{glorot2010understanding} referred to as \textit{GU}, and random normal method with $1$ standard deviation referred to as \textit{RN (1)}.  
In the downstream few-shot learning tasks, we fine-tune the head without the learned PDF, which is in fact the TL method from \cite{desai2021one}, by which the information from the distribution of the head is excluded and the prediction accuracy is fully determined by the level of prior knowledge contained in the basis functions.

\begin{figure}[ht]
    \centering
    \subfigure[]{
        \includegraphics[scale=.25]{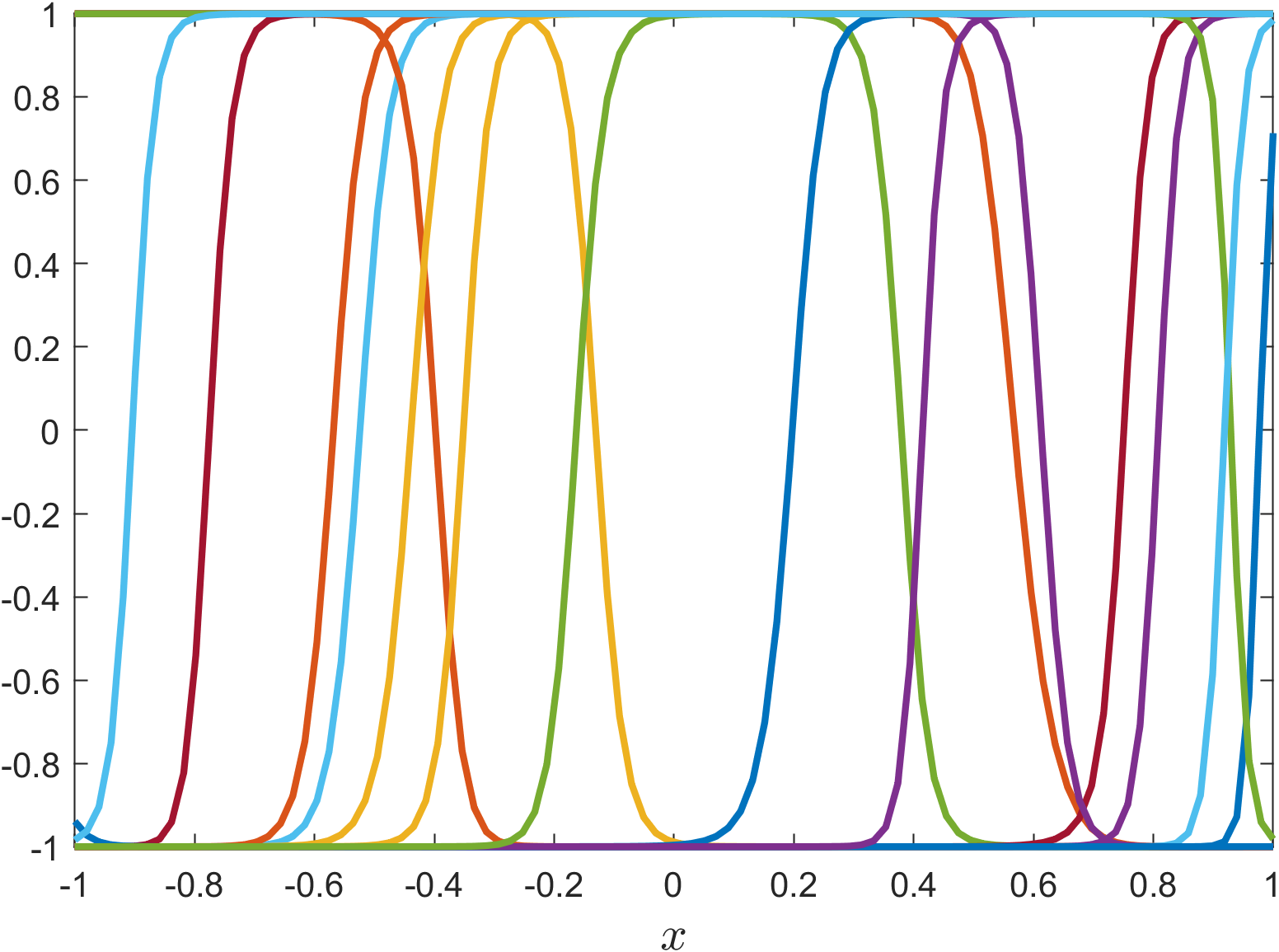}
        \includegraphics[scale=.25]{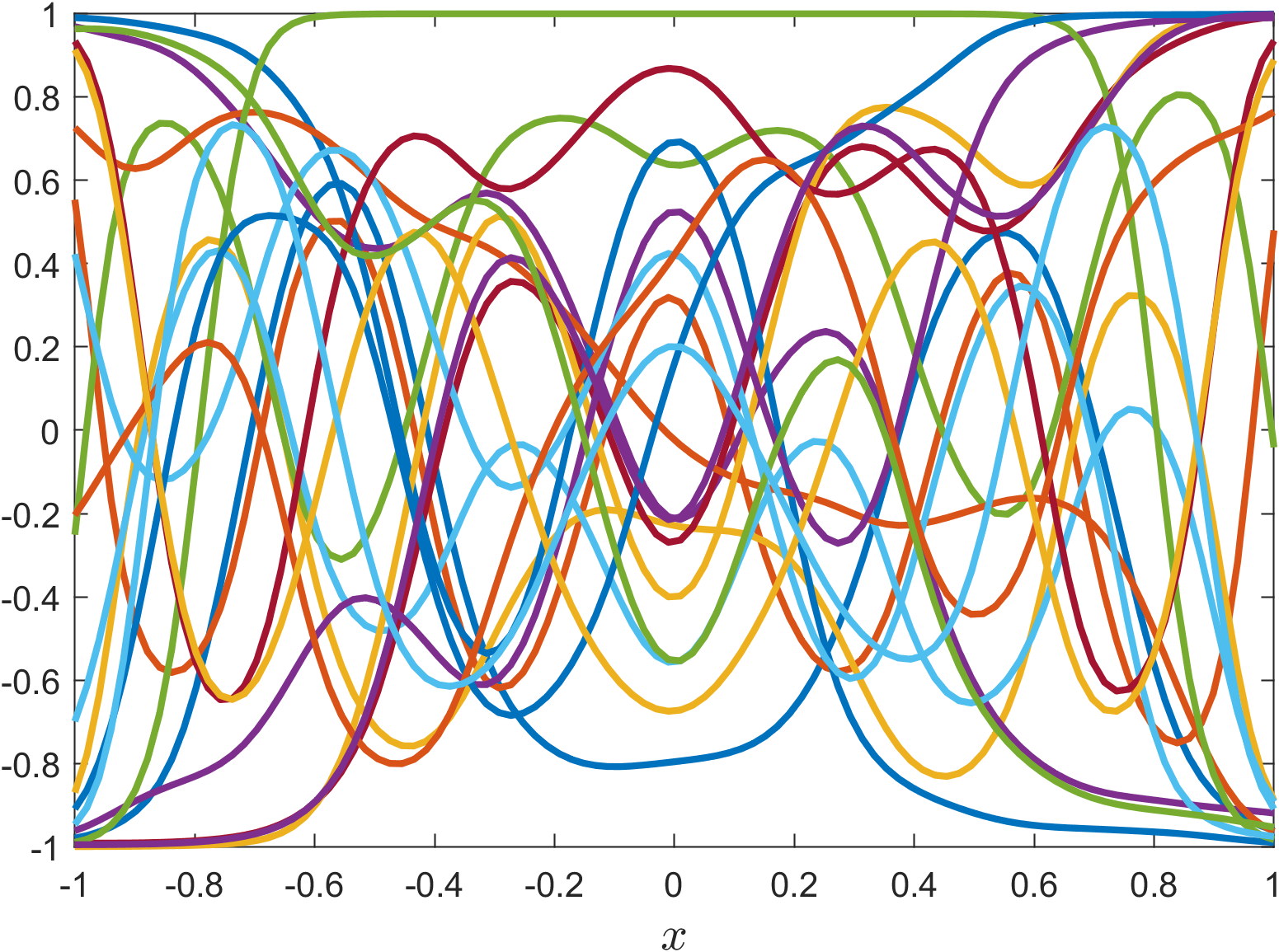}
        \includegraphics[scale=.25]{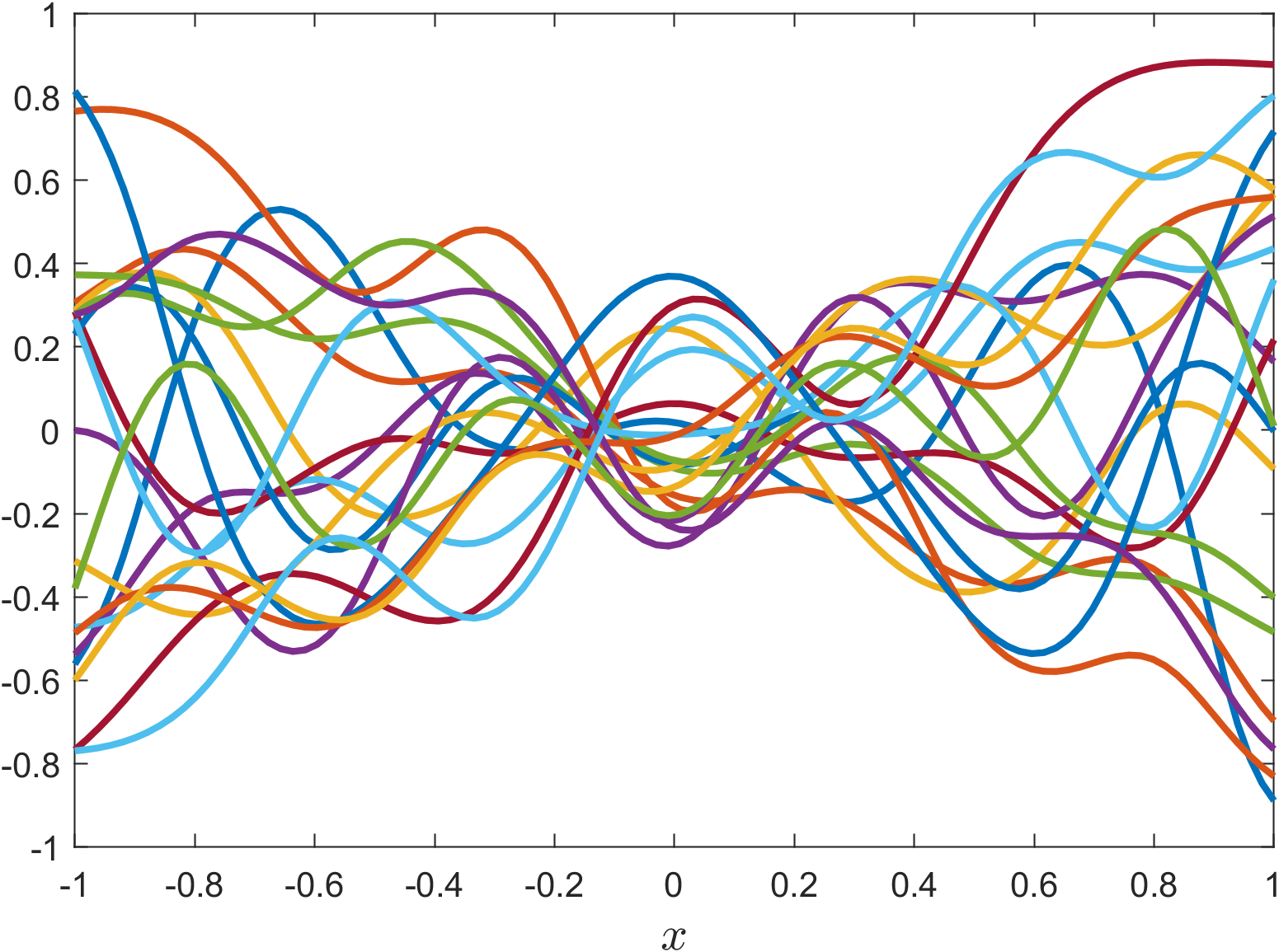}
    }
    \subfigure[]{
        \includegraphics[scale=.25]{figures/example_1/samples_trained.png}
    }
    \subfigure[]{
        \includegraphics[scale=.25]{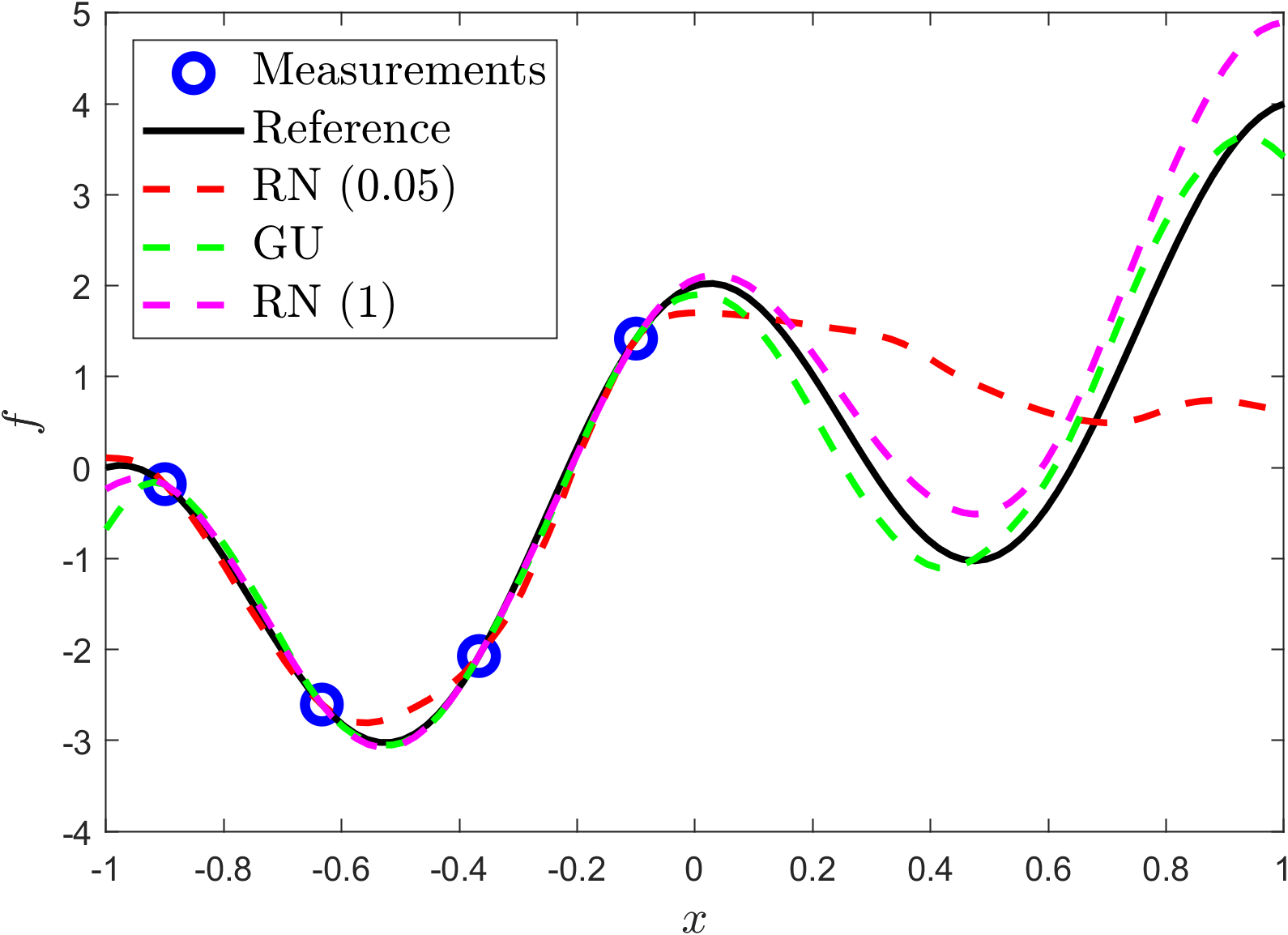}
        \includegraphics[scale=.25]{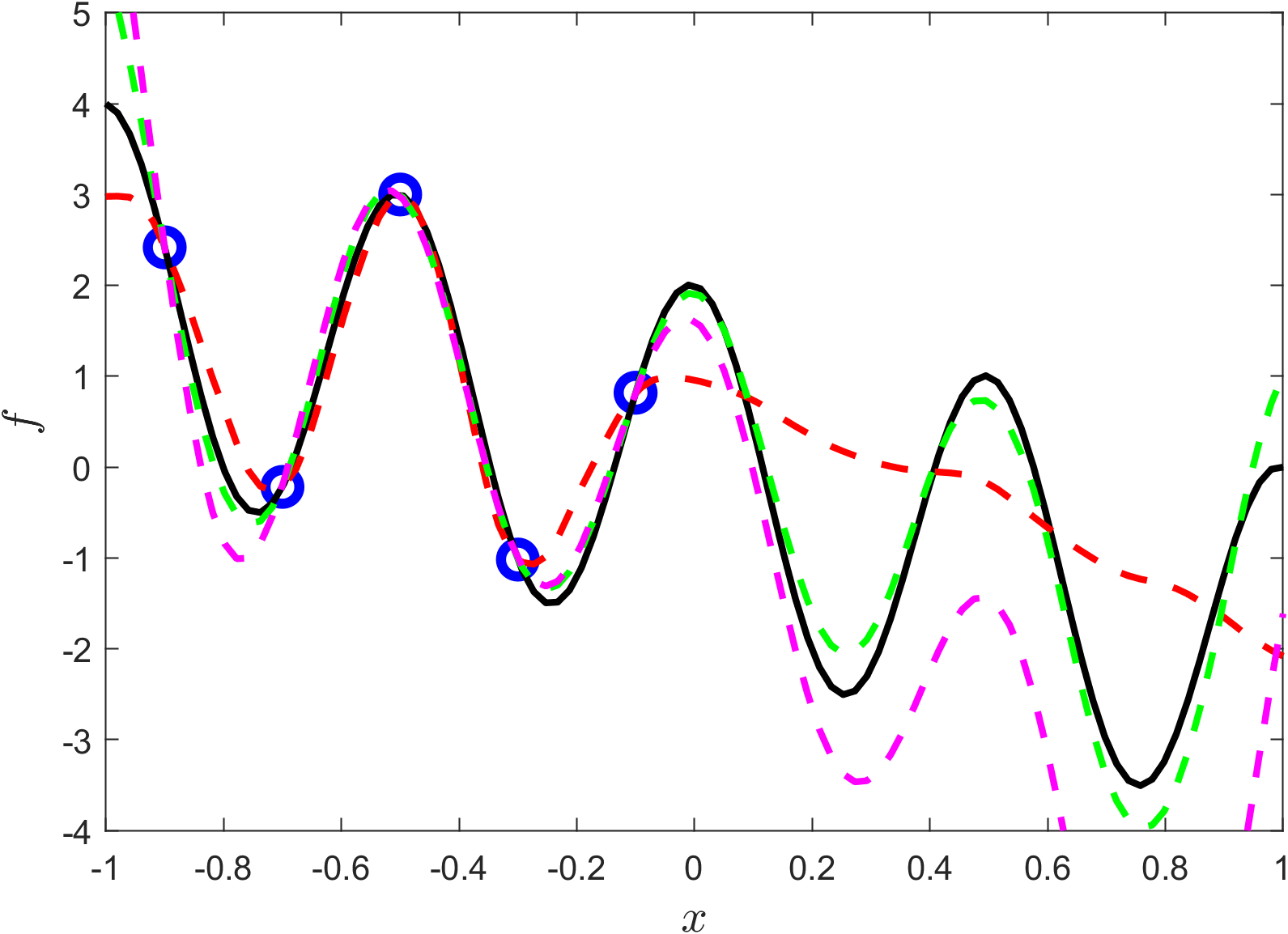}
    }
    \subfigure[]{
        \includegraphics[scale=.25]{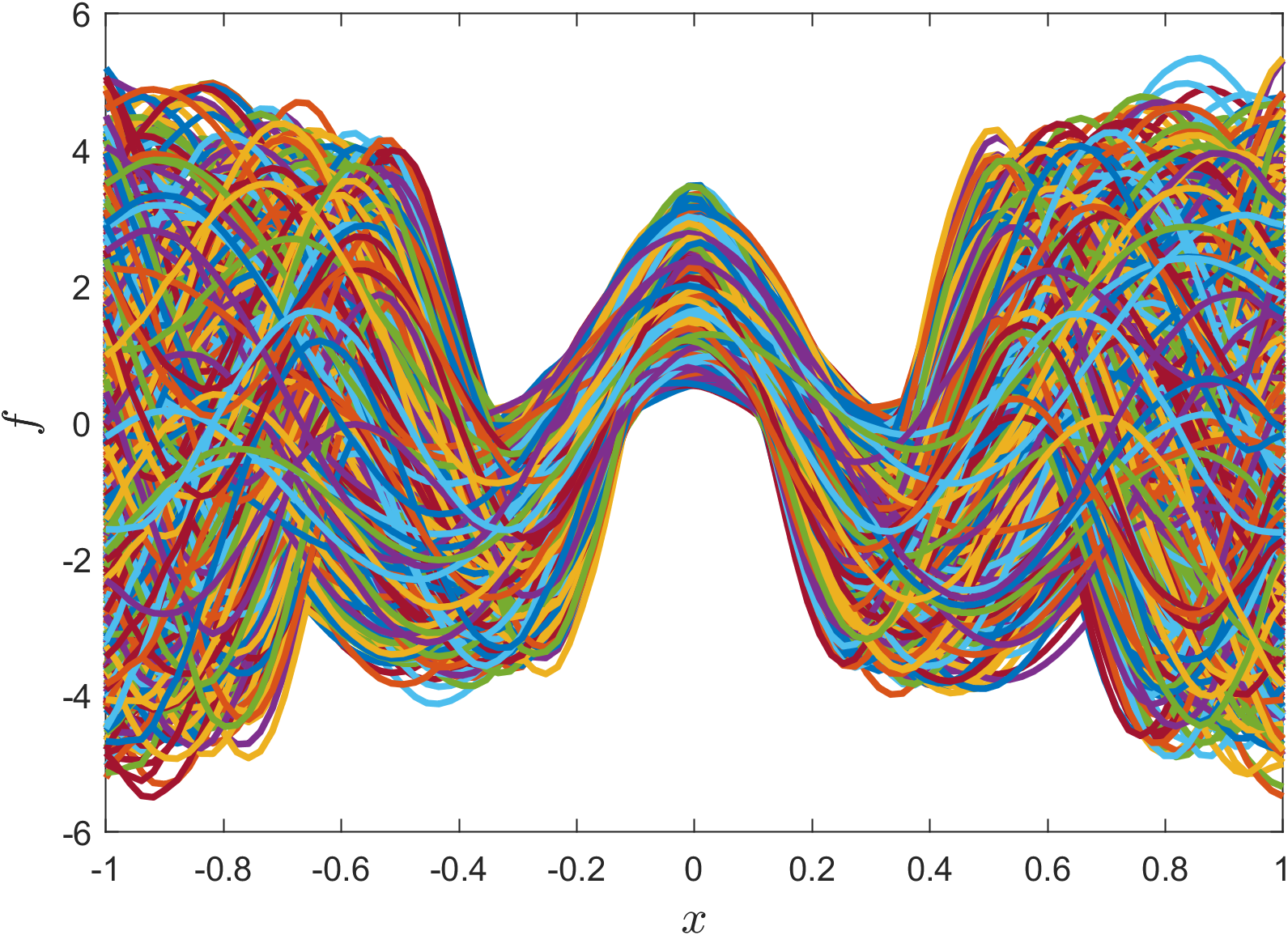}
        \includegraphics[scale=.25]{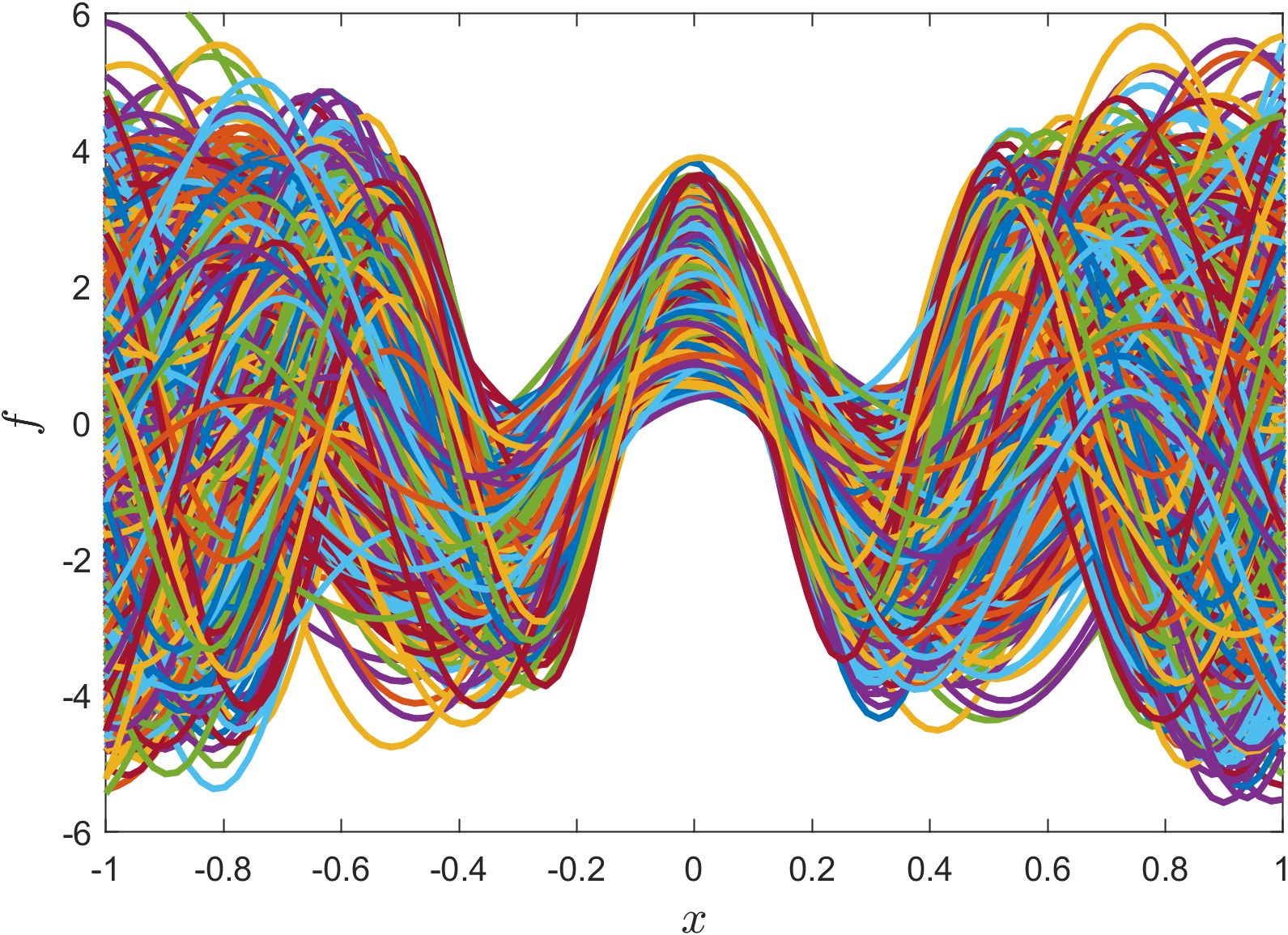}
        \includegraphics[scale=.25]{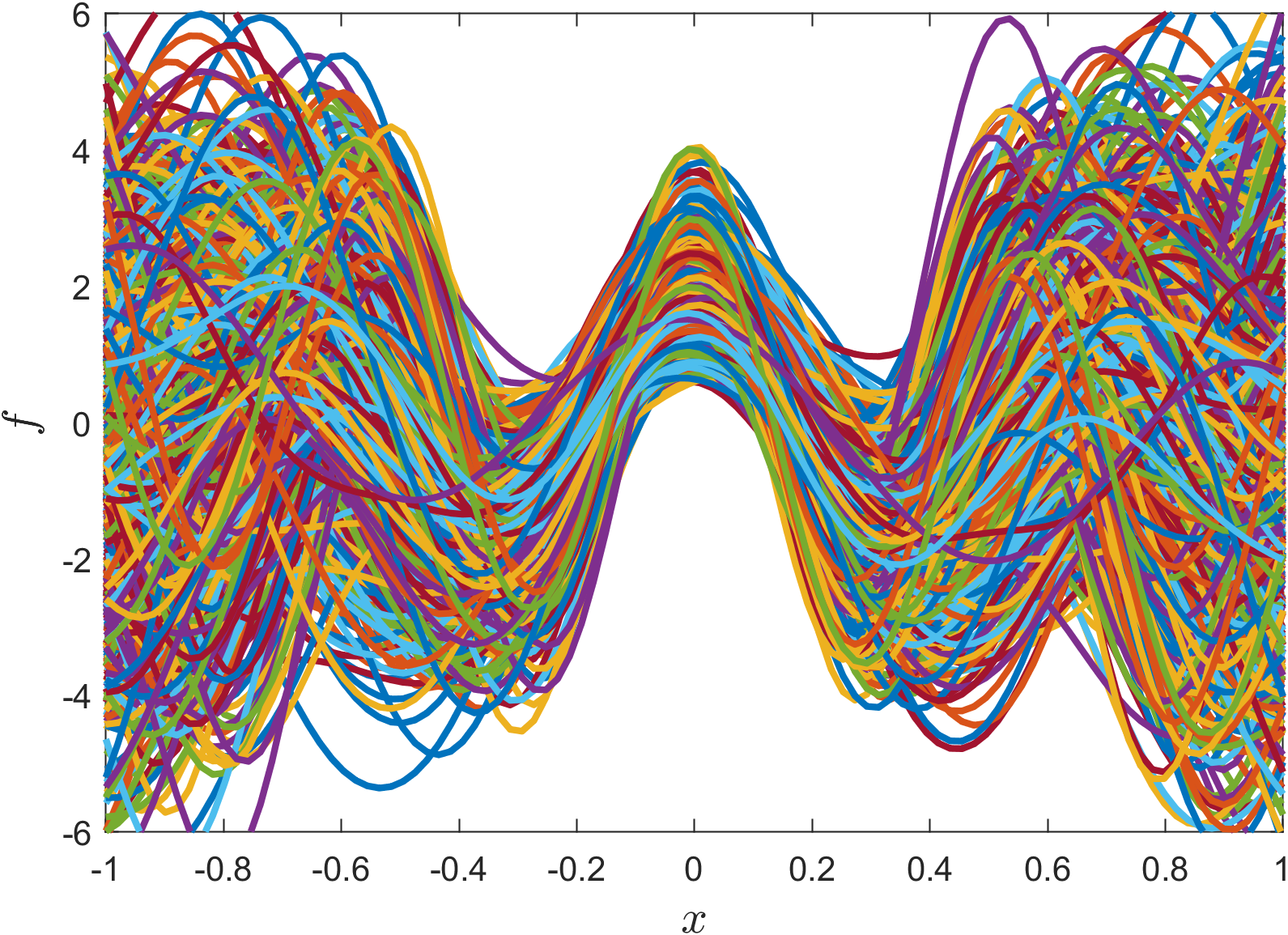}
    }
    \caption{The effect of different initialization methods of the head, in basis functions learning, few-shot learning, and generator learning. (a) Samples of $20$ basis functions from MH-NNs, trained for approximating $1,000$ $f$ generated from Eq.~\eqref{eq:example_1}, using, from left to right, RN ($0.05$), GU and RN ($1$) initialization methods. (b) $1,000$ training samples of $f$. (c) Results for two downstream few-shot regression tasks, using TL method without regularization informed by the learned PDF, as opposite to the proposed approach. (d) Results for generator learning, using, from left to right, RN ($0.05$), GU and RN ($1$) initialization methods.}
    \label{fig:MTL:basis_functions}
\end{figure}

As shown in Fig.~\ref{fig:MTL:basis_functions}, RN ($0.05$) yields the least informative basis functions, whose behavior is dominated by the hyperbolic tangent activation function of NNs. This is further demonstrated in the downstream few-shot learning tasks using the TL method. It also provides the worst prediction accuracy on average in MTL, as presented in Table~\ref{tab:MTL:MTL_STL}. GN and RN ($1$) perform similarly. Plots of some basis functions seemingly indicate that RN ($1$) yields better basis functions, whose behaviors are more similar to the family of functions displayed in Fig.~\ref{fig:MTL:basis_functions}(b), which, however, does not necessarily imply richer prior knowledge in the downstream tasks, as shown in Fig.~\ref{fig:MTL:basis_functions}(c).

It is shown empirically that compared to other two initialization strategies, MH-NNs with RN ($0.05$) does not deliver accurate MTL nor synergistic learning in basis functions. However, we noticed that, in generative modeling, it performs significantly better in terms of accuracy and convergence speed. As shown in Fig.~\ref{fig:MTL:basis_functions}(d), samples from the learned generator are of higher quality. We consistently found that initializing heads with relatively small values often led to easy and fast training of NFs and accurate learning of the generative models. We conjecture that this happens because MH-NNs in MTL tend to contain the representative and informative information in the heads when heads are initialized with small values, while contain it in the basis functions when heads are initialized with relatively large values.
 
\begin{table}[ht]
    \footnotesize
    \centering
    \begin{tabular}{|c|c|c|c|}
    \hline
         & RN ($0.05$) & GU & RN ($1$) \\
         \hline
       Error (\%) & $0.8373\pm0.2341$ & $0.1907\pm0.0690$ & $0.3131\pm 0.0937$  \\
       \hline
    \end{tabular}
    \caption{$L_2$ relative errors, from MTL, for $1,000$ tasks, using different initialization methods. The errors are displayed in the format of mean $\pm$ standard deviation, computed over all tasks.}
\label{tab:MTL:basis_functions}
\end{table}

\subsection{Multi-Task Learning (MTL) versus Single-Task Learning (STL)}

As discussed earlier, MTL with MH-NNs does not necessarily result in synergistic learning nor higher accuracy for all tasks on average. Here, we use again the function approximation example in Sec.~\ref{subsec:example_1}, to investigate the effectiveness of MTL with MH-NNs, as compared to STL. The first case we consider here is assuming that the data is sufficient. For that, we randomly choose $100$ out of the $1,000$ training samples, each one of which is approximated by a NN trained independently, and compare the results with MH-NNs in terms of prediction accuracy. Note that in this case, a MH-NN is trained on $1,000$ functions as before and tested on the chosen $100$ functions, while a single-head NN with the same architecture is trained on $100$ functions directly. Results are shown in Table~\ref{tab:MTL:MTL_STL}, from which it is verified empirically that MTL is outperformed by STL under certain circumstances, e.g., when the random normal initialization methods are used. 

\begin{table}[ht]
    \footnotesize
    \centering
    \begin{tabular}{|c|c|c|c|c|}
    \hline
         & RN ($0.05$) & GU & RN ($1$) & STL\\
         \hline
       Error (\%) & $0.7575\pm0.2477$ & $0.1362\pm0.0259$ & $0.3664\pm 0.1031$ & $0.2102\pm0.0794$  \\
       \hline
    \end{tabular}
    \caption{$L_2$ relative errors of $f$, from MTL with MH-NNs and STL with NNs, on $100$ tasks. Different initialization methods are used for the heads in MH-NNs. The errors are displayed in the format of mean $\pm$ standard deviation, computed over all $100$ tasks.}
\label{tab:MTL:MTL_STL}
\end{table}

\begin{figure}[ht]
    \centering
    \includegraphics[scale=.25]{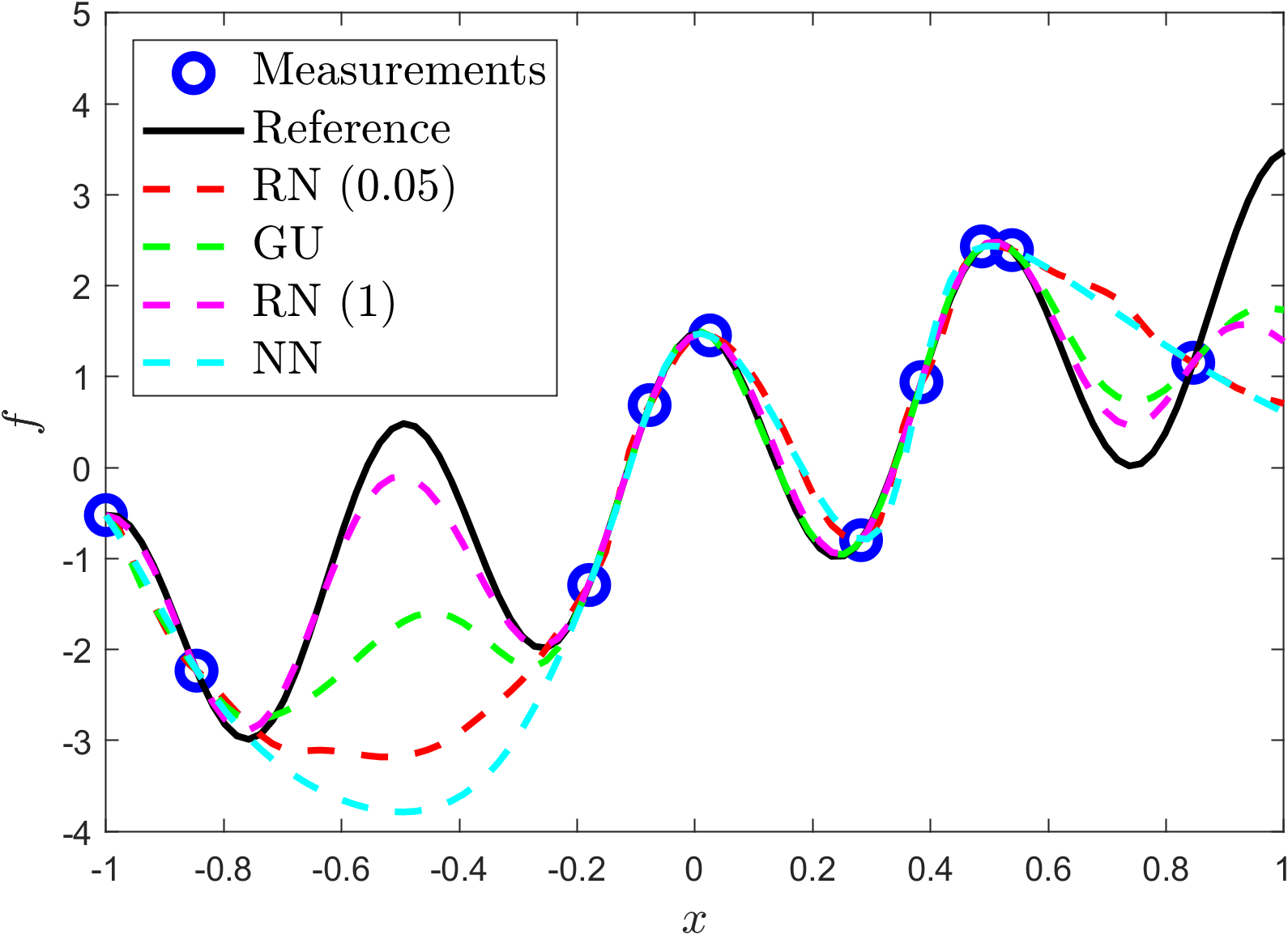}
    \includegraphics[scale=.25]{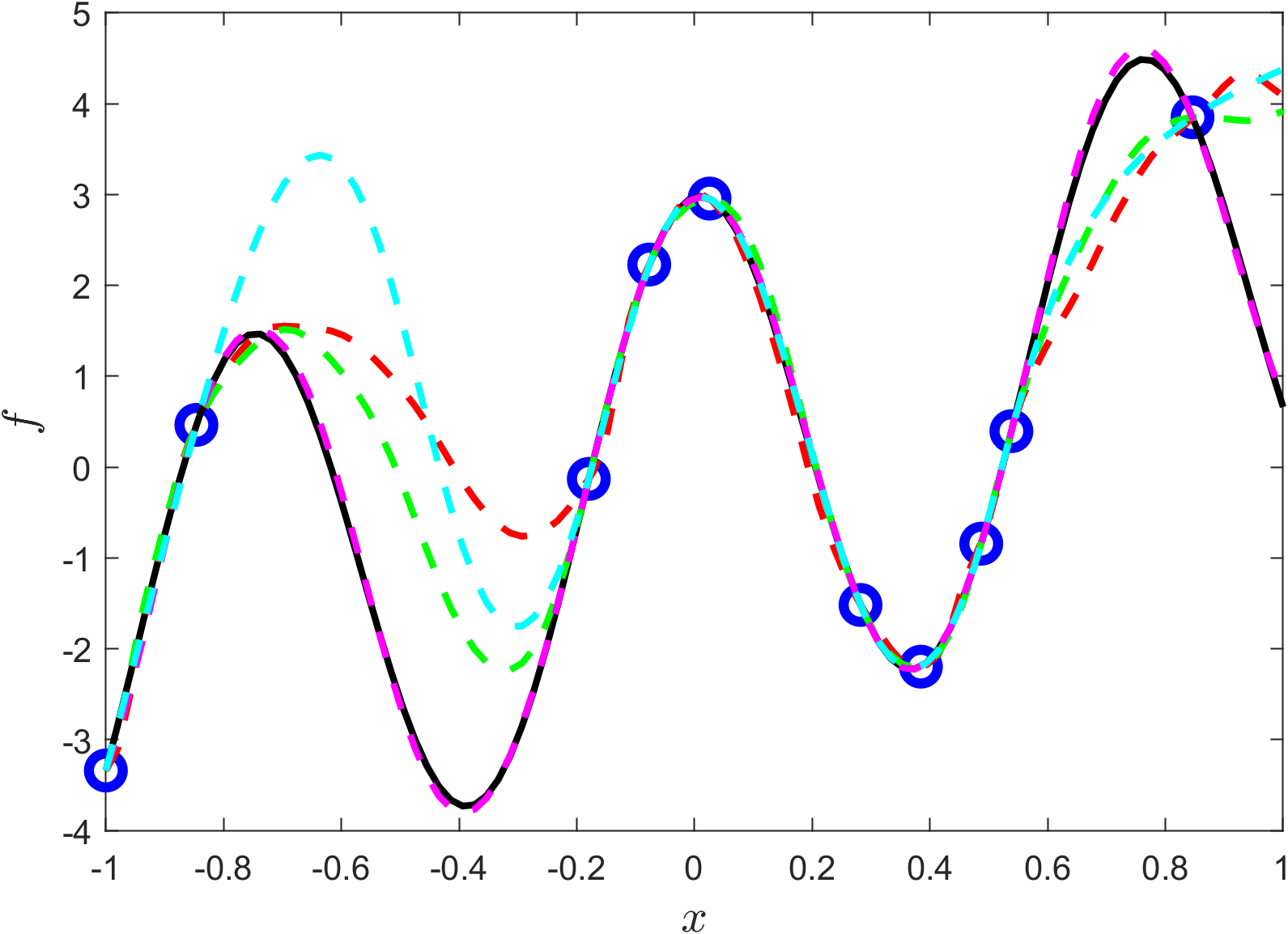}
    \includegraphics[scale=.25]{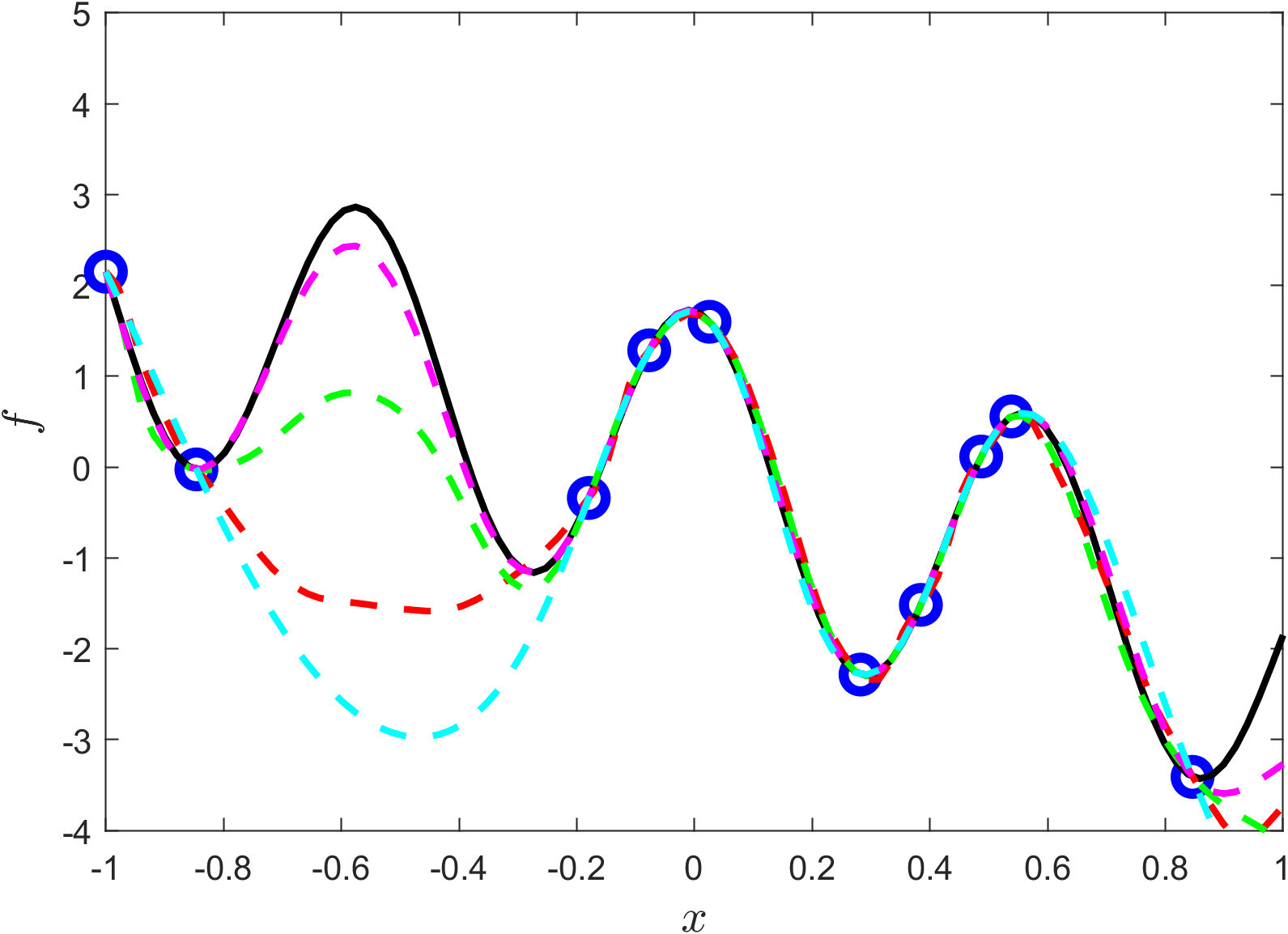}
    \caption{Results for 3 tasks with insufficient data from MTL with MH-NNs, using different initialization methods over the head, and from STL with NNs with the same architecture. We note that tasks with sufficient data and tasks with insufficient data are treated equally in MTL. }
    \label{fig:MTL:MTL_STL_2}
\end{figure}

The second case we consider is assuming that the data is sufficient for some tasks while insufficient for other tasks. For that, we split equally the $1,000$ tasks into two subsets of tasks. For the first $500$ tasks, we assume we only have $10$ measurements randomly sampled on $[-1, 1]$, while for the other $500$ tasks, we assume we have full $40$ measurements equidistantly distributed on $[-1, 1]$. MTL with MH-NNs is performed on those $1,000$ regression tasks all at once, and the tasks are treated as equal. The results are presented in Table~\ref{tab:MTL:MTL_STL_2} and Fig.~\ref{fig:MTL:MTL_STL_2}. We can see that, compared to STL, MTL improves the prediction accuracy on tasks with insufficient data, providing empirical evidence of synergistic learning. Also, interestingly, RN ($1$) initialization method, which yields the worst generative models, performs the best among all three, which agrees with our previous conjecture on the basis functions learning with MH-NNs, that heads initialized with large values tend to force representative and informative information to be encoded in the basis functions.

\begin{table}[ht]
    \footnotesize
    \centering
    \begin{tabular}{|c|c|c|c|}
    \hline
         & RN ($0.05$) & GU & RN ($1$) \\
         \hline
       Error (\%) & $63.60\pm24.08$ & $40.49\pm20.49$ & $16.91\pm 11.08$  \\
       \hline
    \end{tabular}
    \caption{$L_2$ relative errors of $f$, from MTL with MH-NNs, on $500$ tasks equipped with insufficient data. The errors are displayed in the format of mean $\pm$ standard deviation, computed over all $500$ tasks.}
\label{tab:MTL:MTL_STL_2}
\end{table}

\section{Discussion}
\label{sec:dicussion}

We have developed multi-head neural networks (MH-NNs) for physics-informed machine learning, and proposed multi-head physics-informed neural networks (MH-PINNs) as a new method, implemented in the L-HYDRA code. The primary focus of this work is on MH-NNs and MH-PINNs for various learning problems in scientific machine learning, including multi-task learning (MTL), stochastic processes approximation, and few-shot regression learning. We first formulated the problem in Eq.~\eqref{eq:problem}, introduced the architecture design of MH-PINNs, and proposed a method to transform MH-NNs and MH-PINNs to generative models with the help of normalizing flows (NFs) for density estimation and generative modeling. We then studied the applicability and capabilities of MH-PINNs in solving ordinary/paritial differential equations (ODEs/PDEs) as well as approximating stochastic processes. We completed the paper with preliminary and empirical explorations of MH-NNs in synergistic learning, and examined the potential benefits and cost of MTL with MH-NNs. 

This paper can be used in various ways: it proposes a NN approach for MTL in solving ODEs/PDEs; it provides a new approach to approximate stochastic processes; it presents a method to address few-shot physics-informed learning problems, which are often encountered in the context of meta-learning and transfer learning; it contains a systematic study of applying MH-NNs to scientific computing problems; it presents the first empirical evidence of synergistic learning. 

However, there are a few major problems on MH-NNs we did not address, one of which is the expressivity of MH-NNs, or more generally hard-parameter sharing NNs in approximating complicated stochastic processes. Intuitively, if two functions behave very differently, forcing them to share the same basis functions would affect adversely the approximation accuracy. The second problem is the balancing issue of different terms in the loss function in MTL. It is shown in the literature \cite{mcclenny2020self} that PINNs, trained in single-task learning, are already deeply influenced by the weights in front of different terms in the loss function, e.g., data loss, boundary condition loss, PDE residual loss. This issue may be more complex in training MH-PINNs, because in MTL the loss function is commonly defined as weighted summation of task-specific loss. The last major problem is MH-PINNs for synergistic learning. In this paper, we only studied one example in function approximation and presented empirical evidence. More work for the understanding of synergistic learning with MH-PINNs along both the theoretical and computational directions should be pursued in the future.


\section*{Acknowledgments}
We would like to thank Professor Xuhui Meng of Huazhong University of Science and Technology for helpful discussions. This work was supported by: the Vannevar Bush Faculty Fellowship award (GEK) from ONR (N00014-22-1-2795); the U.S. Department of Energy, Advanced Scientific Computing Research program, under the Scalable, Efficient and Accelerated Causal Reasoning Operators, Graphs and Spikes for Earth and Embedded Systems (SEA-CROGS) project, DE-SC0023191; and by the MURI/AFOSR FA9550-20-1-0358 project.

\bibliographystyle{siamplain}
\bibliography{references}

\begin{thebibliography}{10}

\bibitem{tensorflow2015-whitepaper}
{\sc M.~Abadi, A.~Agarwal, P.~Barham, and et. al.}, {\em {TensorFlow}:
  Large-scale machine learning on heterogeneous systems}, 2015,
  \url{https://www.tensorflow.org/}.
\newblock Software available from tensorflow.org.

\bibitem{ablowitz1979explicit}
{\sc M.~J. Ablowitz and A.~Zeppetella}, {\em Explicit solutions of {Fisher's}
  equation for a special wave speed}, Bulletin of Mathematical Biology, 41
  (1979), pp.~835--840.

\bibitem{bahmani2021training}
{\sc B.~Bahmani and W.~Sun}, {\em Training multi-objective/multi-task
  collocation physics-informed neural network with student/teachers transfer
  learnings}, arXiv preprint arXiv:2107.11496,  (2021).

\bibitem{bhattacharya2020model}
{\sc K.~Bhattacharya, B.~Hosseini, N.~B. Kovachki, and A.~M. Stuart}, {\em
  Model reduction and neural networks for parametric {PDEs}}, arXiv preprint
  arXiv:2005.03180,  (2020).

\bibitem{Caruana1993MultitaskLA}
{\sc R.~Caruana}, {\em {Multitask Learning: A Knowledge-Based Source of
  Inductive Bias}}, in International Conference on Machine Learning, 1993.

\bibitem{chen2021learning}
{\sc X.~Chen, J.~Duan, and G.~E. Karniadakis}, {\em Learning and meta-learning
  of stochastic advection--diffusion--reaction systems from sparse
  measurements}, European Journal of Applied Mathematics, 32 (2021),
  pp.~397--420.

\bibitem{desai2021one}
{\sc S.~Desai, M.~Mattheakis, H.~Joy, P.~Protopapas, and S.~Roberts}, {\em
  One-shot transfer learning of physics-informed neural networks}, arXiv
  preprint arXiv:2110.11286,  (2021).

\bibitem{dillon2017tensorflow}
{\sc J.~V. Dillon, I.~Langmore, D.~Tran, E.~Brevdo, S.~Vasudevan, D.~Moore,
  B.~Patton, A.~Alemi, M.~Hoffman, and R.~A. Saurous}, {\em {TensorFlow
  Distributions}}, arXiv preprint arXiv:1711.10604,  (2017).

\bibitem{dinh2016density}
{\sc L.~Dinh, J.~Sohl-Dickstein, and S.~Bengio}, {\em {Density Estimation Using
  Real NVP}}, arXiv preprint arXiv:1605.08803,  (2016).

\bibitem{finn2017model}
{\sc C.~Finn, P.~Abbeel, and S.~Levine}, {\em Model-agnostic meta-learning for
  fast adaptation of deep networks}, in International Conference on Machine
  Learning, PMLR, 2017, pp.~1126--1135.

\bibitem{germain2015made}
{\sc M.~Germain, K.~Gregor, I.~Murray, and H.~Larochelle}, {\em {MADE}:
  {Masked} autoencoder for distribution estimation}, in International
  Conference on Machine Learning, PMLR, 2015, pp.~881--889.

\bibitem{glorot2010understanding}
{\sc X.~Glorot and Y.~Bengio}, {\em Understanding the difficulty of training
  deep feedforward neural networks}, in Proceedings of the thirteenth
  international conference on artificial intelligence and statistics, JMLR
  Workshop and Conference Proceedings, 2010, pp.~249--256.

\bibitem{goodfellow2020generative}
{\sc I.~Goodfellow, J.~Pouget-Abadie, M.~Mirza, B.~Xu, D.~Warde-Farley,
  S.~Ozair, A.~Courville, and Y.~Bengio}, {\em Generative adversarial
  networks}, Communications of the ACM, 63 (2020), pp.~139--144.

\bibitem{goswami2022deep}
{\sc S.~Goswami, K.~Kontolati, M.~D. Shields, and G.~E. Karniadakis}, {\em Deep
  transfer operator learning for partial differential equations under
  conditional shift}, Nature Machine Intelligence, 4 (2022), pp.~1155--1164.

\bibitem{guo2022normalizing}
{\sc L.~Guo, H.~Wu, and T.~Zhou}, {\em {Normalizing field flows: Solving
  forward and inverse stochastic differential equations using physics-informed
  flow models}}, Journal of Computational Physics, 461 (2022), p.~111202.

\bibitem{ho2020denoising}
{\sc J.~Ho, A.~Jain, and P.~Abbeel}, {\em Denoising diffusion probabilistic
  models}, Advances in Neural Information Processing Systems, 33 (2020),
  pp.~6840--6851.

\bibitem{karniadakis2021physics}
{\sc G.~E. Karniadakis, I.~G. Kevrekidis, L.~Lu, P.~Perdikaris, S.~Wang, and
  L.~Yang}, {\em Physics-informed machine learning}, Nature Reviews Physics, 3
  (2021), pp.~422--440.

\bibitem{kass1990validity}
{\sc R.~E. Kass}, {\em {The validity of posterior expansions based on Laplace's
  method}}, Bayesian and Likelihood Methods in Statistics and Econometrics,
  (1990), pp.~473--488.

\bibitem{khoo2021solving}
{\sc Y.~Khoo, J.~Lu, and L.~Ying}, {\em {Solving parametric PDE problems with
  artificial neural networks}}, European Journal of Applied Mathematics, 32
  (2021), pp.~421--435.

\bibitem{kingma2016improved}
{\sc D.~P. Kingma, T.~Salimans, R.~Jozefowicz, X.~Chen, I.~Sutskever, and
  M.~Welling}, {\em Improved variational inference with inverse autoregressive
  flow}, Advances in Neural Information Processing Systems, 29 (2016).

\bibitem{kingma2013auto}
{\sc D.~P. Kingma and M.~Welling}, {\em Auto-encoding variational bayes}, arXiv
  preprint arXiv:1312.6114,  (2013).

\bibitem{kobyzev2020normalizing}
{\sc I.~Kobyzev, S.~J. Prince, and M.~A. Brubaker}, {\em Normalizing flows: An
  introduction and review of current methods}, IEEE Transactions on Pattern
  Analysis and Machine Intelligence, 43 (2020), pp.~3964--3979.

\bibitem{lao2020tfp}
{\sc J.~Lao, C.~Suter, I.~Langmore, C.~Chimisov, A.~Saxena, P.~Sountsov,
  D.~Moore, R.~A. Saurous, M.~D. Hoffman, and J.~V. Dillon}, {\em tfp. mcmc:
  Modern {Markov chain Monte Carlo} tools built for modern hardware}, arXiv
  preprint arXiv:2002.01184,  (2020).

\bibitem{li2020fourier}
{\sc Z.~Li, N.~Kovachki, K.~Azizzadenesheli, B.~Liu, K.~Bhattacharya,
  A.~Stuart, and A.~Anandkumar}, {\em Fourier neural operator for parametric
  partial differential equations}, arXiv preprint arXiv:2010.08895,  (2020).

\bibitem{lin2021to}
{\sc Z.~Lin, Z.~Zhao, Z.~Zhang, H.~Baoxing, and J.~Yuan}, {\em {To Learn
  Effective Features: Understanding the Task-Specific Adaptation of MAML}},
  2021, \url{https://openreview.net/forum?id=FPpZrRfz6Ss}.

\bibitem{linka2022bayesian}
{\sc K.~Linka, A.~Schafer, X.~Meng, Z.~Zou, G.~E. Karniadakis, and E.~Kuhl},
  {\em {Bayesian Physics-Informed Neural Networks} for real-world nonlinear
  dynamical systems}, arXiv preprint arXiv:2205.08304,  (2022).

\bibitem{liu2022novel}
{\sc X.~Liu, X.~Zhang, W.~Peng, W.~Zhou, and W.~Yao}, {\em A novel
  meta-learning initialization method for physics-informed neural networks},
  Neural Computing and Applications,  (2022), pp.~1--24.

\bibitem{lu2021learning}
{\sc L.~Lu, P.~Jin, G.~Pang, Z.~Zhang, and G.~E. Karniadakis}, {\em Learning
  nonlinear operators via {DeepONet} based on the universal approximation
  theorem of operators}, Nature Machine Intelligence, 3 (2021), pp.~218--229.

\bibitem{mcclenny2020self}
{\sc L.~McClenny and U.~Braga-Neto}, {\em Self-adaptive physics-informed neural
  networks using a soft attention mechanism}, arXiv preprint arXiv:2009.04544,
  (2020).

\bibitem{meng2022learning}
{\sc X.~Meng, L.~Yang, Z.~Mao, J.~del {\'A}guila~Ferrandis, and G.~E.
  Karniadakis}, {\em Learning functional priors and posteriors from data and
  physics}, Journal of Computational Physics, 457 (2022), p.~111073.

\bibitem{neal2011mcmc}
{\sc R.~M. Neal et~al.}, {\em {MCMC} using {Hamiltonian} dynamics}, Handbook of
  Markov Chain Monte Carlo, 2 (2011), p.~2.

\bibitem{papamakarios2021normalizing}
{\sc G.~Papamakarios, E.~T. Nalisnick, D.~J. Rezende, S.~Mohamed, and
  B.~Lakshminarayanan}, {\em {Normalizing Flows for Probabilistic Modeling and
  Inference.}}, J. Mach. Learn. Res., 22 (2021), pp.~1--64.

\bibitem{papamakarios2017masked}
{\sc G.~Papamakarios, T.~Pavlakou, and I.~Murray}, {\em Masked autoregressive
  flow for density estimation}, Advances in Neural Information Processing
  Systems, 30 (2017).

\bibitem{penwarden2021physics}
{\sc M.~Penwarden, S.~Zhe, A.~Narayan, and R.~M. Kirby}, {\em Physics-informed
  neural networks ({PINNs}) for parameterized {PDEs}: A metalearning approach},
  arXiv preprint arXiv:2110.13361,  (2021).

\bibitem{perdikaris2016multifidelity}
{\sc P.~Perdikaris, D.~Venturi, and G.~E. Karniadakis}, {\em Multifidelity
  information fusion algorithms for high-dimensional systems and massive data
  sets}, SIAM Journal on Scientific Computing, 38 (2016), pp.~B521--B538.

\bibitem{psaros2022uncertainty}
{\sc A.~F. Psaros, X.~Meng, Z.~Zou, L.~Guo, and G.~E. Karniadakis}, {\em
  {Uncertainty Quantification in Scientific Machine Learning: Methods, Metrics,
  and Comparisons}}, arXiv preprint arXiv:2201.07766,  (2022).

\bibitem{raghu2019rapid}
{\sc A.~Raghu, M.~Raghu, S.~Bengio, and O.~Vinyals}, {\em {Rapid Learning or
  Feature Reuse? Towards Understanding the Effectiveness of MAML}}, arXiv
  preprint arXiv:1909.09157,  (2019).

\bibitem{raissi2019physics}
{\sc M.~Raissi, P.~Perdikaris, and G.~E. Karniadakis}, {\em Physics-informed
  neural networks: A deep learning framework for solving forward and inverse
  problems involving nonlinear partial differential equations}, Journal of
  Computational Physics, 378 (2019), pp.~686--707.

\bibitem{ruder2017overview}
{\sc S.~Ruder}, {\em An overview of multi-task learning in deep neural
  networks}, arXiv preprint arXiv:1706.05098,  (2017).

\bibitem{thanasutives2021adversarial}
{\sc P.~Thanasutives, M.~Numao, and K.-i. Fukui}, {\em Adversarial multi-task
  learning enhanced physics-informed neural networks for solving partial
  differential equations}, in 2021 International Joint Conference on Neural
  Networks (IJCNN), IEEE, 2021, pp.~1--9.

\bibitem{wang2021bridging}
{\sc H.~Wang, H.~Zhao, and B.~Li}, {\em Bridging multi-task learning and
  meta-learning: Towards efficient training and effective adaptation}, in
  International Conference on Machine Learning, PMLR, 2021, pp.~10991--11002.

\bibitem{wang2021learning}
{\sc S.~Wang, H.~Wang, and P.~Perdikaris}, {\em Learning the solution operator
  of parametric partial differential equations with physics-informed
  {DeepONets}}, Science Advances, 7 (2021), p.~eabi8605.

\bibitem{yang2021b}
{\sc L.~Yang, X.~Meng, and G.~E. Karniadakis}, {\em {B-PINNs}: Bayesian
  physics-informed neural networks for forward and inverse {PDE} problems with
  noisy data}, Journal of Computational Physics, 425 (2021), p.~109913.

\bibitem{yang2020physics}
{\sc L.~Yang, D.~Zhang, and G.~E. Karniadakis}, {\em Physics-informed
  generative adversarial networks for stochastic differential equations}, SIAM
  Journal on Scientific Computing, 42 (2020), pp.~A292--A317.

\bibitem{yang2022multi}
{\sc M.~Yang and J.~T. Foster}, {\em Multi-output physics-informed neural
  networks for forward and inverse {PDE} problems with uncertainties}, Computer
  Methods in Applied Mechanics and Engineering,  (2022), p.~115041.

\bibitem{zhong2022pi}
{\sc W.~Zhong and H.~Meidani}, {\em {PI-VAE}: {Physics-Informed Variational
  Auto-Encoder} for stochastic differential equations}, arXiv preprint
  arXiv:2203.11363,  (2022).

\bibitem{zou2022neuraluq}
{\sc Z.~Zou, X.~Meng, A.~F. Psaros, and G.~E. Karniadakis}, {\em {NeuralUQ}: A
  comprehensive library for uncertainty quantification in neural differential
  equations and operators}, arXiv preprint arXiv:2208.11866,  (2022).

\end{thebibliography}

\appendix
\section{Details of NN architectures and training hyperparameters}
\label{sec:nn}
For all examples in Secs.~\ref{sec:examples} and~\ref{sec:MTL}, MH-PINNs are implemented as fully-connected NNs (FNNs) with 3 nonlinear hidden layers, each of which is equipped with $50$ neurons and hyperbolic tangent activation function. The number of heads is the same as the number of tasks in the corresponding examples: $1,000$ in Sec.~\ref{subsec:example_1}, $2,000$ in Secs.~\ref{subsec:example_2},~\ref{subsec:example_3} and \ref{subsec:example_4}, and $10,000$ in Sec.~\ref{subsec:example_5}.
Weights in the body of MH-PINNs are initialized with Glorot uniform initialization \cite{glorot2010understanding} and biases are initialized with zero, while heads are initialized by sampling from random normal distribution with $0.05$ standard deviation, for fast training of NFs and better performance of the learned generators.

Except for the forward problem in Sec.~\ref{subsec:example_2}, NFs in this paper are chosen to be MAF \cite{papamakarios2017masked} with $10$ bijectors, i.e. the invertible map in NFs, each of which is a MADE \cite{germain2015made}, a NN with masked dense layers, with two nonlinear hidden layers equipped with $100$ neurons and ReLU activation function. The RealNVP \cite{dinh2016density} and IAF \cite{kingma2016improved} used in the forward problem in Sec.~\ref{subsec:example_2} also have $10$ bijectors, each of which is a NN with two nonlinear hidden layers equipped with $100$ neurons and ReLU activation function. The implementation mostly follows the instructions of TensorFlow Probability library \cite{dillon2017tensorflow} for NFs.

PI-GANs \cite{yang2020physics} implemented in Sec.~\ref{subsec:example_2} have the following architecture: the discriminator is a FNN with 3 nonlinear hidden layers, each of which is equipped with $128$ neurons and Leaky ReLU activation function; the generator that takes as input $t$ is a FNN with 3 nonlinear hidden layers, each of which is equipped with $50$ neurons and hyperbolic tangent activation function; the other generator takes as input a Gaussian random variable in $50$ dimensions with zero mean and identity covariance matrix, and is implemented as a FNN with 3 nonlinear hidden layers, each of which has $128$ neurons and hyperbolic tangent activation function. The input dimensions of those 3 FNNs are $65$, $1$ and $50$, and the output dimensions are $1$, $50$, $50$, respectively.

For the training of MH-PINNs, full-batch training is deployed with Adam optimizer for $50,000$ iterations. For the training of NFs, except for the forward problem in Sec.~\ref{subsec:example_2}, mini-batch training is deployed with batch size being $100$ and Adam optimizer for $1,000$ epochs. NFs in the forward problem in Sec.~\ref{subsec:example_2} are trained for $500$ epochs instead, and $L_2$ regularization is imposed to the parameters of RealNVP for better performance. For all NFs, to achieve stable training, a hyperbolic tangent function is imposed on the logarithm of the scale, computed from each bijector, such that the logarithm of the scale lies in $(-1, 1)$. For the training of PI-GANs, min-batch training is deployed with batch size being $100$ and Adam optimizer for $100,000$ iterations. Besides, the same as in \cite{yang2020physics, meng2022learning}, physics-informed Wasserstein GANs (PI-WGANs) with gradient penalty are employed, in which the coefficient for gradient penalty is set to be $0.1$. Iteratively, 5 updates of the discriminator are performed and followed by 1 update of the generators. Except in training PI-GANs, the learning rate of Adam optimizer is set to be $10^{-3}$ and other hyperparameters of Adam are set as default. In training PI-GANs, the learning rate is set to be $10^{-4}$, $\beta_1=0.5$ and $\beta_2=0.9$ in Adam optimizer for both discriminator and generators.

Training of MH-PINNs, NFs, and PI-GANs was all performed on a single NVIDIA TITAN Xp GPU. The L-HYDRA code for TensorFlow implementation along with some representative examples will be released on GitHub once the paper is accepted.

\section{Details for performing Bayesian inference}
\label{sec:hmc}
Hamiltonian Monte Carlo (HMC) \cite{neal2011mcmc} is employed in all Bayesian inference examples for uncertainty quantification (UQ) while Laplace approximation \cite{kass1990validity} is only employed in the first example. In this paper, HMC with adaptive step size \cite{lao2020tfp} is used, in which the initial step size is set to be either $0.1$ or $0.01$, tuned for better acceptance rate. The number of burn-in samples and the number of posterior samples are set to be $1,000$. The number of steps for the leapfrog scheme is set to be either $30$ or $50$, also tuned for better acceptance rate. NeuralUQ library \cite{zou2022neuraluq} for UQ in scientific machine learning is used as a tool for physics-informed Bayesian inference in the downstream tasks. The ideal acceptance rate in HMC, as discussed in \cite{meng2022learning, zou2022neuraluq}, is around $60\%$. In this paper, we found chains with acceptance rate from $50\%$ to $80\%$ acceptable. 
\end{document}